\RequirePackage{fix-cm}
\documentclass[smallextended]{svjour3}       
\smartqed  
\usepackage{graphicx}
\usepackage{natbib}
\usepackage{hyperref}
\usepackage{amsmath}
\usepackage{amssymb}
\usepackage{algorithm}
\usepackage{algpseudocode}
\usepackage{multirow}
\usepackage{booktabs}
\usepackage{array}
\usepackage{pifont}
\usepackage{adjustbox}
\usepackage{textcomp}
\usepackage{placeins}
\usepackage[T1]{fontenc}

\newcommand{\Tau}{\mathcal{T}}

\newcommand{\xmark}{\ding{55}}
\usepackage{tabularx}

\DeclareMathOperator*{\argmax}{arg\,max}
\DeclareMathOperator*{\argmin}{arg\,min}
\begin{document}

\title{A Survey of Deep Meta-Learning
}


\author{Mike Huisman         \and
        Jan N. van Rijn      \and 
        Aske Plaat
}


\institute{M. Huisman \and J.N. van Rijn \and A. Plaat \at
              Leiden Institute of Advanced Computer Science \\
              Niels Bohrweg 1, 2333CA Leiden, The Netherlands \\
              \email{m.huisman@liacs.leidenuniv.nl}           %
}

\date{}

\maketitle

\begin{abstract}
Deep neural networks can achieve great successes when presented with large data sets and sufficient computational resources.
However, their ability to learn new concepts \textit{quickly} is limited. 
Meta-learning is one approach to address this issue, by enabling the network to learn how to learn. 
The field of \textit{Deep Meta-Learning} advances at great speed, but lacks a unified, in-depth overview of current techniques. 
With this work, we aim to bridge this gap. After providing the reader with a theoretical foundation, we investigate and summarize key methods, which are categorized into i)~metric-, ii)~model-, and iii)~optimization-based techniques. In addition, we identify the main open challenges, such as performance evaluations on heterogeneous benchmarks, and reduction of the computational costs of meta-learning. 
\keywords{Meta-learning \and Learning to learn \and Few-shot learning \and Transfer learning \and Deep learning}
\end{abstract}

\section{Introduction}\label{sec:intro}

In recent years, deep learning techniques have achieved remarkable successes on various tasks, including game-playing \citep{mnih2013playing, silver2016mastering}, image recognition \citep{krizhevsky2012imagenet,he2015delving}, machine translation \citep{wu2016google}, and automatic classification in biomedical domains \citep{goceri2019capsnet, goceri2020comparative, iqbal2020deep, iqbal2019mitochondrial, iqbal2019efficient}. Despite these advances and recent solutions \citep{goceri2019challenges, gocceri2020convolutional}, ample challenges remain to be solved, such as the large amounts of data and training that are needed to achieve good performance. These requirements severely constrain the ability of deep neural networks to learn new concepts quickly, one of the defining aspects of  human intelligence \citep{jankowski2011meta, lake2017building}.

\textit{Meta-learning} has been suggested as one strategy to overcome this challenge \citep{naik1992meta, schmidhuber1987evolutionary, thrun1998lifelong}. The key idea is that meta-learning agents improve their learning ability over time, or equivalently, learn to learn. The learning process is primarily concerned with tasks (set of observations) and takes place at two different levels: an inner- and an outer-level. At the \textit{inner-level}, a new task is presented, and the agent tries to quickly learn the associated concepts from the training observations. This quick adaptation is facilitated by  knowledge that it has accumulated across earlier tasks at the \textit{outer-level}. Thus, whereas the inner-level concerns a single task, the outer-level concerns a multitude of tasks. 

\begin{figure}[tb]
    \centering
    \includegraphics[width=\linewidth]{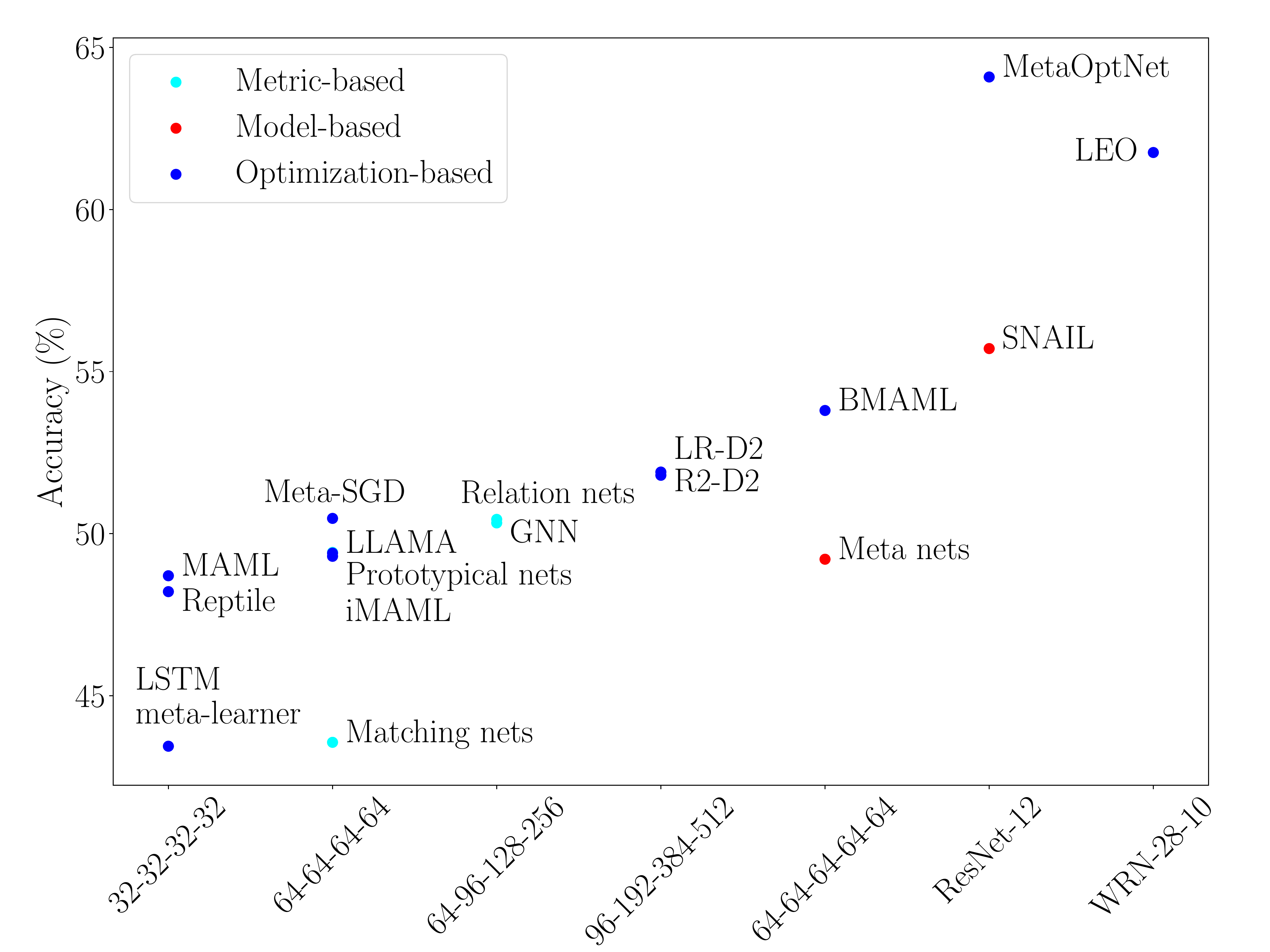}
    \caption{The accuracy scores of the covered techniques on 1-shot miniImageNet classification. The used feature extraction backbone is displayed on the x-axis. As one can see, there is a strong relationship between the network complexity and the classification performance.}
    \label{fig:depthandperformance}
\end{figure}

Historically, the term meta-learning has been used with various scopes. In its broadest sense, it encapsulates all systems that leverage prior learning experience in order to learn new tasks more quickly \citep{vanschoren2018meta}. This broad notion includes more traditional algorithm selection and hyperparameter optimization techniques for Machine Learning \citep{brazdil2008metalearning}. In this work, however, we focus on a subset of the meta-learning field which develops meta-learning procedures to learn a good \textit{inductive bias} for (deep) neural networks.\footnote{Here, inductive bias refers to the assumptions of a model which guide predictions on unseen data \citep{mitchell1980need}.} Henceforth, we use the term \textit{Deep Meta-Learning} to refer to this subfield of meta-learning. 

The field of Deep Meta-Learning is advancing at a quick pace, while it lacks a coherent, unifying overview, providing detailed insights into the key techniques. \citet{vanschoren2018meta} has surveyed meta-learning techniques, where meta-learning was used in the broad sense, limiting  its account of Deep Meta-Learning techniques. Also, many exciting developments in deep meta-learning have happened after the survey was published. A more recent survey by \citet{hospedales2020meta} adopts the same notion of deep meta-learning as we do, but aims to give a broad overview, omitting technical details of the various techniques.

We attempt to fill this gap by providing detailed explications of contemporary Deep Meta-Learning techniques, using a unified notation. 
More specifically, we cover modern techniques in the field for supervised and reinforcement learning, that have achieved state-of-the-art performance, obtained popularity in the field, and presented novel ideas. 
Extra attention is paid to MAML \citep{Finn17}, and related techniques, because of their impact on the field.
We show how the techniques relate to each other, detail their strengths and weaknesses,  identify current challenges, and provide an overview of promising future research directions.
One of the observations that we make is that the network complexity is highly related to the few-shot classification performance (see \autoref{fig:depthandperformance}).
One might expect that in a few-shot setting, where only a few examples are available to learn from, the number of network parameters should be kept small to prevent overfitting. 
Clearly, the figure shows that this does not hold, as techniques that use larger backbones tend to achieve better performance. 
One important factor might be that due to the large number of tasks that have been seen by the network, we are in a setting where similarly large amounts of observations have been evaluated.
This result suggests that the size of the network should be taken into account when comparing algorithms.

This work can serve as an educational introduction to the field of Deep Meta-Learning, and as reference material for  experienced researchers in the field. Throughout, we will adopt the taxonomy used by \citet{vinyals2017talk}, which identifies three categories of Deep Meta-Learning approaches: i)~metric-based, ii)~model-based, and iii)~optimization-based meta-learning techniques.   

The remainder of this work is structured as follows. Section 2 builds a common foundation on which we will base our overview of Deep Meta-Learning techniques. Sections 3, 4, and 5 cover the main metric-, model-, and optimization-based meta-learning techniques, respectively. Section 6 provides a helicopter view of the field and summarizes the key challenges and open questions. \autoref{tab:notation} gives an overview of notation that we will use throughout this paper. 

\begin{table}[tb]
    \begin{tabularx}{\linewidth}{ll}
    \toprule
         Expression & Meaning  \\
         \midrule
         Meta-learning & Learning to learn \\
         $\Tau_{j} = (D^{tr}_{\Tau_{j}}, D^{test}_{\Tau_{j}})$& A task consisting of a labeled support and query set \\ 
         Support set & The train set $D^{tr}_{\Tau_{j}}$ associated with a task $\Tau_{j}$ \\
         Query set & The test set $D^{test}_{\Tau_{j}}$ associated with a task $\Tau_{j}$ \\
         $\boldsymbol{x}_{i}$ & Example input vector $i$ in the support set \\
         $y_{i}$ & (One-hot encoded) label of example input $\boldsymbol{x}_{i}$ from the support set\\
         $k$ & Number of examples per class in the support set \\
         $N$ & Number of classes in the support and query sets of a task \\
         $\boldsymbol{x}$ & Input in the query set \\
         $y$ & A (one-hot encoded) label for input $\boldsymbol{x}$ \\
         $(f/g/h)_{\circ}$ & Neural network function with parameters $\circ$ \\
         Inner-level & At the level of a single task \\
         Outer-level & At the meta-level: across tasks \\
         Fast weights & A term used in the literature to denote task-specific parameters \\
         Base-learner & Learner that works at the inner-level \\
         Meta-learner & Learner that operates at the outer-level \\
         $\boldsymbol{\theta}$ & The parameters of the base-learner network \\
         $\mathcal{L}_D$ & Loss function with respect to task/dataset $D$ \\
         Input embedding & Penultimate layer representation of the input \\
         Task embedding & An internal representation of a task in a network/system \\
         SL & Supervised Learning \\
         RL & Reinforcement Learning \\
         \bottomrule
    \end{tabularx}
    \caption{Some notation and meaning, which we use throughout this paper.}
    \label{tab:notation}
\end{table}

\section{Foundation}
\label{sec:foundation}
In this section, we build the necessary foundation for investigating Deep Meta-Learning techniques in a consistent manner. To begin with, we contrast regular learning and meta-learning. Afterwards, we briefly discuss how Deep Meta-Learning relates to different fields, what the usual training and evaluation procedure looks like, and which benchmarks are often used for this purpose. We finish this section by describing the context and some applications of the meta-learning field. 

\subsection{The Meta Abstraction}

In this subsection, we contrast base-level (regular) learning and meta-learning for two different paradigms, i.e., supervised and reinforcement learning.  

\subsubsection{Regular Supervised Learning}
In \textit{supervised learning}, we wish to learn a function $f_{\boldsymbol{\theta}}: X \rightarrow Y$ that learns to map inputs $\boldsymbol{x}_{i} \in X$ to their corresponding outputs $y_{i} \in Y$. Here, $\boldsymbol{\theta}$ are model parameters (e.g.\ weights in a neural network) that determine the function's behavior. To learn these parameters, we are given a data set of $m$ observations: $D = \{(\boldsymbol{x}_{i}, y_{i})\}_{i=1}^{m}$. Thus, given a data set $\mathcal{D}$, learning boils down to finding the correct setting for $\boldsymbol{\theta}$ that minimizes an empirical loss function $\mathcal{L}_{D}$, which must capture how the model is performing, such that appropriate adjustments to its parameters can be made. In short, we wish to find 
\begin{align}
    \boldsymbol{\theta}_{SL} := \argmin_{\boldsymbol{\theta}} \, \mathcal{L}_{D}(\boldsymbol{\theta}),
\end{align}\label{eq:suplearning}where $SL$ stands for ``supervised learning". Note that this objective is specific to data set $\mathcal{D}$, meaning that our model $f_{\boldsymbol{\theta}}$ may not \textit{generalize} to examples outside of $\mathcal{D}$. To measure generalization, one could evaluate the performance on a separate test data set, which contains unseen examples. A popular way to do this is through \textit{cross-validation}, where one repeatedly creates train and test splits $D^{tr}, D^{test} \subset D$ and uses these to train and evaluate a model respectively \citep{hastie2009elements}. 

Finding globally optimal parameters $\boldsymbol{\theta}_{SL}$ is often computationally infeasible. We can, however, approximate them,  guided by \textit{pre-defined} meta-knowledge $\omega$ \citep{hospedales2020meta}, which includes, e.g., the initial model parameters $\boldsymbol{\theta}$, choice of optimizer, and learning rate schedule. As such, we approximate 
\begin{align}
    \boldsymbol{\theta}_{SL} \approx g_{\omega}(D, \mathcal{L}_{D}),
\end{align}
where $g_{\omega}$ is an optimization procedure that uses \textit{pre-defined} meta-knowledge $\omega$, data set $\mathcal{D}$, and loss function $\mathcal{L}_{D}$, to produce updated weights $g_{\omega}(D, \mathcal{L}_{D})$ that (presumably) perform well on $\mathcal{D}$. 

\subsubsection{Supervised Meta-Learning}
In contrast, \textit{supervised meta-learning} does not  assume that any meta-knowledge $\omega$ is given, or pre-defined. Instead, the goal of meta-learning is to find the best $\omega$, such that our (regular) base-learner can learn new \textit{tasks} (data sets) as quickly as possible. Thus, whereas  supervised regular learning involves one data set, supervised meta-learning involves a group of data sets. The goal is  to learn meta-knowledge $\omega$ such that our model can learn many different tasks well. Thus, our model is learning to learn.  

More formally, we have a probability distribution of tasks $p(\Tau)$ and wish to find optimal meta-knowledge
\begin{align}
    \omega^{*} := \argmin_{\omega} \, \underbrace{\mathbb{E}_{\Tau_{j} \backsim p(\Tau)}}_\textrm{Outer-level} [ \underbrace{ \mathcal{L}_{\Tau_{j}}(g_{\omega}(\Tau_{j}, \mathcal{L}_{\Tau_{j}}))}_\textrm{Inner-level}] .
\end{align}
Here, the inner-level concerns task-specific learning, while the outer-level concerns multiple tasks.
One can now easily see why this is meta-learning: we learn $\omega$, which allows for quick learning of tasks $\Tau_{j}$ at the inner-level. Hence, we are learning to learn.

\subsubsection{Regular Reinforcement Learning}\label{sec:RL}

In \textit{reinforcement learning}, we have an agent that learns from  experience. That is, it interacts with an environment, modeled by a Markov Decision Process (MDP) $M = (S, A, P, r, p_{0}, \gamma, T)$. Here, $S$ is the set of states, $A$ the set of actions, $P$ the transition probability distribution defining $P(s_{t+1}| s_{t}, a_{t})$, $r: S \times A \rightarrow \mathbb{R}$ the reward function, $p_{0}$ the probability distribution over initial states, $\gamma \in [0,1]$ the discount factor, and $T$ the time horizon (maximum number of time steps) \citep{sutton2018reinforcement, duan2016rl}. 

At every time step $t$, the agent finds itself in state $s_{t}$, in which the agent performs an action $a_{t}$, computed by a policy function $\pi_{\boldsymbol{\theta}}$ (i.e., $a_{t} = \pi_{\boldsymbol{\theta}}(s_{t})$), which is parameterized by weights $\boldsymbol{\theta}$. In turn, it receives a reward $r_{t} = r(s_{t}, \pi_{\boldsymbol{\theta}}(s_{t})) \in \mathbb{R}$ and a new state $s_{t+1}$. This process of interactions continues until a termination criterion is met (e.g.\ fixed time horizon $T$ reached). The goal of the agent is to learn how to act in order to maximize its expected reward. The reinforcement learning (RL) goal is to find 

\begin{align}
    \boldsymbol{\theta}_{RL} := \argmin_{\boldsymbol{\theta}} \, \mathbb{E}_{\mbox{traj}} \sum_{t=0}^{T} \gamma^{t}r(s_{t}, \pi_{\boldsymbol{\theta}}(s_{t})),
\end{align}
where we take the expectation over the possible \textit{trajectories} $\mbox{traj} = (s_{0}, \pi_{\boldsymbol{\theta}}(s_{0}), \allowbreak \ldots s_{T}, \pi_{\boldsymbol{\theta}}(s_{T}))$ due to the random nature of MDPs \citep{duan2016rl}. Note that $\gamma$ is a hyperparameter that can prioritize short- or long-term rewards by decreasing or increasing it, respectively.

Also in the case of reinforcement learning it is often infeasible to find  the global optimum $\boldsymbol{\theta}_{RL}$, and thus we  settle for  approximations. In short, given a learning method $\omega$, we approximate 
\begin{align}
    \boldsymbol{\theta}_{RL} \approx g_{\omega}(\Tau_{j}, \mathcal{L}_{\Tau_{j}}),
\end{align}
where again $\Tau_{j}$ is the given MDP, and $g_{\omega}$ is the optimization algorithm, guided by pre-defined meta-knowledge $\omega$.

Note that in a Markov Decision Process (MDP), the agent knows the state at any given time step $t$. When this is not the case, it becomes a Partially Observable Markov Decision Process (POMDP), where the agent receives only observations $O$, and uses these to update its belief with regard to the state it is in \citep{sutton2018reinforcement}.

\subsubsection{Meta Reinforcement Learning}
The meta abstraction has as its object a group of tasks, or Markov Decision Processes (MDPs) in the case of reinforcement learning. Thus, instead of maximizing the expected reward on a single MDP, the meta reinforcement learning objective is to maximize the expected reward over various MDPs, by learning meta-knowledge $\omega$. Here, the MDPs are sampled from some distribution $p(\Tau)$. So, we wish to find a set of parameters 
\begin{align}
    \boldsymbol{\omega}^{*} := \argmin_{\boldsymbol{\omega}} \, \underbrace{\mathbb{E}_{\Tau_{j} \backsim p(\Tau)}}_\textrm{Outer-level} \left[ \underbrace{\mathbb{E}_{traj} \sum_{t=0}^{T} \gamma^{t}r(s_{t}, \pi_{g_{\omega}(\Tau_{j}, \mathcal{L}_{\Tau_{j}})}(s_{t}))}_\textrm{Inner-level} \right].
\end{align}

\subsubsection{Contrast with other Fields}
Now that we have provided a formal basis for our discussion for both supervised and reinforcement meta-learning, it is time to contrast meta-learning briefly with two related areas of machine learning that also have the goal to improve the speed of learning. We will start with transfer learning.

\textbf{Transfer Learning}
In Transfer Learning, one tries to \textit{transfer} knowledge of previous tasks to new, unseen tasks \citep{pan2009survey, taylor2009transfer}, which can be challenging when the new task comes from a different distribution than the one used for training  \cite{iqbal2018heterogeneous}. The distinction between Transfer Learning and Meta-Learning has become more opaque over time. 
A key property of meta-learning techniques, however, is their \textit{meta-objective}, which explicitly aims to optimize performance across a distribution over tasks (as seen in previous sections by taking the expected loss over a distribution of tasks). This objective need not always be present in Transfer Learning techniques, e.g., when one \textit{pre-trains} a model on a large data set, and \textit{fine-tunes} the learned weights on a smaller data set.

\begin{figure}[tb]
    \centering
    \includegraphics[width=\linewidth]{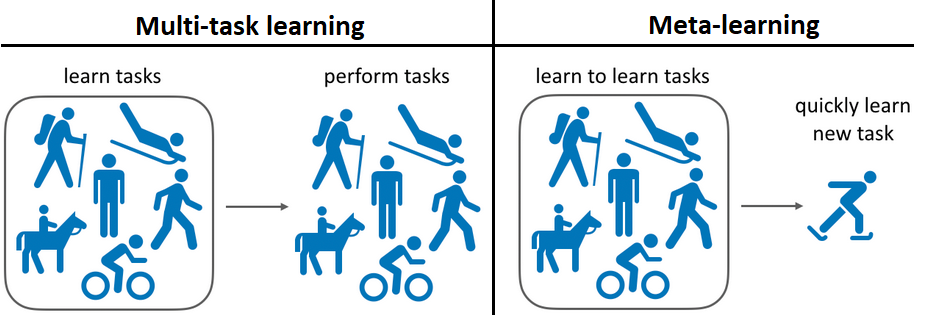}
    \caption[The difference between multi-task learning and meta-learning]{The difference between multi-task learning and meta-learning\protect\footnotemark.}
    \label{fig:multitask}
\end{figure}

\textbf{Multi-task learning}
Another,  closely related field, is that of multi-task learning. In multi-task learning, a model is jointly trained to perform well on multiple fixed tasks \citep{hospedales2020meta}. Meta-learning, in contrast, aims to find a model that can learn new (previously unseen) tasks quickly. This difference is illustrated in \autoref{fig:multitask}. 

\FloatBarrier
\footnotetext{Adapted from~\url{https://meta-world.github.io/}}

\subsection{The Meta-Setup}
In the previous section, we have described the learning objectives for (meta) supervised and reinforcement learning. We will now describe the general setting that can be  used  to achieve these objectives. In general, one optimizes a meta-objective by using various tasks, which are data sets in the context of supervised learning, and (Partially Observable) Markov Decision Processes in the case of reinforcement learning. 
This is done in three stages: the i)~\textit{meta-train} stage, ii)~\textit{meta-validation} stage, and iii)~\textit{meta-test} stage, each of which is associated with a set of tasks. 

First, in the meta-train stage, the meta-learning algorithm is applied to the meta-train tasks. Second, the meta-validation tasks can then be used to evaluate the performance on unseen tasks, which were not used for training. Effectively, this measures the \textit{meta-generalization} ability of the trained network, which serves as feedback to tune, e.g., hyper-parameters of the meta-learning algorithm.
Third, the meta-test tasks are used to give a final performance estimate of the meta-learning technique. 

\begin{figure}[tb!]
    \centering
    \includegraphics[scale=0.38]{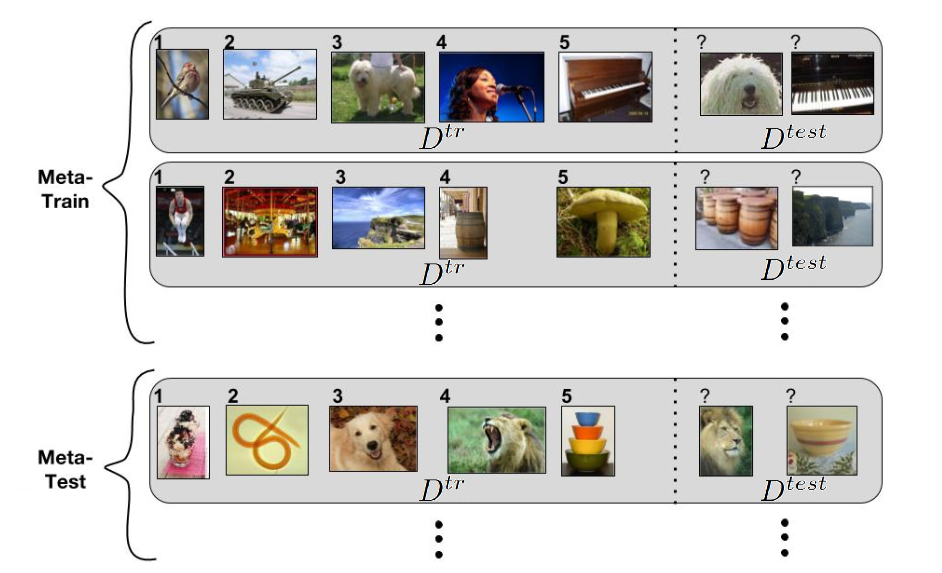}
    \caption{Illustration of $N$-way, $k$-shot classification, where $N = 5$, and $k = 1$. Meta-validation tasks are not displayed. Adapted from \citet{Ravi2017}. }
    \label{fig:fewshot}
\end{figure}

\subsubsection{\texorpdfstring{$N$}{N}-way, \texorpdfstring{$k$}{k}-shot Learning}

A frequently used instantiation of this general meta-setup is called $N$-way, $k$-shot\index{$n$-way, $k$-shot learning} classification (see \autoref{fig:fewshot}). This setup is also divided into the three stages---meta-train, meta-validation, and meta-test---which are used for meta-learning, meta-learner hyperparameter optimization, and evaluation, respectively. Each stage has a corresponding set of disjoint labels, i.e., $L^{tr}, L^{val}, L^{test} \subset Y$, such that $L^{tr} \cap L^{val} = \emptyset, L^{tr} \cap L^{test} = \emptyset$, and $L^{val} \cap L^{test} = \emptyset$. In a given stage $s$, \textit{tasks/episodes} $\Tau_{j} = (D^{tr}_{\Tau_{j}}, D^{test}_{\Tau_{j}})$ are obtained by sampling examples $(\boldsymbol{x}_{i}, y_{i})$ from the full data set $\mathcal{D}$, such that every $y_{i} \in L^{s}$. Note that this requires access to a data set $\mathcal{D}$. The sampling process is guided by the $N$-way, $k$-shot principle, which states that every training data set $D^{tr}_{\Tau_{j}}$ should contain exactly $N$ classes and $k$ examples per class, implying that $|D^{tr}_{\Tau_{j}}| = N \cdot k$. Furthermore, the true labels of examples in the test set $D_{\Tau_{j}}^{test}$ must be present in the train set $D^{tr}_{\Tau_{j}}$ of a given task $\Tau_{j}$. $D^{tr}_{\Tau{j}}$ acts as a \textit{support set}\index{support set}, literally supporting classification decisions on the \textit{query set}\index{query set} $D^{test}_{\Tau_{j}}$.  Importantly, note that with this terminology, the query set (or test set) of a task is actually used during the meta-training phase. Furthermore, the fact that the labels across stages are disjoint ensures that we test the ability of a model to learn \textit{new} concepts. 

The meta-learning objective in the training phase is to minimize the loss function of the model predictions on the query sets, conditioned on the support sets. As such, for a given task $\Tau_j$, the model `sees' the support set, and extracts information from the support set to guide its predictions on the query set. By applying this procedure to different episodes/tasks 
$\Tau_j$, the model will slowly accumulate meta-knowledge $\omega$, which can ultimately speed up learning on new tasks. 

The easiest way to achieve this is by doing this with regular neural networks, but as was pointed out by various authors (see, e.g., \cite{Finn17}) more sophisticated architectures will vastly outperform such networks. In the remainder of this work, we will review such architectures. 

At the meta-validation and meta-test stages, or evaluation phases, the learned meta-information in $\omega$ is fixed. The model is, however, still allowed to make task-specific updates to its parameters $\boldsymbol{\theta}$ (which implies that it is learning). After task-specific updates, we can evaluate the performance on the test sets. In this way, we test how well a technique performs at meta-learning.

$N$-way, $k$-shot classification is often performed for small values of $k$ (since we want our models to learn new concepts quickly, i.e., from few examples). In that case, one can refer to it as \textit{few-shot learning}\index{few-shot learning}.

\subsubsection{Common Benchmarks}\label{sec:benchmarks}
Here, we briefly describe some benchmarks that can be used to evaluate meta-learning algorithms. 

\begin{itemize}
    \item \textbf{Omniglot \citep{lake2011one}:} This data set presents an image recognition task. Each image corresponds to one out of 1\,623 characters from 50 different alphabets. Every character was drawn by 20 people. Note that in this case, the characters are the classes/labels. \item \textbf{ImageNet \citep{deng2009imagenet}:} This is the largest image classification data set, containing more than 20K classes and over 14 million colored images. \textit{miniImageNet} is a mini variant of the large ImageNet data set \citep{deng2009imagenet} for image classification, proposed by \citet{vinyals2016matching} to reduce the engineering efforts to run experiments. The mini data set contains 60\,000 colored images of size $84 \times 84$. There are a total of 100 classes present, each accorded by 600 examples. \textit{tieredImageNet} \citep{ren2018meta} is another variation of the large ImageNet data set. It is similar to miniImageNet, but contains a hierarchical structure. That is, there are 34 classes, each with its own sub-classes. 
    \item \textbf{CIFAR-10 and CIFAR-100 \citep{krizhevsky2009cifar}}: Two other image recognition data sets. Each one contains 60K RGB images of size $32 \times 32$. CIFAR-10 and CIFAR-100 contain 10 and 100 classes respectively, with a uniform number of examples per class (6\,000 and 600 respectively). Every class in CIFAR-100 also has a super-class, of which there are 20 in the full data set. Many variants of the CIFAR data sets can be sampled, giving rise to e.g.\ \textit{CIFAR-FS} \citep{Bertinetto19} and \textit{FC-100} \citep{oreshkin2018tadam}.
    \item \textbf{CUB-200-2011 \citep{wah2011caltech}:} The CUB-200-2011 data set contains roughly 12K RGB images of birds from 200 species. Every image has some labeled attributes (e.g.\ crown color, tail shape).
    \item \textbf{MNIST \citep{lecun2010mnist}:} MNIST presents a hand-written digit recognition task, containing ten classes (for digits 0 through 9). In total, the data set is split into a 60K train and 10K test gray scale images of hand-written digits.    
    \item \textbf{Meta-Dataset \citep{triantafillou2019meta}:} This data set comprises several other data sets such as Omniglot \citep{lake2011one}, CUB-200 \citep{wah2011caltech}, ImageNet \citep{deng2009imagenet}, and more \citep{triantafillou2019meta}. An episode is then constructed by sampling a data set (e.g.\ Omniglot) and selecting a subset of labels to create train and test splits as before. In this way, broader generalization is enforced since the tasks are more distant from each other.  
    \item \textbf{Meta-world \citep{yu2019meta}}: A meta reinforcement learning data set, containing 50 robotic manipulation tasks (control a robot arm to achieve some pre-defined goal, e.g.\ unlocking a door, or playing soccer). It was specifically designed to cover a broad range of tasks, such that meaningful generalization can be measured \citep{yu2019meta}.
\end{itemize}

\begin{figure}[tb]
    \centering
    \includegraphics[width=\linewidth]{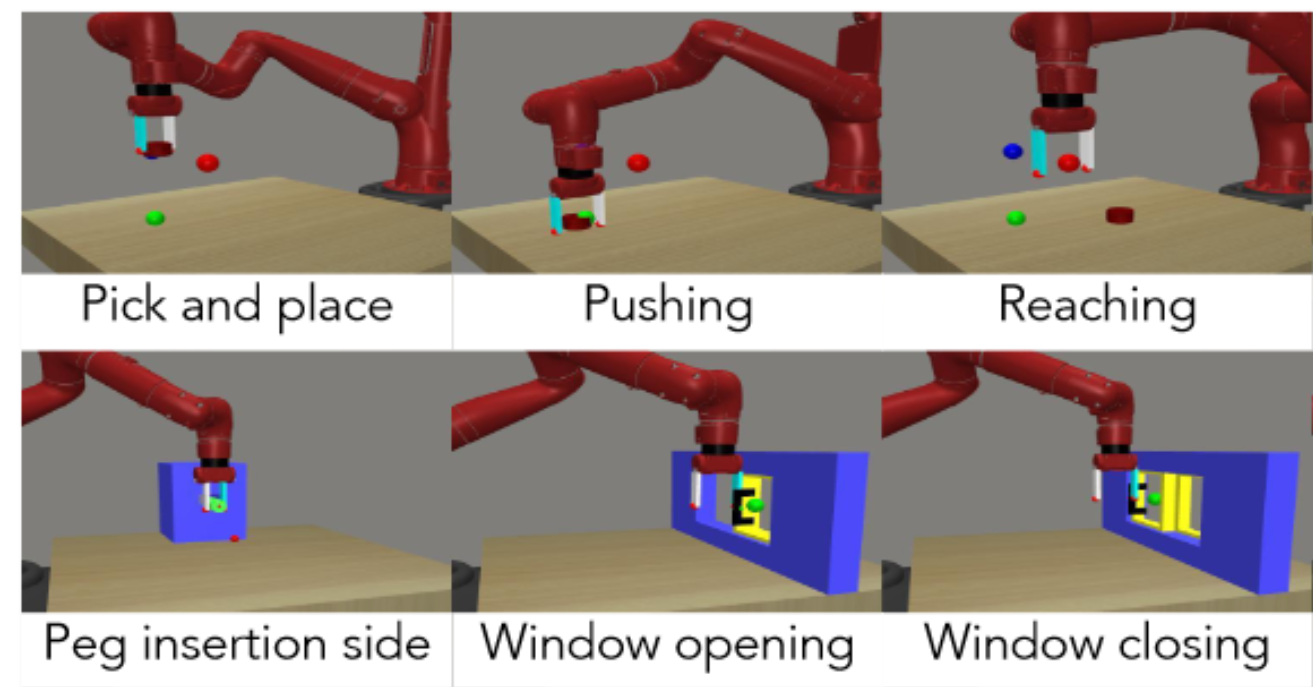}
    \caption{Learning continuous robotic control tasks is an important application of Deep Meta-Learning techniques. Image taken from \citep{yu2019meta}.}
    \label{fig:robot}
\end{figure}

\subsubsection{Some Applications of Meta-Learning}
Deep neural networks have achieved remarkable results on various tasks including image recognition, text processing, game playing, and robotics \citep{silver2016mastering, mnih2013playing, wu2016google}, but their success depends on the amount of available data \citep{sun2017revisiting} and computing resources. Deep meta-learning reduces this dependency by allowing deep neural networks to learn new concepts quickly. As a result, meta-learning widens the applicability of deep learning techniques to many application domains. Such areas include few-shot image classification \citep{Finn17, snell2017prototypical,Ravi2017}, robotic control policy learning \citep{gupta2018meta, nagabandi2018learning} (see \autoref{fig:robot}), hyperparameter optimization \citep{antoniou2018how, schmidhuber1997shifting}, meta-learning learning rules \citep{bengio1990learning, bengio1992optimization, miconi2018differentiable, miconi2020backpropamine}, abstract reasoning \citep{barrett2018measuring}, and many more. For a larger overview of applications, we refer interested readers to \citet{hospedales2020meta}.     

\subsection{The Meta-Learning Field}
As mentioned in the introduction, meta-learning is a broad area of research, as it encapsulates all techniques that leverage prior learning experience to learn new tasks more quickly \citep{vanschoren2018meta}. We can classify two distinct communities in the field with a different focus: i) algorithm selection and hyperparameter optimization for machine learning techniques, and ii) search for inductive bias in deep neural networks. We will refer to these communities as group i) and group ii) respectively. Now, we will give a brief description of the first field, and a historical overview of the second.

Group i) uses a more traditional approach, to select a suitable machine learning algorithm and hyperparameters for a new data set $\mathcal{D}$ \citep{peng2002improved}. This selection can for example be made by leveraging prior model evaluations on various data sets $D'$, and by using the model which achieved the best performance on the most similar data set \citep{vanschoren2018meta}. Such traditional approaches require (large) databases of prior model evaluations, for many different algorithms. This has led to initiatives such as OpenML \citep{vanschoren2014openml}, where researchers can share such information. 
The usage of these systems would limit the freedom in picking the neural network architecture as they would be constrained to using architectures that have been evaluated beforehand.  

In contrast, group ii) adopts the view of a self-improving (neural) agent, which improves its learning ability over time by finding a good inductive bias (a set of assumptions that guide predictions).
We now present a brief historical overview of developments in this field of Deep Meta-Learning, based on \citet{hospedales2020meta}. 

Pioneering work was done by \citet{schmidhuber1987evolutionary} and \citet{hinton1987using}. Schmidhuber developed a theory of \textit{self-referential} learning, where the weights of a neural network can serve as input to the model itself, which then predicts updates \citep{schmidhuber1987evolutionary, schmidhuber1993neural}. In that same year, \citet{hinton1987using} proposed to use two weights per neural network connection, i.e., \textit{slow} and \textit{fast} weights, which serve as long- and short-term memory respectively. Later came the idea of meta-learning learning rules \citep{bengio1990learning, bengio1992optimization}. Meta-learning techniques that use gradient-descent and backpropagation were proposed by \citet{hochreiter2001learning} and \citet{younger2001meta}. These two works have been pivotal to the current field of Deep Meta-Learning, as the majority of techniques rely on backpropagation, as we will see on our journey of contemporary Deep Meta-Learning techniques. 

\subsection{Overview of the rest of this Work}
In the remainder of this work, we will look in more detail at individual meta-learning methods. 
As indicated before, the techniques can be grouped into three main categories \citep{vinyals2017talk}, namely
i)~metric-, ii)~model-, and iii)~opti\-mization-based methods. We will discuss them in that order. 

To help give an overview of the methods, we draw your attention to the following tables. 
\autoref{tab:categorization} summarizes the three categories and provides  key ideas, and strengths of the approaches. The terms and technical details are explained more fully in the remainder of this paper. \autoref{tab:overview} contains an  overview of all techniques that are discussed further on.

\begin{table}[tb]
    \begin{tabularx}{\linewidth}{ccXX}
    \toprule
         & \textbf{Metric} & \textbf{Model} & \textbf{Optimization} \\
        \midrule
         \textbf{Key idea}& Input similarity & Internal task \hfill \allowbreak representation & Optimize for fast \hfill \allowbreak adaptation \hfill \\
         \textbf{Strength}& Simple and effective & Flexible & More robust generalizability \\
         \textbf{$p_{\boldsymbol{\theta}}(Y | \boldsymbol{x}, D^{tr}_{\Tau_{j}})$} & $\sum\limits_{(\boldsymbol{x}_{i}, y_{i}) \in D^{tr}_{\Tau_{j}}} k_{\boldsymbol{\theta}}(\boldsymbol{x}, \boldsymbol{x}_{i})y_{i}$ & $f_{\boldsymbol{\theta}}(\boldsymbol{x}, D^{tr}_{\Tau_{j}})$ & $f_{g_{\boldsymbol{\varphi}(\boldsymbol{\theta}, D_{\Tau_{j}}^{tr}, \mathcal{L}_{D^{tr}_{\Tau_{j}}})} }(\boldsymbol{x})$ \\
         \bottomrule
    \end{tabularx}
    \caption{High-level overview of the three Deep Meta-Learning categories, i.e., i)~metric-, ii)~model-, and iii)~optimization-based techniques, and their main strengths and weaknesses. Recall that $\Tau_{j}$ is a task, $D^{tr}_{\Tau_{j}}$ the corresponding support set, $k_{\boldsymbol{\theta}}(\boldsymbol{x}, \boldsymbol{x}_{i})$ a kernel function returning the similarity between the two inputs $\boldsymbol{x}$ and $\boldsymbol{x}_{i}$, $y_{i}$ are true labels for known inputs $\boldsymbol{x}_{i}$, $\theta$ are base-learner parameters, and $g_{\boldsymbol{\varphi}}$ is a (learned) optimizer with parameters $\boldsymbol{\varphi}$.}
    \label{tab:categorization}
\end{table}

\begin{table}[phtb]
    \begin{tabularx}{\linewidth}{ccXc}
         Name & RL &Key idea & Bench. \\
         \toprule 
         \textbf{Metric-based}  & & \textbf{Input similarity} & -\\
         \, Siamese networks   & \xmark & Two-input, shared-weight, class identity network & 1, 8  \\
         \, Matching networks & \xmark & Learn input embeddings for cosine-similarity weighted predictions & 1, 2  \\
         \, Prototypical networks & \xmark & Input embeddings for class prototype clustering & 1, 2, 7 \\
         \, Relation networks   & \xmark & Learn input embeddings and similarity metric & 1, 2, 7\\
         \, ARC  & \xmark & LSTM-based input fusion through interleaved glimpses & 1, 2 \\
         \, GNN  & \xmark & Propagate label information to unlabeled inputs in a graph & 1, 2 \\
         \midrule
         \textbf{Model-based} &  & \textbf{Internal and stateful latent task representations} & - \\
         \, Reccurrent ml.   & \checkmark & Deploy Recurrent networks on RL problems & -\\
         \, MANNs  & \xmark & External short-term memory module for fast learning & 1  \\
         \, Meta networks  & \checkmark & Fast reparameterization of base-learner by distinct meta-learner & 1, 2 \\
         \, SNAIL   & \checkmark & Attention mechanism coupled with temporal convolutions & 1, 2 \\
         \, CNP   & \xmark & Condition predictive model on embedded contextual task data & 1, 8 \\
         \, Neural stat.  & \xmark & Similarity between latent task embeddings & 1, 8 \\
         \midrule
         \textbf{Opt.-based}  & & \textbf{Optimize for fast task-specific adaptation} & - \\
         \, LSTM optimizer   & \xmark & RNN proposing weight updates for base-leaner & 6, 8 \\
         \, LSTM ml.  & \checkmark & Embed base-learner parameters in cell state of LSTM & 2\\
         \, RL optimizer   & \xmark & View optimization as RL problem & 4, 6 \\
         \, MAML  & \checkmark & Learn initialization weights $\boldsymbol{\theta}$ for fast adaptation & 1, 2\\
         \, iMAML   & \checkmark & Approx. higher-order gradients, independent of optimization path  & 1, 2\\
         \, Meta-SGD   & \checkmark & Learn both the initialization and updates & 1, 2  \\
         \, Reptile  & \checkmark & Move initialization towards task-specific updated weights & 1, 2 \\
         \, LEO   & \xmark & Optimize in lower-dimensional latent parameter space & 2, 3 \\
         \, Online MAML  & \xmark & Accumulate task data for MAML-like training & 4, 8 \\
         \, LLAMA   & \xmark & Maintain probability distribution over post-update parameters $\boldsymbol{\theta}'_{j}$ & 2 \\
         \, PLATIPUS  & \xmark & Learn a probability distribution over weight initializations $\boldsymbol{\theta}$ & -\\
         \, BMAML  & \checkmark & Learn multiple initializations $\boldsymbol{\Theta}$, jointly optimized by SVGD & 2\\
         \, Diff. solvers   & \xmark & Learn input embeddings for simple base-learners & 1, 2, 3, 4, 5
    \end{tabularx}
    \caption{Overview of the discussed Deep Meta-Learning techniques. The table is partitioned into three sections, i.e., metric-, model-, and optimization-based techniques. All methods in one section adhere to the key idea of its corresponding category, which is mentioned in bold font. The columns RL and Bench show whether the techniques are applicable to reinforcement learning settings and the used benchmarks for testing the performance of the techniques. Note that all techniques are applicable to supervised learning, with the exception of RMLs. The benchmark column displays which benchmarks from \autoref{sec:benchmarks} were used in the paper proposing the technique. The used coding scheme for this column is the following. 1: Omniglot, 2: miniImageNet, 3: tieredImageNet, 4: CIFAR-100, 5: CIFAR-FS, 6: CIFAR-10, 7: CUB, 8: MNIST, ``-": used other evaluation method that are non-standard in Deep Meta-Learning and thus not covered in \autoref{sec:benchmarks}. Used abbreviations: ``opt.": optimization, ``diff.": differentiable, ``bench.": benchmarks.}
    \label{tab:overview}
\end{table}

\section{Metric-based Meta-Learning}\label{sec:metriclearning}

At a high level, the goal of metric-based\index{meta-learning!metric-based} techniques is to acquire---among others---meta-knowledge $\omega$ in the form of a good feature space that can be used for various new tasks. In the context of neural networks, this feature space coincides with the weights $\boldsymbol{\theta}$ of the networks. Then, new tasks can be 
learned by comparing new inputs to example inputs (of which we know the labels) in the meta-learned feature space. The higher the similarity between a new input and an example, the more likely it is that the new input will have the same label as the example input.

Metric-based techniques are a form of meta-learning as they leverage their prior learning experience (meta-learned feature space) to `learn' new tasks more quickly. Here, `learn' is used in a non-standard way since metric-based techniques do not make any network changes when presented with new tasks, as they  rely solely on input comparisons in the already meta-learned feature space. These input comparisons are a form of \textit{non-parametric learning}, i.e., new task information is not absorbed into the network parameters.   

More formally, metric-based learning techniques aim to learn a similarity kernel, or equivalently, \textit{attention mechanism} $k_{\boldsymbol{\theta}}$ (parameterized by $\boldsymbol{\theta}$), that takes two inputs $\boldsymbol{x}_{1}$ and $\boldsymbol{x}_{2}$, and outputs their similarity score. Larger scores indicate larger similarity. Class predictions for new inputs $\boldsymbol{x}$ can then be made by comparing $\boldsymbol{x}$ to example inputs $\boldsymbol{x}_{i}$, of which we know the true labels $y_{i}$. The underlying idea being that the larger the similarity between $\boldsymbol{x}$ and $\boldsymbol{x}_{i}$, the more likely it becomes that $\boldsymbol{x}$ also has label $y_{i}$. 

Given a task $\Tau_{j} = (D^{tr}_{\Tau_{j}}, D^{test}_{\Tau_{j}})$ and an unseen input vector $\boldsymbol{x} \in D^{test}_{\Tau_{j}}$, a probability distribution over classes $Y$ is computed/predicted as a weighted combination of labels from the support set $D^{tr}_{\Tau_{j}}$, using similarity kernel $k_{\boldsymbol{\theta}}$, i.e.,

\begin{align}
    p_{\boldsymbol{\theta}}(Y|\boldsymbol{x}, D^{tr}_{\Tau_{j}}) = \sum_{(\boldsymbol{x}_{i}, y_{i}) \in D^{tr}_{\Tau_{j}}} k_{\boldsymbol{\theta}}(\boldsymbol{x}, \boldsymbol{x}_{i})y_{i}.\label{eq:metricLearning}
\end{align}
Importantly, the labels $y_{i}$ are assumed to be \textit{one-hot encoded}, meaning that they are represented by zero vectors with a `1' on the position of the true class. For example, suppose there are five classes in total, and our example $\boldsymbol{x}_{1}$ has true class 4. Then, the one-hot encoded label is $y_{1} = [0,0,0,1,0]$. Note that
the probability distribution $ p_{\boldsymbol{\theta}}(Y|\boldsymbol{x}, D^{tr}_{\Tau_{j}})$ over classes is a vector of size $|Y|$, in which the $i$-th entry corresponds to the probability that input $\boldsymbol{x}$ has class $Y_{i}$ (given the support set). The predicted class is thus $\hat{y} = \argmax_{i=1,2, \ldots ,|Y|} p_{\boldsymbol{\theta}}(Y|\boldsymbol{x},S)_{i}$, where $p_{\boldsymbol{\theta}}(Y|\boldsymbol{x},S)_{i}$ is the computed probability that input $\boldsymbol{x}$ has class $Y_{i}$.

\begin{figure}[tb]
    \centering
    \includegraphics[scale=0.5]{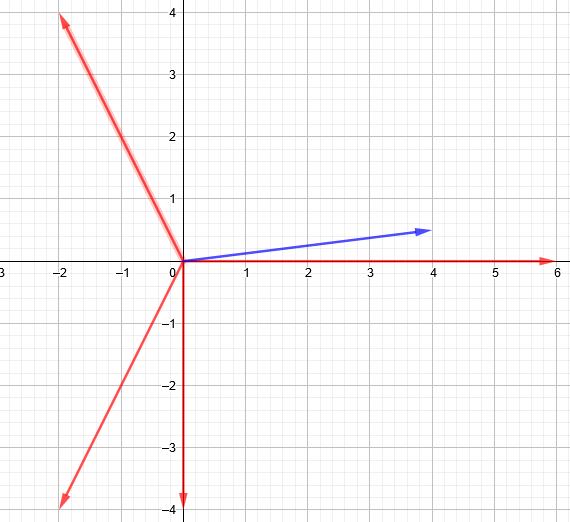}
    \caption{Illustration of our metric-based example. The blue vector represents the new input from the query set, whereas the red vectors are inputs from the support set which can be used to guide our prediction for the new input.}
    \label{fig:metrixample}
\end{figure}

\subsection{Example}
Suppose that we are given a task $\Tau_{j} = (D^{tr}_{\Tau_{j}}, D^{test}_{\Tau_{j}})$. Furthermore, suppose that $D^{tr}_{\Tau_{j}} = \{ ([0,-4], 1), ([-2,-4],2), ([-2,4],3), ([6,0], 4) \}$, where a tuple denotes a pair $(\boldsymbol{x}_{i},y_{i})$. For simplicity, the example will not use an embedding function, which maps example inputs onto an (more informative) embedding space. Our query set only contains one example $D^{test}_{\Tau_{j}} = \{ ([4, 0.5], y) \}$. Then, the goal is to predict the correct label for new input $[4, 0.5]$ using only examples in $D^{tr}_{\Tau_{j}}$. The problem is visualized in \autoref{fig:metrixample}, where red vectors correspond to example inputs from our support set. The blue vector is the new input that needs to be classified. Intuitively, this new input is most similar to the vector $[6,0]$, which means that we expect the label for the new input to be the same as that for $[6,0]$, i.e., $4$. 

Suppose we use a fixed similarity kernel, namely the cosine similarity, i.e.,
$k(\boldsymbol{x}, \boldsymbol{x}_{i}) = \frac{\boldsymbol{x} \cdot \boldsymbol{x}_{i}^{T}}{||\boldsymbol{x}|| \cdot ||\boldsymbol{x}_{i}||}$, where $||\boldsymbol{v}||$ denotes the length of vector $\boldsymbol{v}$, i.e., $||\boldsymbol{v}|| = \sqrt{(\sum_{n}v_{n}^{2})}$. Here, $v_{n}$ denotes the $n$-th element of placeholder vector $\boldsymbol{v}$ (substitute $\boldsymbol{v}$ by $\boldsymbol{x}$ or $\boldsymbol{x}_{i}$). We can now compute the cosine similarity between the new input $[4,0.5]$ and every example input $\boldsymbol{x}_{i}$, as done in \autoref{tab:exampleMetric}, where we used the facts that $||\boldsymbol{x}|| = ||\, [4,0.5] \, || = \sqrt{4^{2}+0.5^{2}} \approx 4.03$, and $\frac{\boldsymbol{x}}{||\boldsymbol{x}||} \approx \frac{[4,0.5]}{4.03} = [0.99,0.12]$. 

From this table and \autoref{eq:metricLearning}, it follows that the predicted probability distribution $p_{\boldsymbol{\theta}}(Y|\boldsymbol{x}, D^{tr}_{\Tau_{j}}) = -0.12y_{1} -0.58y_{2} - 0.37y_{3} + 0.99y_{4} = -0.12 [1,0,0,0] - 0.58 [0,1,0,0] -0.37[0,0,1,0] + 0.99[0,0,0,1] =\allowbreak [-0.12,\allowbreak -0.58,\allowbreak -0.37,\allowbreak 0.99]$. Note that this is not really a probability distribution. That would require normalization such that every element is at least $0$ and the sum of all elements is $1$. For the sake of this example, we do not perform this normalization, as it is clear that class 4 (the class of the most similar example input $[6,0]$) will be predicted. 

\begin{table}[tb]
    \centering
    \begin{tabular}{cccccc}
    \toprule
         & $\boldsymbol{x}_{i}$ & $y_{i}$ & $||\boldsymbol{x}_{i}||$ & $\frac{\boldsymbol{x}_{i}}{||\boldsymbol{x}_{i}||}$ & $\frac{\boldsymbol{x}_{i}}{||\boldsymbol{x}_{i}||} \cdot \frac{\boldsymbol{x}}{||\boldsymbol{x}||}$\\
         \midrule
         & $[0,-4]$ & $[1,0,0,0]$ & $4$ & $[0,-1]$ & $-0.12$\\
         & $[-2,-4]$ & $[0,1,0,0]$ &$4.47$ & $[-0.48,-0.89]$ & $-0.58$\\
         & $[-2,4]$ & $[0,0,1,0]$ & $4.47$ & $[-0.48,0.89]$ & $-0.37$\\
         & $[6,0]$ & $[0,0,0,1]$& $6$ & $[1,0]$& $0.99$ \\
         \bottomrule
    \end{tabular}
    \caption{Example showing pair-wise input comparisons. Numbers were rounded to two decimals.}
    \label{tab:exampleMetric}
\end{table}

One may wonder why such techniques are meta-learners, for we could take any single data set $\mathcal{D}$ and use pair-wise comparisons to compute predictions. 
At the outer-level, metric-based meta-learners are trained on a distribution of different tasks, in order to learn (among others) a good input embedding function. 
This embedding function facilitates inner-level learning, which is achieved through pair-wise comparisons. 
As such, one learns an embedding function across tasks to facilitate task-specific learning, which is equivalent to ``learning to learn", or meta-learning.  

After this introduction to metric-based methods, we will now cover some key metric-based techniques.

\subsection{Siamese Neural Networks}\label{sec:siamese}

\begin{figure}[tb]
    \centering
    \includegraphics[scale=0.6]{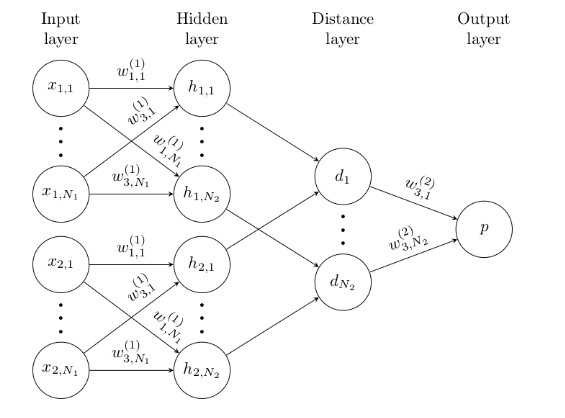}
    \caption{Example of a Siamese neural network. Source: \citet{Koch15}.}
    \label{fig:siamesenets}
\end{figure}

A Siamese neural network\index{siamese neural networks} \citep{Koch15} consists of two neural networks $f_{\boldsymbol{\theta}}$ that share the same weights $\boldsymbol{\theta}$. Siamese neural networks take two inputs $\boldsymbol{x}_{1}, \boldsymbol{x}_{2}$, and compute two hidden states $f_{\boldsymbol{\theta}}(\boldsymbol{x}_{1}), f_{\boldsymbol{\theta}}(\boldsymbol{x}_{2})$, corresponding to the activation patterns in the final hidden layers. These hidden states are fed into a distance layer, which computes a distance vector $\boldsymbol{d} = |f_{\boldsymbol{\theta}}(\boldsymbol{x}_{1}) - f_{\boldsymbol{\theta}}(\boldsymbol{x}_{2})|$, where $d_{i}$ is the absolute distance between the $i$-th elements of $f_{\boldsymbol{\theta}}(\boldsymbol{x}_{1})$ and $f_{\boldsymbol{\theta}}(\boldsymbol{x}_{2})$. From this distance vector, the similarity between $\boldsymbol{x}_{1}, \boldsymbol{x}_{2}$ is computed as $\sigma (\boldsymbol{\alpha}^{T} \boldsymbol{d})$, where $\sigma$ is the sigmoid function (with output range [0,1]), and $\boldsymbol{\alpha}$ is a vector of free weighting parameters, determining the importance of each $d_{i}$. This network structure can be seen in \autoref{fig:siamesenets}. 

\citet{Koch15} applied this technique to few-shot image recognition in two stages. In the first stage, they train the twin network on an \textit{image verification} task, where the goal is to output whether two input images $\boldsymbol{x}_{1}$ and $\boldsymbol{x}_{2}$ have the same class. The network is thus stimulated to learn discriminative features. In the second stage, where the model is confronted with a new task, the network leverages its prior learning experience. That is, given a task $\Tau_{j} = (D^{tr}_{\Tau_{j}}, D^{test}_{\Tau_{j}})$, and previously unseen input $\boldsymbol{x} \in D^{test}_{\Tau_{j}}$, the predicted class $\hat{y}$ is equal to the label $y_{i}$ of the example $(\boldsymbol{x}_{i},y_{i}) \in D^{tr}_{\Tau_{j}}$ which yields the highest similarity score to $\boldsymbol{x}$. 
In contrast to other techniques mentioned further in this section, Siamese neural networks do not directly optimize for good performance across tasks (consisting of support and query sets).
However, they do leverage learned knowledge from the verification task to learn new tasks quickly.

In summary, Siamese neural networks are a simple and elegant approach to perform few-shot learning. However, they are not readily applicable outside the supervised learning setting.

\subsection{Matching Networks}\label{sec:matchingNets}

\begin{figure}[tb]
  \centering
  \includegraphics[scale=0.55]{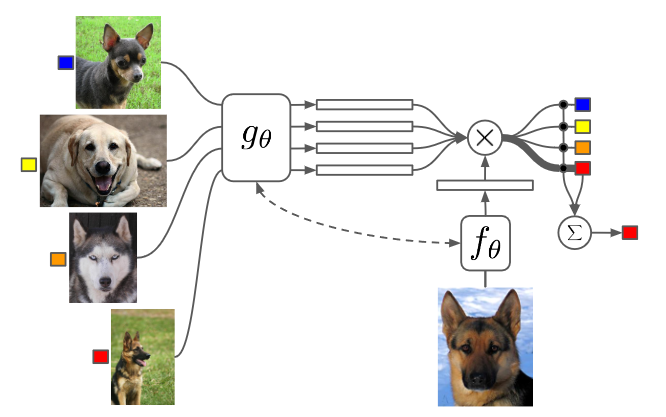}
  \caption{Architecture of matching networks. Source: \citet{vinyals2016matching}.}
  \label{fig:matchingNets}
\end{figure}

Matching networks\index{matching networks} \citep{vinyals2016matching} build upon the idea that underlies Siamese neural networks \citep{Koch15}. That is, they leverage pair-wise comparisons between the given support set $D^{tr}_{\Tau_{j}} = \{ (\boldsymbol{x}_{i}, y_{i}) \}_{i=1}^{m}$ (for a task $\Tau_{j}$), and new inputs $\boldsymbol{x} \in D^{test}_{\Tau_{j}}$ from the query set which we want to classify. However, instead of assigning the class $y_{i}$ of the most similar example input $\boldsymbol{x}_{i}$, matching networks use a weighted combination of \textit{all} example labels $y_{i}$ in the support set, based on the similarity of inputs $\boldsymbol{x}_{i}$ to new input $\boldsymbol{x}$. More specifically, predictions are computed as follows: $\hat{y} = \sum_{i=1}^{m} a(\boldsymbol{x}, \boldsymbol{x}_{i})y_{i}$, where $a$ is a non-parametric (non-trainable) attention mechanism, or similarity kernel. This classification process is shown in \autoref{fig:matchingNets}. In this figure, the input to $f_{\boldsymbol{\theta}}$ has to be classified, using the support set $D^{tr}_{\Tau_{j}}$ (input to $g_{\boldsymbol{\theta}}$).

The  attention that is used consists of a softmax over the cosine similarity $c$ between the input representations, i.e., 
\begin{align}
  a(\boldsymbol{x}, \boldsymbol{x}_{i}) = \frac{e^{c( f_{\boldsymbol{\phi}}(\boldsymbol{x}), g_{\boldsymbol{\varphi}}(\boldsymbol{x}_{i}) )}}{\sum_{j=1}^{m} e^{c( f_{\boldsymbol{\phi}}(\boldsymbol{x}), g_{\boldsymbol{\varphi}}(\boldsymbol{x}_{j}) )}},
\end{align}
where $f_{\boldsymbol{\phi}}$ and $g_{\boldsymbol{\varphi}}$ are neural networks, parameterized by $\boldsymbol{\phi}$ and $\boldsymbol{\varphi}$, that map raw inputs to a (lower-dimensional) latent vector, which corresponds to the output of the final hidden layer of a neural network. As such, the neural networks act as embedding functions. The larger the cosine similarity between the embeddings of $\boldsymbol{x}$ and $\boldsymbol{x}_{i}$, the larger $a(\boldsymbol{x}, \boldsymbol{x}_{i})$, and thus the influence of label $y_{i}$ on the predicted label $\hat{y}$ for input $\boldsymbol{x}$. 

\citet{vinyals2016matching} propose two main choices for the embedding functions. The first is to use a single neural network, granting us $\boldsymbol{\theta} = \boldsymbol{\phi} = \boldsymbol{\varphi}$ and thus $f_{\boldsymbol{\phi}} = g_{\boldsymbol{\varphi}}$. This setup is the default form of matching networks, as shown in \autoref{fig:matchingNets}. The second choice is to make $f_{\boldsymbol{\phi}}$ and $g_{\boldsymbol{\varphi}}$ dependent on the support set $D^{tr}_{\Tau_{j}}$ using Long Short-Term Memory networks (LSTMs). In that case, $f_{\boldsymbol{\phi}}$ is represented by an attention LSTM, and $g_{\boldsymbol{\varphi}}$ by a bidirectional one. This choice for embedding functions is called \textit{Full Context Embeddings} (FCE), and yielded an accuracy improvement of roughly 2\% on miniImageNet compared to the regular matching networks, indicating that task-specific embeddings can aid the classification of new data points from the same distribution. 

Matching networks learn a good feature space across tasks for making pair-wise comparisons between inputs. In contrast to Siamese neural networks \citep{Koch15}, this feature space (given by weights $\boldsymbol{\theta}$) is learned across tasks, instead of on a distinct verification task. 

In summary, matching networks are an elegant and simple approach to metric-based meta-learning. However, these networks are not readily applicable outside of supervised learning settings and suffer from performance degradation when label distributions are biased \citep{vinyals2016matching}.

\subsection{Prototypical Networks}\label{sec:proto}

Just like matching networks \citep{vinyals2016matching}, prototypical networks \citep{snell2017prototypical} base their class predictions on the entire support set $D^{tr}_{\Tau_{j}}$. However, instead of computing the similarity between new inputs and examples in the support set, prototypical networks only compare new inputs to \textit{class prototypes} (centroids), which are single vector representations of classes in some embedding space. Since there are fewer (or equal) class prototypes than the number of examples in the support set, the amount of required pair-wise comparisons decreases, saving computational costs. 

\begin{figure}[thb]
    \centering
    \includegraphics[scale=0.8]{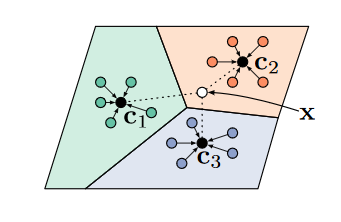}
    \caption{Prototypical networks for the case of few-shot learning. The $\boldsymbol{c}_{k}$ are class prototypes for class $k$ which are computed by averaging the representations of inputs (colored circles) in the support set. Note that the representation space is partitioned into three disjoint areas, where each area corresponds to one class. The class with the closest prototype to the new input $\boldsymbol{x}$ in the query set is then given as prediction.  Source: \citet{snell2017prototypical}.}
    \label{fig:protoNets}
\end{figure}

The underlying idea of class prototypes is that for a task $\Tau_{j}$, there exists an embedding function that maps the support set onto a space where class instances cluster nicely around the corresponding class prototypes \citep{snell2017prototypical}. Then, for a new input $\boldsymbol{x}$, the class of the prototype nearest to that input will be predicted. As such, prototypical networks perform nearest centroid/prototype classification in a meta-learned embedding space. This is visualized in \autoref{fig:protoNets}.  

More formally, given a distance function $d: X \times X \rightarrow [0, +\infty)$ (e.g.\ Euclidean distance) and embedding function $f_{\boldsymbol{\theta}}$, parameterized by $\boldsymbol{\theta}$, prototypical networks compute class probabilities $p_{\boldsymbol{\theta}}(Y | \boldsymbol{x}, D^{tr}_{\Tau_{j}})$ as follows
\begin{align}
    p_{\boldsymbol{\theta}}(y = k | \boldsymbol{x}, D^{tr}_{\Tau_{j}}) = \frac{exp[-d(f_{\theta}(\boldsymbol{x}), \boldsymbol{c}_{k})]}{\sum_{y_{i}} exp[-d(f_{\theta}(\boldsymbol{x}), \boldsymbol{c}_{y_{i}}) ]},
\end{align}
where $\boldsymbol{c}_{k}$ is the prototype/centroid for class $k$ and $y_{i}$ are the classes in the support set $D^{tr}_{\Tau_{j}}$. Here, a class prototype for class $k$ is defined as the average of all vectors $\boldsymbol{x}_{i}$ in the support set such that $y_{i} = k$. Thus, classes with prototypes that are nearer to the new input $\boldsymbol{x}$ obtain larger probability scores.  

\citet{snell2017prototypical} found that the squared Euclidean distance function as $d$ gave rise to the best performance. With that distance function, prototypical networks can be seen as linear models. To see this, note that $-d(f_{\theta}(\boldsymbol{x}), \boldsymbol{c}_{k}) = -|| f_{\theta}(\boldsymbol{x}) - \boldsymbol{c}_{k}||^{2} = - f_{\theta}(\boldsymbol{x})^{T}f_{\theta}(\boldsymbol{x}) + 2\boldsymbol{c}_{k}^{T}f_{\theta}(\boldsymbol{x}) - \boldsymbol{c}_{k}^{T}\boldsymbol{c}_{k}$. The first term does not depend on the class $k$, and does thus not affect the classification decision. The remainder can
be written as $\boldsymbol{w}_{k}^{T}f_{\theta}(\boldsymbol{x}) + \boldsymbol{b}_{k}$, where $\boldsymbol{w}_{k} = 2\boldsymbol{c}_{k}$ and $\boldsymbol{b}_{k} = -\boldsymbol{c}_{k}^{T}\boldsymbol{c}_{k}$. 
Note that this is linear in the output of network $f_\theta$, not linear in the input of the network $\boldsymbol{x}$. 
Also, \citet{snell2017prototypical} show that prototypical networks (coupled with Euclidean distance) are equivalent to matching networks in one-shot learning settings, as every example in the support set will be its prototype. 

In short, prototypical networks save computational costs by reducing the required number of pair-wise comparisons between new inputs and the support set, by adopting the concept of class prototypes. Additionally, prototypical networks were found to outperform matching networks \citep{vinyals2016matching} in 5-way, $k$-shot learning for $k=1,5$ on Omniglot \citep{lake2011one} and miniImageNet \citep{vinyals2016matching}, even though they do not use complex task-specific embedding functions. Despite these advantages, prototypical networks are not readily applicable outside of supervised learning settings.   

\subsection{Relation Networks}\label{sec:relnets}
\begin{figure}[tb]
    \centering
    \includegraphics[scale=0.72]{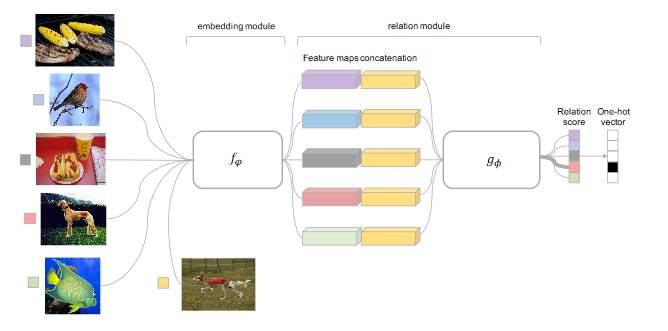}
    \caption{Relation network architecture. First, the embedding network $f_{\boldsymbol{\varphi}}$ embeds all inputs from the support set $D^{tr}_{\Tau_{j}}$ (the five example inputs on the left), and the query input (below the $f_{\boldsymbol{\varphi}}$ block). All support set embeddings $f_{\boldsymbol{\varphi}}(\boldsymbol{x}_{i})$ are then concatenated to the query embedding $f_{\boldsymbol{\varphi}}(\boldsymbol{x})$. These concatenated embeddings are passed into a relation network $g_{\boldsymbol{\phi}}$, which computes a relation score for every pair $(\boldsymbol{x}_{i}, \boldsymbol{x})$. The class of the input $\boldsymbol{x}_{i}$ that yields the largest relation score $g_{\boldsymbol{\phi}}([f_{\boldsymbol{\varphi}}(\boldsymbol{x}), f_{\boldsymbol{\varphi}}(\boldsymbol{x}_{i})])$ is then predicted. Source: \citet{sung2018learning}.}
    \label{fig:relnet}
\end{figure}

In contrast to previously discussed metric-based techniques, Relation networks \citep{sung2018learning} employ a trainable similarity metric, instead of a pre-defined one (e.g.\ cosine similarity as used in  matching networks \citep{vinyals2016matching}). More specifically, matching networks consist of two chained, neural network modules: the \textit{embedding} network/module $f_{\boldsymbol{\varphi}}$ which is responsible for embedding inputs, and the \textit{relation} network $g_{\boldsymbol{\phi}}$ which computes similarity scores between new inputs $\boldsymbol{x}$ and example inputs $\boldsymbol{x}_{i}$ of which we know the labels. A classification decision is then made by picking the class of the example input which yields the largest \textit{relation score} (or similarity). Note that Relation networks thus do not use the idea of class prototypes, and simply compare new inputs $\boldsymbol{x}$ to all example inputs $\boldsymbol{x}_{i}$ in the support set, as done by, e.g., matching networks \citep{vinyals2016matching}.  

More formally, we are given a support set $D^{tr}_{\Tau_{j}}$ with some examples $(\boldsymbol{x}_{i}, y_{i})$, and a new (previously unseen) input $\boldsymbol{x}$. Then, for every combination $(\boldsymbol{x}, \boldsymbol{x}_{i})$, the Relation network produces a \textit{concatenated} embedding $[f_{\boldsymbol{\varphi}}(\boldsymbol{x}), f_{\boldsymbol{\varphi}}(\boldsymbol{x}_{i})]$, which is vector obtained by concatenating the respective embeddings of $\boldsymbol{x}$ and $\boldsymbol{x}_{i}$.
This concatenated embedding is then fed into the \textit{relation} module $g_{\boldsymbol{\phi}}$. Finally, $g_{\boldsymbol{\phi}}$ computes the relation score between $\boldsymbol{x}$ and $\boldsymbol{x}_{i}$ as
\begin{align}
    r_{i} = g_{\boldsymbol{\phi}}([f_{\boldsymbol{\varphi}}(\boldsymbol{x}), f_{\boldsymbol{\varphi}}(\boldsymbol{x}_{i}) ]).
\end{align}
The predicted class is then $\hat{y} = y_{\argmax_{i} r_{i}}$. This entire process is shown in \autoref{fig:relnet}. Remarkably enough, Relation networks use the Mean-Squared Error (MSE) of the relation scores, rather than the more standard cross-entropy loss. The MSE is then propagated backwards through the entire architecture (\autoref{fig:relnet}).

The key advantage of Relation networks is their expressive power, induced by the usage of a trainable similarity function. This expressivity makes this technique very powerful. As a result, it yields better performance than previously discussed techniques that use a fixed similarity metric. 

\subsection{Graph Neural Networks}\label{sec:graph}

Graph neural networks\index{graph neural networks} \citep{garcia2017few} use a more general and flexible approach than previously discussed techniques for $N$-way, $k$-shot classification. 
As such, graph neural networks subsume Siamese \citep{Koch15} and prototypical networks \citep{snell2017prototypical}. The graph neural network approach represents each task $\Tau_{j}$ as a fully-connected graph $G = (V,E)$, where $V$ is a set of nodes/vertices and $E$ a set of edges connecting nodes. In this graph, nodes $\boldsymbol{v}_{i}$ correspond to input embeddings $f_{\boldsymbol{\theta}}(\boldsymbol{x}_{i})$, concatenated with their one-hot encoded labels $y_{i}$, i.e., $\boldsymbol{v}_{i} = [f_{\boldsymbol{\theta}}(\boldsymbol{x}_{i}), y_{i}]$. 
For inputs $\boldsymbol{x}$ from the query set (for which we do not have the labels), a uniform prior over all $N$ possible labels is used: $y = [\frac{1}{N}, \ldots ,\frac{1}{N}]$. 
Thus, each node contains an input and label section. Edges are weighted links that connect these nodes.

The graph neural network then propagates information in the graph using a number of local operators. The underlying idea is that label information can be transmitted from nodes of which we do have the labels, to nodes for which we have to predict labels. Which local operators are used, is out of scope for this paper, and the reader is referred to \citet{garcia2017few} for details. 

By exposing the graph neural network to various tasks $\Tau_{j}$, the propagation mechanism can be altered to improve the flow of label information in such a way that predictions become more accurate. As such, in addition to learning a good input representation function $f_{\boldsymbol{\theta}}$, graph neural networks also learn to propagate label information from labeled examples to unlabeled inputs.

Graph neural networks achieve good performance in few-shot settings \citep{garcia2017few} and are also applicable in semi-supervised and active learning settings.  

\subsection{Attentive Recurrent Comparators (ARCs)}

\begin{figure}[tb]
    \centering
    \includegraphics[scale=0.45]{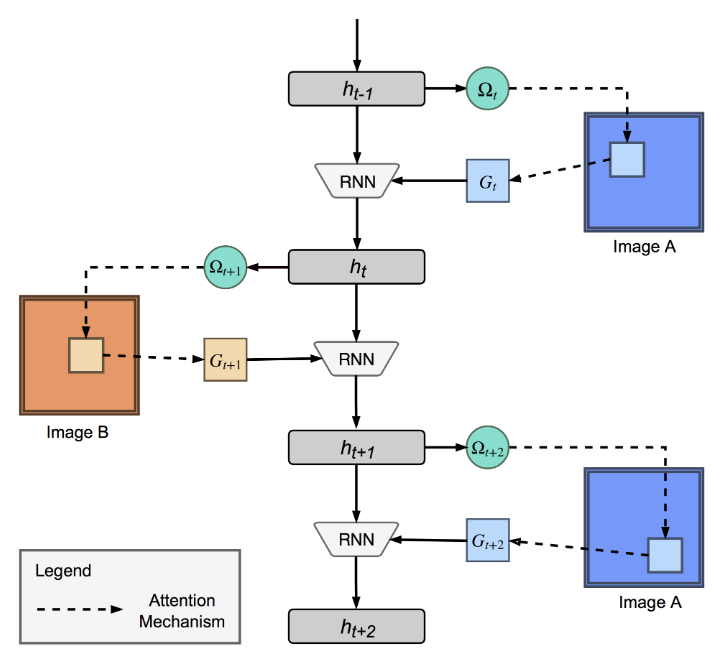}
    \caption{Processing in an attentive recurrent comparator. At every time step, the model takes a glimpse of a part of an image and incorporates this information into the hidden state $h_t$. The final hidden state after taking various glimpses of a pair of images is then used to compute a class similarity score.  Source: \citet{shyam2017attentive}.}
    \label{fig:arc}
\end{figure}

Attentive recurrent comparators\index{attentive recurrent comparators} (ARCs) \citep{shyam2017attentive} differ from previously discussed techniques as they do not compare inputs as a whole, but by parts. This approach is inspired by how humans would make a decision concerning the similarity of objects. That is, we  shift our attention from one object to the other, and move back and forth to take glimpses of different parts of both objects. In this way, information of  two objects is fused from the  beginning, whereas other techniques (e.g., matching networks \citep{vinyals2016matching} and graph neural networks \citep{garcia2017few}) only combine information at the end (after embedding both images) \citep{shyam2017attentive}.  

Given two inputs $\boldsymbol{x}_{i}$ and $\boldsymbol{x}$, we feed them in interleaved fashion repeatedly into a recurrent neural network (controller): $\boldsymbol{x}_{i}, \boldsymbol{x}, \ldots ,\boldsymbol{x}_{i},\boldsymbol{x}$. Thus, the image at time step $t$ is given by $I_{t} = \boldsymbol{x}_{i}$ if $t$ is even else $\boldsymbol{x}$. Then, at each time step $t$, the attention mechanism focuses on a square region of the current image: $G_{t} = attend(I_{t}, \Omega_{t})$, where $\Omega_{t} = W_{g}h_{t-1}$ are attention parameters, which are computed from the previous hidden state $h_{t-1}$. The next hidden state $h_{t+1} = \mbox{RNN}(G_{t}, h_{t-1})$ is given by the glimpse at time t, i.e., $G_{t}$, and the previous hidden state $h_{t-1}$. The entire sequence consists of $g$ glimpses per image. After this sequence is fed into the recurrent neural network (indicated by RNN($\circ$)), the final hidden state $h_{2g}$ is used as combined representation of $\boldsymbol{x}_{i}$ relative to $\boldsymbol{x}$. This process is summarized in \autoref{fig:arc}. Classification decisions can then be made by feeding the combined representations into a classifier. Optionally, the combined representations can be processed by bi-directional LSTMs before passing them to the classifier.

The attention approach is biologically inspired, and biologically plausible. A downside  of attentive recurrent comparators is the higher computational cost, while the performance is often not better than less biologically plausible techniques, such as graph neural networks \citep{garcia2017few}.

\subsection{Metric-based Techniques, in conclusion}
\begin{figure}
    \centering
    \includegraphics[width=\linewidth]{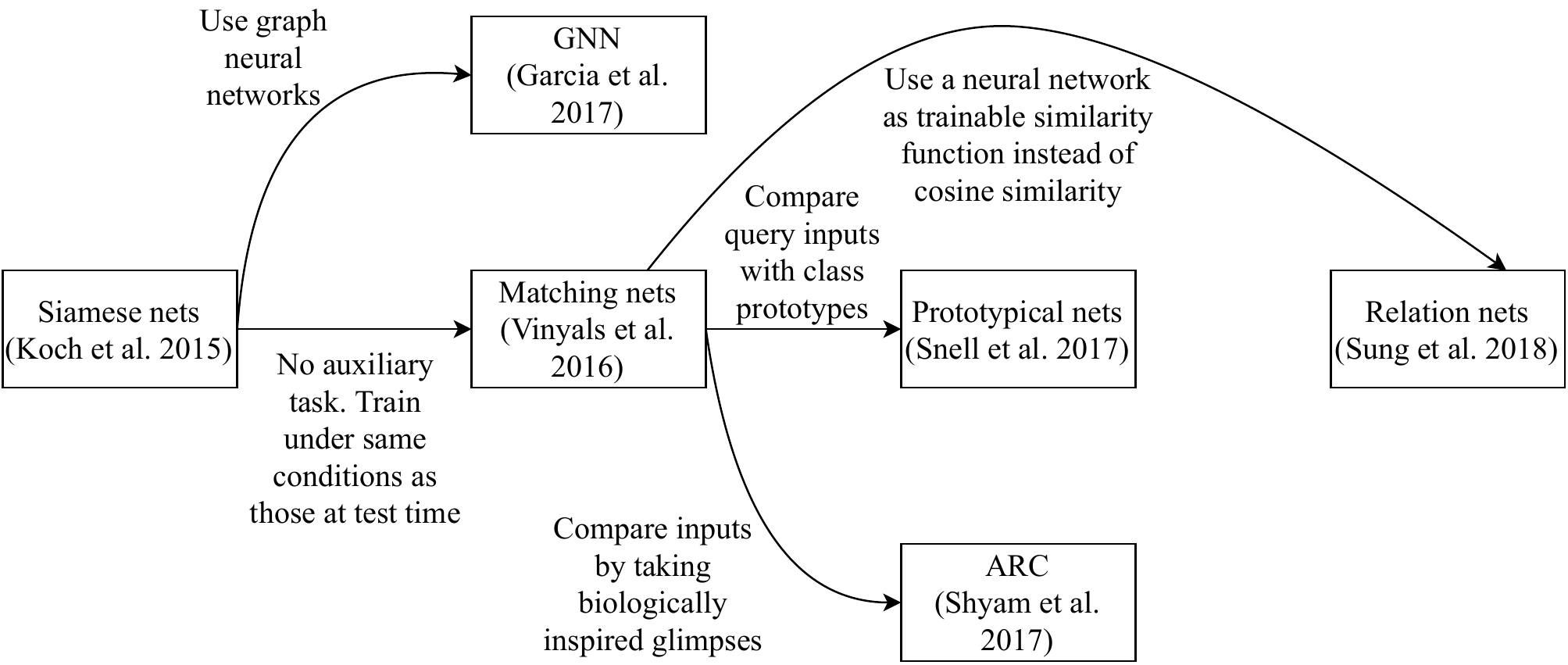}
    \caption{The relationships between the covered metric-based meta-learning techniques.}
    \label{fig:metricbasedrels}
\end{figure}

In this section, we have seen various metric-based\index{meta-learning!metric-based} techniques. The metric-based techniques meta-learn an informative feature space that can be used to compute class predictions based on input similarity scores. 
\autoref{fig:metricbasedrels} shows the relationships between the various metric-based techniques that we have covered. 

As we can see, Siamese networks \citep{Koch15} mark the beginning of metric-based, deep meta-learning techniques in few-shot learning settings.
They are the first to use the idea of predicting classes by comparing inputs from the support and query sets. 
This idea was generalized in graph neural networks (GNNs) \citep{hamiltonips, garcia2017few} where the information flow between support and query inputs is parametric and thus more flexible.
Matching networks \citep{vinyals2016matching} are directly inspired by Siamese networks as they use the same core idea (comparing inputs for making predictions), but directly train in the few-shot setting and use cosine similarity as a similarity function. 
Thus, the auxiliary, binary classification task used by Siamese networks is left out, and matching networks directly train on tasks. 
Prototypical networks \citep{snell2017prototypical} increase the robustness of input comparisons by comparing every query set input with a class prototype instead of individual support set examples. 
This reduces the number of required input comparisons for a single query input to $N$ instead of $k \cdot N$.  
Relation networks \citep{sung2018learning} replace the fixed, pre-defined similarity metrics used in matching and prototypical networks by a neural network, which allows for learning a domain-specific similarity function.
Lastly, attentive recurrent comparators \citep{shyam2017attentive} take a more biologically plausible approach by not comparing entire inputs but by taking multiple interleaved glimpses at various parts of the inputs that are being compared.   

Key advantages of these metric-based techniques are that i)~the underlying idea of similarity-based predictions is conceptually simple, and ii)~they can be  fast at test-time when tasks are small, as the networks do not need to make task-specific adjustments. However, when tasks at meta-test time become more distant from the tasks that were used at meta-train time, metric-learning techniques are unable to absorb new task information into the network weights. Consequently, performance may degrade.   

Furthermore, when tasks become larger, pair-wise comparisons may become prohibitively expensive. Lastly, most metric-based techniques rely on the presence of labeled examples, which make them inapplicable outside of supervised learning settings. 

\section{Model-based Meta-Learning}

A  different approach to Deep Meta-Learning is the model-based\index{meta-learning!model-based} approach. On a high level, model-based techniques rely upon an adaptive, internal state, in contrast to metric-based techniques, which generally use a fixed neural network at test-time.

More specifically, model-based techniques maintain a stateful, internal representation of a task. When presented with a task, a model-based neural network processes the support set in a sequential fashion. At every time step, an input enters and alters the internal state of the model. Thus, the internal state can capture relevant task-specific information, which can be used to make predictions for new inputs.

Because the predictions are based on internal dynamics that are hidden from the outside, model-based techniques are also called \textit{black-boxes}. Information from previous inputs must be remembered, which is why model-based techniques have a memory component, either in- or externally. 

Recall that the mechanics of metric-based techniques were limited to pair-wise input comparisons. This is not the case for model-based techniques, where the human designer has the freedom to choose the internal dynamics of the algorithm. As a result, model-based techniques are not restricted to meta-learning good feature spaces, as they can also learn internal dynamics, used to process and predict input data of tasks.

More formally, given a support set $D^{tr}_{\Tau_{j}}$ corresponding to task $\Tau_{j}$, model-based techniques compute a class probability distribution for a new input $\boldsymbol{x}$ as  
\begin{align}
    p_{\boldsymbol{\theta}}(Y|\boldsymbol{x}, D^{tr}_{\Tau_{j}}) = f_{\boldsymbol{\theta}}(\boldsymbol{x}, D^{tr}_{\Tau_{j}}),
\end{align}
where $f$ represents the black-box neural network model, and $\boldsymbol{\theta}$ its parameters. 

\subsection{Example}
Using the same example as in Section~\ref{sec:metriclearning}, suppose we are given a task support set $D^{tr}_{\Tau_{j}} = \{ ([0,-4], 1), ([-2,-4],2), ([-2,4],3), ([6,0], 4) \}$, where a tuple denotes a pair $(\boldsymbol{x}_{i},y_{i})$. Furthermore, suppose our query set only contains one example $D^{test}_{\Tau_{j}} = \{ ([4, 0.5], 4) \}$. This problem has been visualized in \autoref{fig:metrixample} (in Section~\ref{sec:metriclearning}). For the sake of the example, we do not use an input embedding function: our model will operate on the raw inputs of $D^{tr}_{\Tau_{j}}$ and $D^{test}_{\Tau_{j}}$. As an internal state, our model uses an external \textit{memory matrix} $M \in \mathbb{R}^{4 \times (2+1)}$, with four rows (one for each example in our support set), and three columns (the dimensionality of input vectors, plus one dimension for the correct label). Our model proceeds to process the support set sequentially, reading the examples from $D^{tr}_{\Tau_{j}}$ one by one, and by storing the $i$-th example in the $i$-th row of the memory module. After processing the support set, the memory matrix contains all examples, and as such, serves as internal task representation.

Given the new input $[4,0.5]$, our model could use many different techniques to make a prediction based on this representation. For simplicity, assume that it computes the dot product between $\boldsymbol{x}$, and every memory $M(i)$ (the 2-D vector in the $i$-th row of $M$, ignoring the correct label), and predicts the class of the input which yields the largest dot product. This would produce scores $-2, -10, -6,$ and $24$ for the examples in $D^{tr}_{\Tau_{j}}$ respectively. Since the last example $[6,0]$ yields the largest dot product, we predict that class, i.e., $4$. 

Note that this example could be seen as a metric-based technique where the dot product is used as a similarity function.
However, the reason that this technique is model-based is that it stores the entire task inside a memory module. 
This example was deliberately easy for illustrative purposes. More advanced and successful techniques have been proposed, which we will now cover.

\subsection{Recurrent Meta-Learners}

Recurrent meta-learners \citep{duan2016rl, wang2016learning} are, as the name suggests, meta-learners based on recurrent neural networks. 
The recurrent network serves as dynamic task embedding storage. These recurrent meta-learners were specifically proposed for reinforcement learning problems, hence we will explain them in that setting. 

The recurrence is implemented by e.g.\ an LSTM \citep{wang2016learning} or a GRU \citep{duan2016rl}. The internal dynamics of the chosen Recurrent Neural Network (RNN) allows for fast adaptation to new tasks, while the algorithm used to train the recurrent net gradually accumulates knowledge about the task structure, where each task is modelled as an episode (or set of episodes). 

\begin{figure}[tb]
    \centering
    \includegraphics[width=\linewidth]{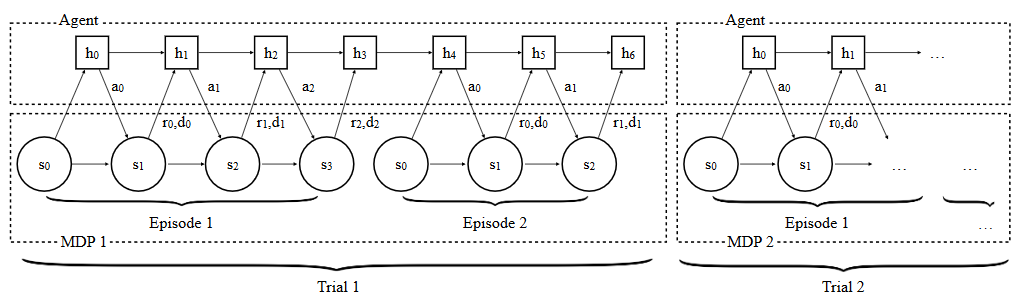}
    \caption{Workflow of recurrent meta-learners in reinforcement learning contexts. As mentioned in \autoref{sec:RL}, $s_{t}, r_{t},$ and $d_{t}$ denote the state, reward, and termination flag at time step $t$. $h_{t}$ refers to the hidden state at time $t$. Source: \citet{duan2016rl}.}
    \label{fig:rlRNN}
\end{figure}

The idea of recurrent meta-learners is quite simple. That is, given a task $\Tau_{j}$, we simply feed the (potentially processed) environment variables $[s_{t+1},a_{t},r_{t},d_{t}]$ (see \autoref{sec:RL}) into an RNN at every time step $t$. Recall that $s,a,r,d$ denote the state, action, reward, and termination flag respectively. At every time step $t$, the RNN outputs an action and a hidden state. Conditioned on its hidden state $h_{t}$, the network outputs an action $a_{t}$. The goal is to maximize the expected reward in each trial. See \autoref{fig:rlRNN} for a visual depiction. From this figure, it also becomes clear why these techniques are model-based. That is, they embed information from previously seen inputs in the hidden state. 

Recurrent meta-learners have shown to perform almost as well as asymptotically optimal algorithms on simple reinforcement learning tasks \citep{wang2016learning, duan2016rl}. However, their performance degrades in more complex settings, where temporal dependencies can span a longer horizon. Making recurrent meta-learners better at such complex tasks is a direction for future research. 

\subsection{Memory-Augmented Neural Networks (MANNs)}\label{sec:MANN}

The key idea of memory-augmented\index{memory-augmented neural networks} neural networks (MANNs) \citep{Santoro16} is to enable neural networks to learn quickly with the help of an \textit{external memory}. The main \textit{controller} (the recurrent neural network interacting with the memory) then gradually accumulates knowledge across tasks, while the external memory allows for quick task-specific adaptation. For this, \citet{Santoro16} used Neural Turing Machines \citep{Graves14}. Here, the controller is parameterized by $\boldsymbol{\theta}$ and acts as the long-term memory of the memory-augmented neural network, while the external memory module is the short-term memory. 

\begin{figure}[!htb]
    \centering
    \includegraphics[width=\linewidth]{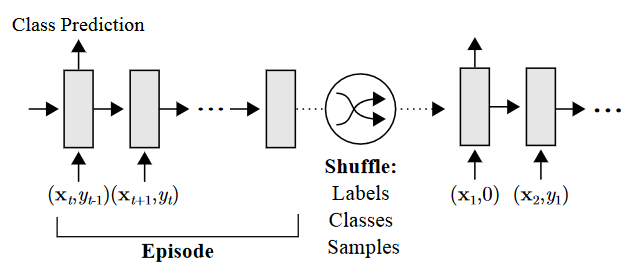}-
    \caption{Workflow of memory-augmented neural networks. Here, an episode corresponds to  a given task $\Tau_j$. After every episode, the order of labels, classes, and samples should be shuffled to minimize dependence on arbitrarily assigned orders. Source: \citet{Santoro16}.}
    \label{fig:flowMANN}
\end{figure}

The workflow of memory-augmented neural networks is displayed in \autoref{fig:flowMANN}. Note that the data from a task is processed as a sequence, i.e., data are fed into the network one by one. The support set is fed into the memory-augmented neural network first. Afterwards, the query set is processed. 
During the meta-train phase, training tasks can be fed into the network in arbitrary order.
At time step $t$, the model receives input $\boldsymbol{x}_{t}$ with the label of the previous input, i.e., $y_{t-1}$. This was done to prevent the network from mapping class labels directly to the output \citep{Santoro16}. 

\begin{figure}[tb!]
    \centering
    \includegraphics[width=\linewidth]{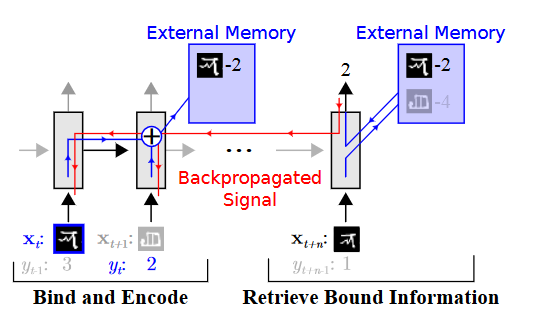}
    \caption{Controller-memory interaction in memory-augmented neural networks. Source: \citet{Santoro16}.}
    \label{fig:memoryMANN}
\end{figure}

The interaction between the controller and memory is visualized in \autoref{fig:memoryMANN}. The idea is that the external memory module, containing representations of previously seen inputs, can be used to make predictions for new inputs. In short, previously obtained knowledge is leveraged to aid the classification of new inputs. Note that neural networks also attempt to do this, however, their prior knowledge is slowly accumulated into the network weights, while an external memory module can directly store such information. 

Given an input $\boldsymbol{x}_{t}$ at time $t$, the controller generates a key $\boldsymbol{k}_{t}$, which can be stored in memory matrix $M$ and can be used to retrieve previous representations from memory matrix $M$. 
When reading from memory, the aim is to produce a linear combination of stored keys in memory matrix $M$, giving greater weight to those which have a larger cosine similarity with the current key $\boldsymbol{k}_{t}$. More specifically, a read vector $\boldsymbol{w}^{r}_{t}$ is created, in which each entry $i$ denotes the cosine similarity between key $\boldsymbol{k}_{t}$ and the memory (from a previous input) stored in row $i$, i.e., $M_{t}(i)$. Then, the representation $\boldsymbol{r}_{t} = \sum_{i}w_{t}^{r}(i)M(i)$ is retrieved, which is simply a linear combination of all keys (i.e., rows) in memory matrix $M$. 

Predictions are made as follows. Given an input $\boldsymbol{x}_{t}$, memory-augmented neural networks use the external memory to compute the corresponding representation $\boldsymbol{r}_{t}$, which could be fed into a softmax layer, resulting in class probabilities. 
Across tasks, memory-augmented neural networks learn a good input embedding function $f_{\boldsymbol{\theta}}$ and classifier weights, which can be exploited when presented with new tasks.

To write input representations to memory, \citet{Santoro16} propose a new mechanism called Least Recently Used Access (LRUA). LRUA either writes to the least, or most recently used memory location. In the former case, it preserves recent memories, and in the latter it updates recently obtained information. The writing mechanism works by keeping track of how often every memory location is accessed in a usage vector $\boldsymbol{w}_{t}^{u}$, which is updated at every time step according to the following update rule: $\boldsymbol{w}_{t}^{u} := \gamma \boldsymbol{w}^{u}_{t-1} + \boldsymbol{w}_{t}^{r} + \boldsymbol{w}_{t}^{w}$, where superscripts $u,w$ and $r$ refer to usage, write and read vectors, respectively. In words, the previous usage vector is decayed (using parameter $\gamma$), while current reads ($\boldsymbol{w}_{t}^{r}$) and writes ($\boldsymbol{w}_{t}^{w}$) are added to the usage.  Let $n$ be the total number of reads to memory, and $\ell u(n)$ ($\ell u$ for `least used') be the $n$-th smallest value in the usage vector $\boldsymbol{w}^{u}_{t}$. Then, the least-used weights are defined as follows: $$\boldsymbol{w}^{\ell u}_{t}(i) = 
\begin{cases} 
  0 & \text{if $w^{u}_{t}(i) > \ell u(n)$} \\
  1 & else
\end{cases}.$$ Then, the write vector $\boldsymbol{w}_{t}^{w}$ is computed as $\boldsymbol{w}^{w}_{t} = \sigma(\alpha) \boldsymbol{w}^{r}_{t-1} + (1 - \sigma(\alpha))\boldsymbol{w}^{\ell u}_{t-1}$, where $\alpha$ is a parameter that interpolates between the two weight vectors. As such, if $\sigma(\alpha) = 1$, we write to the most recently used memory, whereas when $\sigma(\alpha) = 0$, we write to the least recently used memory locations. Finally, writing is performed as follows: $M_{t}(i) := M_{t-1}(i) + w_{t}^{w}(i)\boldsymbol{k}_{t}$, for all $i$. 

In summary, memory-augmented neural networks \citep{Santoro16} combine external memory and a neural network to achieve meta-learning. The interaction between a controller, with long-term memory parameters $\boldsymbol{\theta}$, and memory $M$, may also be interesting for studying human meta-learning \citep{Santoro16}. In contrast to many metric-based techniques, this model-based technique is applicable to both classification and regression problems. A downside of this approach is the architectural complexity.  

\subsection{Meta Networks}\label{sec:metanets}

\begin{figure}[tb]
    \centering
    \includegraphics[scale=0.6]{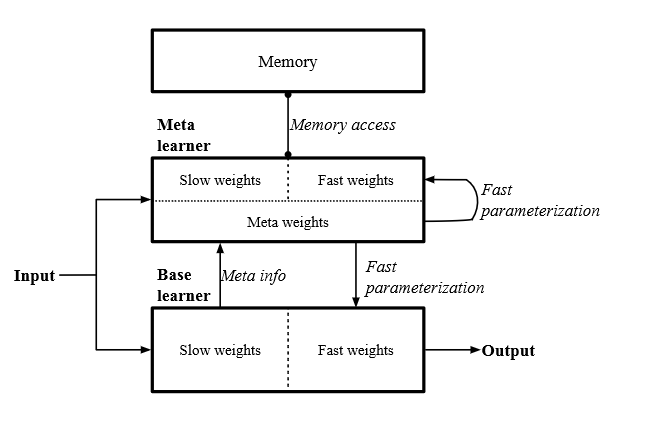}
    \caption{Architecture of a Meta Network. Source: \citet{munkhdalai2017meta}.}
    \label{fig:metaNet}
\end{figure}

Meta networks are divided into two distinct subsystems (consisting of neural networks), i.e., the base- and meta-learner (whereas in memory-augmented neural networks the base- and meta-components are intertwined). The base-learner is responsible for performing tasks, and for providing the meta-learner with meta-information, such as loss gradients. The meta-learner can then compute fast task-specific weights for itself and the base-learner, such that it can perform better on the given task $\Tau_{j} = (D^{tr}_{\Tau_{j}}, D^{test}_{\Tau_{j}})$. This workflow is depicted in \autoref{fig:metaNet}. 

The meta-learner consists of neural networks $u_{\boldsymbol{\phi}}, m_{\boldsymbol{\varphi}}$, and $d_{\boldsymbol{\psi}}$. Network $u_{\boldsymbol{\phi}}$ is used as input representation function. Networks $d_{\boldsymbol{\psi}}$ and $m_{\boldsymbol{\varphi}}$ are used to compute task-specific weights $\boldsymbol{\phi}^{*}$ and example-level fast weights $\boldsymbol{\theta}^{*}$. Lastly, $b_{\boldsymbol{\theta}}$ is the base-learner which performs input predictions. Note that we used the term fast-weights throughout, which refers to task- or input-specific versions of slow (initial) weights. 

In similar fashion to memory-augmented neural networks \citep{Santoro16}, meta networks\index{meta networks} \citep{munkhdalai2017meta} also leverage the idea of an external memory module. However, meta networks use the memory for a different purpose.
The memory stores for each observation $\boldsymbol{x}_i$ in the support set two components, i.e., its representation $\boldsymbol{r}_i$ and the fast weights $\boldsymbol{\theta}_i^*$. 
These are then used to compute a attention-based representation and fast weights for new inputs, respectively. 

\begin{algorithm}
\caption{Meta networks, by \citet{munkhdalai2017meta}}\label{alg:metanets}
\begin{algorithmic}[1]
\State Sample $S = \{ (\boldsymbol{x}_{i},y_{i}) \backsim D^{tr}_{\Tau_{j}} \}_{i=1}^{T}$ from the support set
\For{$(\boldsymbol{x}_{i},y_{i}) \in S$}
    \State $\mathcal{L}_{i} = \mbox{error}(u_{\boldsymbol{\phi}}(\boldsymbol{x}_{i}), y_{i})$ 
\EndFor
\State $\boldsymbol{\phi}^{*} = d_{\boldsymbol{\psi}}(\{ \nabla_{\boldsymbol{\phi}} \mathcal{L}_{i} \}_{i=1}^{T})$
\For{$(\boldsymbol{x}_{i},y_{i}) \in D^{tr}_{\Tau_{j}}$}
    \State $\mathcal{L}_{i} = \mbox{error}(b_{\boldsymbol{\theta}}(\boldsymbol{x}_{i}),y_{i})$
    \State $\boldsymbol{\theta}_{i}^{*} = m_{\boldsymbol{\varphi}}(\nabla_{\boldsymbol{\theta}}\mathcal{L}_{i})$
    \State Store $\boldsymbol{\theta}_{i}^{*}$ in $i$-th position of example-level weight memory $M$
    \State $\boldsymbol{r}_{i} = u_{\boldsymbol{\phi}, \boldsymbol{\phi}^{*}}(\boldsymbol{x}_{i})$
    \State Store $\boldsymbol{r}_{i}$ in $i$-th position of representation memory $R$
\EndFor
\State $\mathcal{L}_{task} = 0$
\For{$(\boldsymbol{x},y) \in D^{test}_{\Tau_{j}}$}
    \State $\boldsymbol{r} = u_{\boldsymbol{\phi}, \boldsymbol{\phi}^{*}}(\boldsymbol{x})$
    \State $\boldsymbol{a} = \mbox{attention}(R, \boldsymbol{r})$ \Comment{$a_{k}$ is the cosine similarity between $\boldsymbol{r}$ and $R(k)$}
    \State $\boldsymbol{\theta}^{*} = \mbox{softmax}(\boldsymbol{a})^{T}M$
    \State $\mathcal{L}_{task} = \mathcal{L}_{task} + \mbox{error}(b_{\boldsymbol{\theta}, \boldsymbol{\theta}^{*}}(\boldsymbol{x}), y)$
\EndFor
\State Update $\Theta = \{ \boldsymbol{\theta}, \boldsymbol{\phi}, \boldsymbol{\psi}, \boldsymbol{\varphi} \}$ using $\nabla_{\Theta} \mathcal{L}_{task}$
\end{algorithmic}
\end{algorithm}

The pseudocode for meta networks is displayed in \autoref{alg:metanets}. First, a sample of the support set is created (line 1), which is used to compute task-specific weights $\boldsymbol{\phi}^{*}$ for the representation network $u_{\boldsymbol{\phi}}$ (lines 2-5). 
Note that  $u_{\boldsymbol{\phi}}$ has two tasks, i) it should compute a representation for inputs $(\boldsymbol{x}_{i}$ (line 10 and 15), and ii) it needs to make predictions for inputs $(\boldsymbol{x}_{i}$, in order to compute a loss (line 3). 
To achieve both goals, a conventional neural network can be used that makes class predictions. The states of the final hidden layer are then used as representations. 
Typically, the cross entropy is calculated over the predictions of representation network $u_{\boldsymbol{\phi}}$. When there are multiple examples per class in the support set, an alternative is to use a contrastive loss function \citep{munkhdalai2017meta}. 

Then, meta networks iterate over every example $(\boldsymbol{x}_{i}, y_{i})$ in the support set $D^{tr}_{\Tau_{j}}$. The base-learner $b_{\boldsymbol{\theta}}$ attempts to make class predictions for these examples, resulting in loss values $\mathcal{L}_{i}$ (line 7-8). The gradients of these losses are used to compute fast weights $\boldsymbol{\theta}^{*}$ for example $i$ (line 8), which are then stored in the $i$-th row of memory matrix $M$ (line 9). Additionally, input representations $\boldsymbol{r}_{i}$ are computed and stored in memory matrix $R$ (lines 10-11). 

Now, meta networks are ready to address the query set $D^{test}_{\Tau_{j}}$. They iterate over every example $(\boldsymbol{x}, y)$, and compute a representation $\boldsymbol{r}$ of it (line 15). This representation is matched against the representations of the support set, which are stored in memory matrix $R$. This matching gives us a similarity vector $\boldsymbol{a}$, where every entry $k$ denotes the similarity between input representation $\boldsymbol{r}$ and the $k$-th row in memory matrix R, i.e., $R(k)$ (line 16). A softmax over this similarity vector is performed to normalize the entries. The resulting vector is used to compute a linear combination of weights that were generated for inputs in the support set (line 17). These weights $\boldsymbol{\theta}^{*}$ are specific for input $\boldsymbol{x}$ in the query set and can be used by the base-learner $b$ to make predictions for that input (line 18). The observed error is added to the task loss. After the entire query set is processed, all involved parameters can be updated using backpropagation (line 20).     

\begin{figure}[tb]
    \centering
    \includegraphics[scale=0.6]{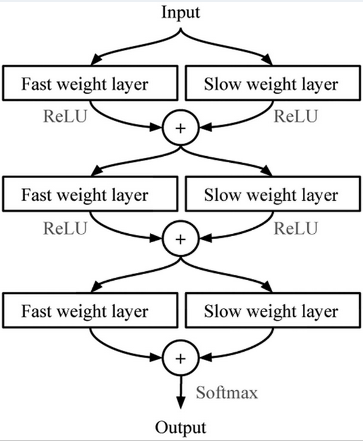}
    \caption{Layer augmentation setup used to combine slow and fast weights. Source: \citet{munkhdalai2017meta}.}
    \label{fig:layeraugmentation}
\end{figure}

Note that some neural networks use both slow- and fast-weights at the same time.  \citet{munkhdalai2017meta} use a so-called augmentation setup for this, as depicted in \autoref{fig:layeraugmentation}.

In short, meta networks rely on a reparameterization of the meta- and base-learner for every task. Despite the flexibility and applicability to both supervised and reinforcement learning settings, the approach is quite complex. It consists of many components, each with its own set of parameters, which can be  a burden on memory usage and computation time. Additionally, finding the correct architecture for all the involved components can be  time-consuming. 

\subsection{Simple Neural Attentive Meta-Learner (SNAIL)}

\begin{figure}[tb]
    \centering
    \includegraphics[scale=0.5]{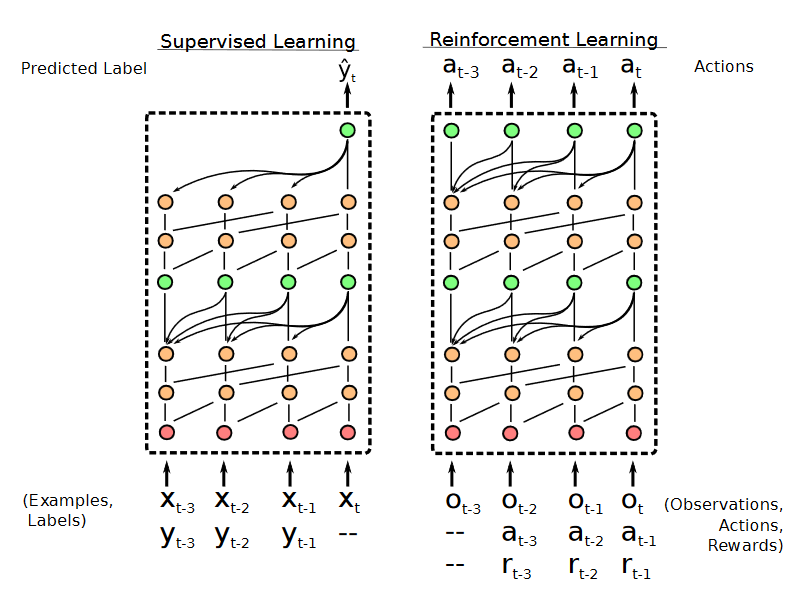}
    \caption{Architecture and workflow of SNAIL for supervised and reinforcement learning settings. The input layer is red. Temporal Convolution blocks are orange; attention blocks are green. Source: \citet{mishra2018simple}.}
    \label{fig:SNAIL}
\end{figure}

Instead of an external memory matrix, SNAIL\index{simple neural attentive meta-learner} \citep{mishra2018simple} relies on a special model architecture to serve as memory. \citet{mishra2018simple} argue that it is not possible to use Recurrent Neural Networks for this, as they have limited memory capacity, and cannot pinpoint specific prior experiences \citep{mishra2018simple}. Hence, SNAIL uses a different architecture, consisting of 1D \textit{temporal convolutions} \citep{oord2016wavenet} and a \textit{soft attention} mechanism \citep{vaswani2017attention}. The temporal convolutions allow for `high bandwidth' memory access, and the attention mechanism allows one to pinpoint specific experiences. \autoref{fig:SNAIL} visualizes the architecture and workflow of SNAIL for supervised learning problems. From this figure, it becomes clear why this technique is model-based. That is, model outputs are based upon the internal state, computed from earlier inputs.

SNAIL consists of three building blocks. The first is the \textit{DenseBlock}, which applies a single 1D convolution to the input, and concatenates (in the feature/horizontal direction) the result. The second is a \textit{TCBlock}, which is simply a series of DenseBlocks with exponentially increasing dilation rate of the temporal convolutions \citep{mishra2018simple}. Note that the dilation is nothing but the temporal distance between two nodes in a network. For example, if we use a dilation of 2, a node at position $p$ in layer $L$ will receive the activation from node $p-2$ from layer $L-1$. The third block is the \textit{AttentionBlock}, which learns to focus on the important parts of prior experience. 

In similar fashion to memory-augmented neural networks \citep{Santoro16} (\autoref{sec:MANN}), SNAIL also processes task data in sequence, as shown in \autoref{fig:SNAIL}. However, the input at time $t$ is accompanied by the label at time $t$, instead of $t-1$ (as was the case for memory-augmented neural networks).  SNAIL learns internal dynamics from  seeing various tasks so that it can make good predictions on the query set, conditioned upon the support set.

A key advantage of SNAIL is that it can be applied to both supervised and reinforcement learning tasks. In addition, it achieves  good performance compared to previously discussed techniques. A downside of SNAIL is that finding the correct architecture of TCBlocks and DenseBlocks can be  time-consuming. 

\subsection{Conditional Neural Processes (CNPs)}

\begin{figure}[tb]
    \centering
    \includegraphics[scale=0.4]{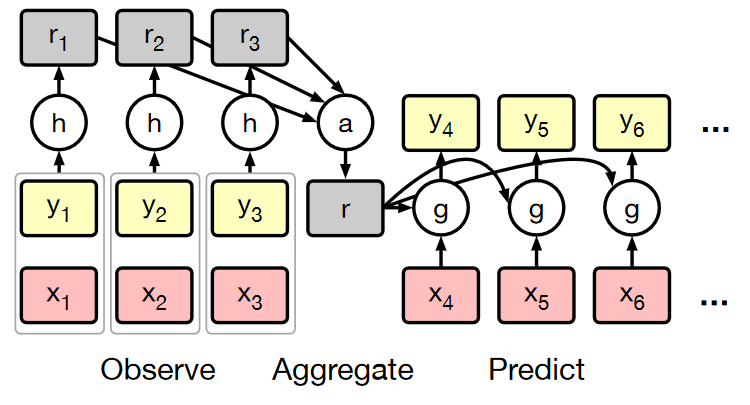}
    \caption{Schematic view of how conditional neural processes work. Here, $h$ denotes a network outputting a representation for a observation, $a$ denotes an aggregation function for these representations, and $g$ denotes a neural network that makes predictions for unlabelled observations, based on the aggregated representation. Source: \citet{garnelo2018conditional}.}
    \label{fig:cnp}
\end{figure}
In contrast to previous techniques, a conditional neural process\index{conditional neural processes} (CNP) \citep{garnelo2018conditional} does not rely on an  external memory module. Instead, it aggregates the  support set into a single aggregated latent representation. The general architecture is shown in \autoref{fig:cnp}. As we can see, the conditional neural process operates in three phases on task $\Tau_{j}$. First, it observes the support set $D^{tr}_{\Tau_{j}}$, including the ground-truth outputs $y_{i}$. Examples $(\boldsymbol{x}_{i},y_{i}) \in D^{tr}_{\Tau_{j}}$ are embedded using a neural network $h_{\boldsymbol{\theta}}$ into representations $\boldsymbol{r}_{i}$. Second, these representations are aggregated using operator $a$ to produce a single representation $\boldsymbol{r}$ of $D^{tr}_{\Tau_{j}}$ (hence it is model-based). Third, a neural network $g_{\boldsymbol{\phi}}$ processes this single representation $\boldsymbol{r}$, new inputs $\boldsymbol{x}$, and produces predictions $\hat{y}$.

Let the entire conditional neural process model be denoted by $Q_{\boldsymbol{\Theta}}$, where $\Theta$ is a set of all involved parameters $\{ \boldsymbol{\theta}, \boldsymbol{\phi} \}$. The training process is different compared to other techniques. Let $\boldsymbol{x}_{\Tau_{j}}$ and $\boldsymbol{y}_{\Tau_{j}}$ denote all inputs and corresponding outputs in $D_{\Tau_{j}}^{tr}$. Then, the first $\ell \backsim U(0, \ldots , k \cdot N -1)$ examples in $D^{tr}_{\Tau_{j}}$ are used as a conditioning set $D^{c}_{\Tau_{j}}$ (effectively splitting the support set in a true training set and a validation set). Given a value of $\ell$, the goal is to maximize the log likelihood (or minimize the negative log likelihood) of the labels $\boldsymbol{y}_{\Tau_{j}}$ in the entire support set $D^{tr}_{\Tau_{j}}$   

\begin{align}
    \mathcal{L}(\boldsymbol{\Theta}) = -\mathbb{E}_{\Tau_{j} \backsim p(\Tau)}\left[ \mathbb{E}_{\ell \backsim U(0, \ldots ,k \cdot N-1)} \left( Q_{\boldsymbol{\Theta}} (\boldsymbol{y}_{\Tau_{j}} | D^{c}_{\Tau_{j}}, \boldsymbol{x}_{\Tau_{j}})  \right) \right].
\end{align} Conditional neural processes are trained by repeatedly sampling various tasks and values of $\ell$, and propagating the observed loss backwards. 

In summary, conditional neural processes use compact representations of previously seen inputs to aid the classification of new observations. Despite its simplicity and elegance, a disadvantage of this technique is that it is often outperformed in few-shot settings by other techniques such as matching networks \citep{vinyals2016matching} (see \autoref{sec:matchingNets}).

\subsection{Neural Statistician}
\begin{figure}[tb]
    \centering
    \includegraphics[scale=0.4]{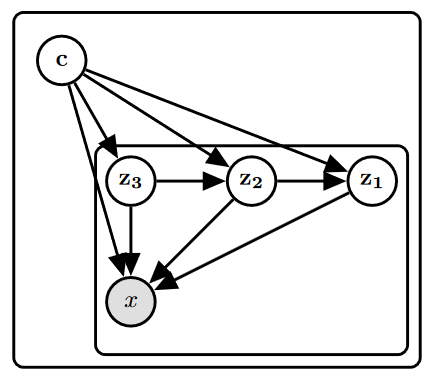}
    \caption{Neural statistician architecture. Edges are neural networks. All incoming inputs to a node are concatenated.}
    \label{fig:neuralStat}
\end{figure}

A neural statistician \citep{edwards2016towards} differs from earlier approaches as it learns to compute \textit{summary statistics}, or \textit{meta-features}, of data sets in an unsupervised manner. These latent embeddings (making the approach model-based) can then later be used for making predictions.  Despite the broad applicability of the model, we discuss it in the context of Deep Meta-Learning. 

A neural statistician performs both \textit{learning} and \textit{inference}. In the learning phase, the model attempts to produce generative models $\hat{P}_{i}$ for every data set $D_{i}$. The key assumption that is made by  \citet{edwards2016towards} is that there exists a generative process $P_{i}$, which conditioned on a latent context vector $\boldsymbol{c}_{i}$, can produce data set $D_{i}$. At inference time, the goal is to infer a (posterior) probability distribution over the context $q(\boldsymbol{c}|D)$. 

The model uses a variational autoencoder, which consists of an encoder and decoder. The encoder is responsible for producing a distribution over latent vectors $\boldsymbol{z}$: $q(\boldsymbol{z}|\boldsymbol{x}; \boldsymbol{\phi})$, where $\boldsymbol{x}$ is an input vector, and $\boldsymbol{\phi}$ are the encoder parameters. The encoded input $\boldsymbol{z}$, which is often of lower dimensionality than the original input $\boldsymbol{x}$, can then be decoded by the decoder $p(\boldsymbol{x}|\boldsymbol{z};\boldsymbol{\theta})$. Here, $\boldsymbol{\theta}$ are the parameters of the decoder. To capture more complex patterns in data sets, the model uses multiple latent layers $\boldsymbol{z}_{1}, \ldots ,\boldsymbol{z}_{L}$, as shown in \autoref{fig:neuralStat}. Given this architecture, the posterior over $c$ and $\boldsymbol{z}_{1},..,\boldsymbol{z}_{L}$ (shorthand $\boldsymbol{z}_{1:L}$) is given by 
\begin{align}
    q(\boldsymbol{c}, \boldsymbol{z}_{1:L}| D; \boldsymbol{\phi}) = q(\boldsymbol{c}|D;\boldsymbol{\phi}) \prod_{\boldsymbol{x} \in D} q(z_{L}| \boldsymbol{x}, \boldsymbol{c};\boldsymbol{\phi})\prod_{i=1}^{L-1} q(\boldsymbol{z}_{i} | \boldsymbol{z}_{i+1},\boldsymbol{x}, \boldsymbol{c};\boldsymbol{\phi}). 
\end{align} The neural statistician is trained to minimize a three-component loss function, consisting of the reconstruction loss (how well it models the data), context loss (how well the inferred context $q(\boldsymbol{c}|D;\boldsymbol{\phi})$ corresponds to the prior $P(\boldsymbol{c})$, and latent loss (how well the inferred latent variables $\boldsymbol{z}_{i}$ are modelled). 

This model can be applied to $N$-way, few-shot learning as follows. Construct $N$ data sets for every of the $N$ classes, such that one data set contains only examples of the same class. Then, the neural statistician is provided with a new input $\boldsymbol{x}$, and has to predict its class. It computes a context posterior $N_{\boldsymbol{x}} = q(\boldsymbol{c}|\boldsymbol{x};\boldsymbol{\phi})$ depending on new input $\boldsymbol{x}$. In similar fashion, context posteriors are computed for all of the data sets $N_{i} = q(\boldsymbol{c}|D_{i};\boldsymbol{\phi})$. Lastly, it assigns the label $i$ such that the difference between $N_{i}$ and $N_{\boldsymbol{x}}$ is minimal. 

In summary, the neural statistician \citep{edwards2016towards} allows for quick learning on new tasks through data set modeling. Additionally, it is applicable to both supervised and unsupervised settings. A downside is that the approach requires many data sets to achieve good performance \citep{edwards2016towards}.

\subsection{Model-based Techniques, in conclusion}
\begin{figure}
    \centering
    \includegraphics[width=\linewidth]{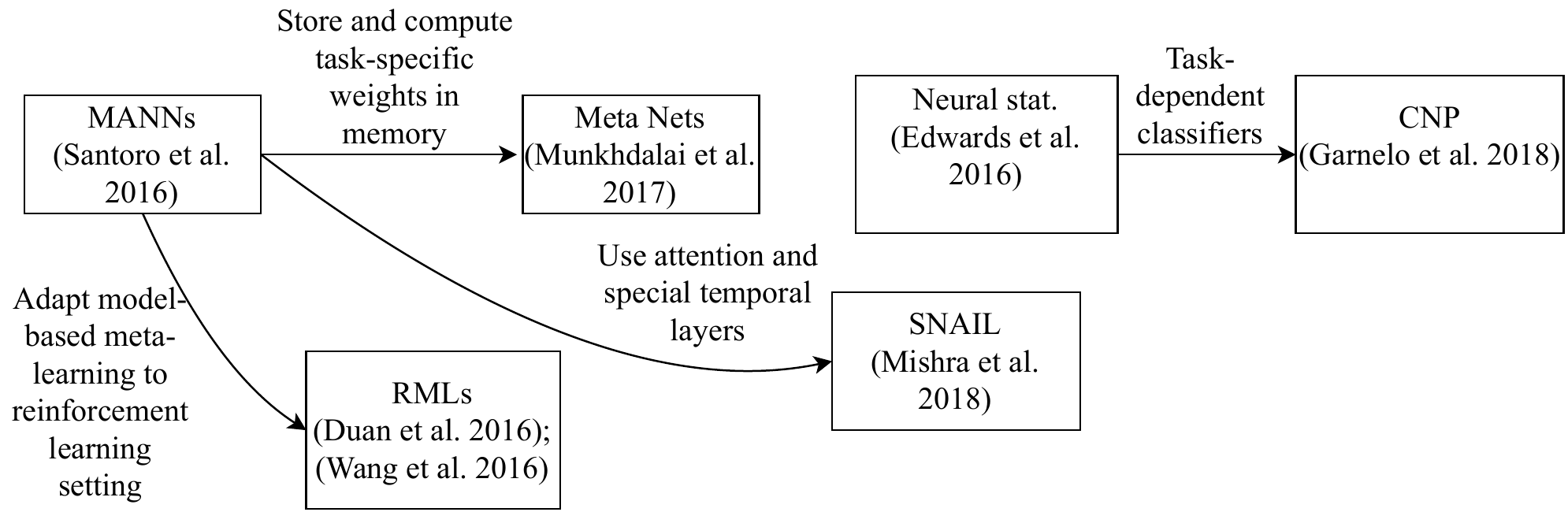}
    \caption{The relationships between the covered model-based meta-learning techniques. The neural statistician and conditional neural process (CNP) form an island in the model-based approaches.}
    \label{fig:modelbasedrels}
\end{figure}

In this section, we have discussed various model-based\index{meta-learning!model-based} techniques. Despite apparent differences, they all build on  the notion of task internalization. That is, tasks are processed and represented in the state of the model-based system. This state can then be used to make predictions. 
\autoref{fig:modelbasedrels} displays the relationships between the covered model-based techniques.

Memory-augmented neural networks (MANNs) \citep{Santoro16} mark the beginning of the deep model-based meta-learning techniques. 
They use the idea of feeding the entire support set in sequential fashion into the model and then making predictions for the query set inputs using the internal state of the model. 
Such a model-based approach, where inputs sequentially enter the model was also taken by recurrent meta-learners \citep{duan2016rl, wang2016learning} in the reinforcement learning setting. 
Meta networks \citep{munkhdalai2017meta} also use a large black-box solution but generate task-specific weights for every task that is encountered.
SNAIL \citep{mishra2018simple} tries to improve the memory capacity and ability to pinpoint memories, which is limited in recurrent neural networks, by using attention mechanisms coupled with special temporal layers.
Lastly, the neural statistician and conditional neural process (CPN) are two techniques that try to learn the meta-features of data sets in an end-to-end fashion.
The neural statistician uses the distance between meta-features to make class predictions, while the conditional neural process conditions classifiers on these features.

Advantages of model-based approaches include the flexibility of the internal dynamics of the systems, and their broader applicability compared to most metric-based techniques. However, model-based techniques are often outperformed by metric-based techniques in supervised settings (e.g. graph neural networks \citep{garcia2017few}; \autoref{sec:graph}), may not perform well when presented with larger data sets \citep{hospedales2020meta}, and generalize less well to more distant tasks than optimization-based techniques \citep{finn2018meta}. We discuss this optimization-based approach next.  

\section{Optimization-based Meta-Learning}

Optimization-based\index{meta-learning!optimization-based} techniques adopt a  different perspective on meta-learning than the previous two approaches. 
They explicitly optimize for fast learning. Most optimization-based techniques do so by approaching meta-learning as a bi-level optimization problem. 
At the inner-level, a base-learner makes task-specific updates using some optimization strategy (such as gradient descent). At the outer-level, the performance across tasks is optimized. 

More formally, given a task $\Tau_{j} = (D^{tr}_{\Tau_{j}}, D^{test}_{\Tau_{j}})$ with new input $\boldsymbol{x} \in D^{test}_{\Tau_{j}}$ and base-learner parameters $\boldsymbol{\theta}$, optimization-based meta-learners return 
\begin{align}
    p(Y|\boldsymbol{x}, D^{tr}_{\Tau_{j}}) = f_{g_{\boldsymbol{\varphi}(\boldsymbol{\theta}, D_{\Tau_{j}}^{tr}, \mathcal{L}_{\Tau_{j}})} }(\boldsymbol{x}),
\end{align}
where $f$ is the base-learner, $g_{\boldsymbol{\varphi}}$ is a (learned) optimizer that makes task-specific updates to the base-learner parameters $\boldsymbol{\theta}$ using the support data $D_{\Tau_{i}}^{tr}$, and loss function $\mathcal{L}_{\Tau_{j}}$. 

\subsection{Example}
\begin{figure}[tb]
    \centering
    \includegraphics[scale=0.6]{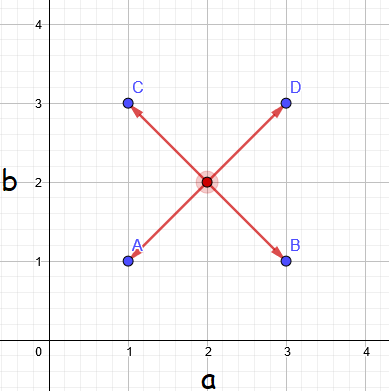}
    \caption{Example of an optimization-based technique, inspired by \citet{Finn17}.}
    \label{fig:optExample}
\end{figure}

Suppose we are faced with a linear regression problem, where every task is associated with a different function $f(x)$. For this example, suppose our model only has two parameters: $a$ and $b$, which together form the function $\hat{f}(x) = ax + b$. Suppose further that our meta-training set consists of four different tasks, i.e., A, B, C, and D. Then, according to the optimization-based view, we wish to find a single set of parameters $\{a, b\}$ from which we can quickly learn the optimal parameters for each of the four tasks, as displayed in \autoref{fig:optExample}. In fact, this is the intuition behind the popular optimization-based technique MAML \citep{Finn17}. 
By exposing our model to various meta-training tasks, we can update the parameters $a$ and $b$ to facilitate quick adaptation. 

We will now discuss the core optimization-based techniques in more detail.

\subsection{LSTM Optimizer}\label{sec:l2l}

Standard gradient update rules have the form
\begin{align}
    \boldsymbol{\theta}_{t+1} := \boldsymbol{\theta}_{t} - \alpha \nabla_{\boldsymbol{\theta}_{t}} \mathcal{L}_{\Tau_{j}}(\boldsymbol{\theta}_{t}),\label{eq:gradientdescent}
\end{align} where $\alpha$ is the learning rate, and $\mathcal{L}_{\Tau_{j}}(\boldsymbol{\theta}_{t})$ is the loss function with respect to task $\Tau_{j}$ and network parameters at time $t$, i.e., $\boldsymbol{\theta}_{t}$. The key idea underlying LSTM optimizers\index{LSTM optimizer} \citep{andrychowicz2016learning} is to replace the update term ($- \alpha \nabla \mathcal{L}_{\Tau_{j}}(\boldsymbol{\theta}_{t})$) by an update proposed by an LSTM $g$ with parameters $\boldsymbol{\varphi}$. Then, the new update becomes 
\begin{align}
    \boldsymbol{\theta}_{t+1} := \boldsymbol{\theta}_{t} + g_{\boldsymbol{\varphi}}(\nabla_{\boldsymbol{\theta}_{t}} \mathcal{L}_{\Tau_{j}}(\boldsymbol{\theta}_{t})).
\end{align}

\begin{figure}[tb]
    \centering
    \includegraphics[width=\linewidth]{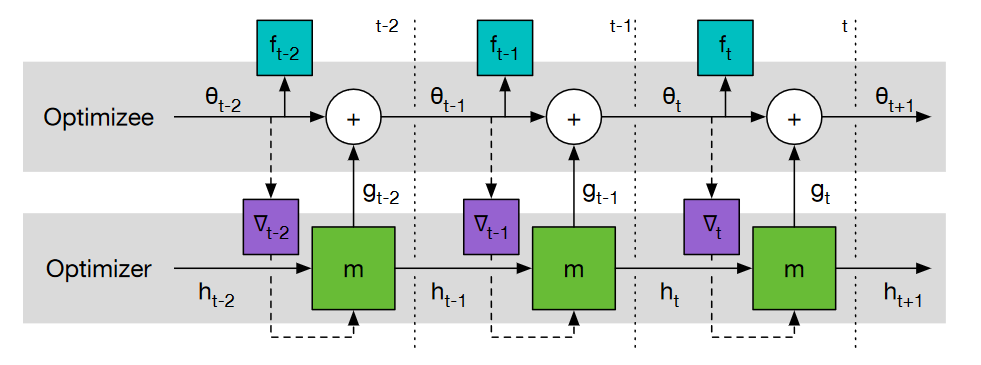}
    \caption{Workflow of the LSTM optimizer. Gradients can only propagate backwards through solid edges. $f_{t}$ denotes the observed loss at time step $t$. Source: \citet{andrychowicz2016learning}.}
    \label{fig:l2l}
\end{figure}
This new update allows the optimization strategy to be tailored to a specific family of tasks. Note that this is meta-learning, i.e., the LSTM learns to learn. As such, this technique basically learns an update policy. 

The loss function used to train an LSTM optimizer is:
\begin{align}
    \mathcal{L}(\boldsymbol{\varphi}) = \mathbb{E}_{\mathcal{L}_{\Tau_{j}}}\left[ \sum_{t=1}^{T}w_{t}\mathcal{L}_{\Tau_{j}}(\boldsymbol{\theta}_{t}) \right],
\end{align} where $T$ is the number of parameter updates that are made, and $w_{t}$ are weights indicating the importance of performance after $t$ steps. 
Note that generally, we are only interested in the final performance after $T$ steps. However, the authors found that the optimization procedure was better guided by equally weighting the performance after each gradient descent step. 
As is often done, second-order derivatives (arising from the dependency between the updated weights and the LSTM optimizer) were ignored due to the computational expenses associated with the computation thereof. This loss function is fully differentiable and thus allows for training an LSTM optimizer (see \autoref{fig:l2l}). To prevent a parameter explosion, the same network is used for every \textit{coordinate}/weight in the base-learner's network, causing the update rule to be the same for every parameter. Of course, the updates depend on their prior values and gradients.

The key advantage of LSTM optimizers is that they can enable faster learning compared to hand-crafted optimizers, also on different data sets than those used to train the optimizer. However, \citet{andrychowicz2016learning} did not apply this technique to few-shot learning. In fact, they did not apply it across tasks at all. Thus, it is unclear whether this technique can perform well in few-shot settings, where few data per class are available for training. Furthermore, the question remains whether it can scale to larger base-learner architectures. 

\subsection{LSTM Meta-Learner}

\begin{figure}[tb]
    \centering
    \includegraphics[width=\linewidth]{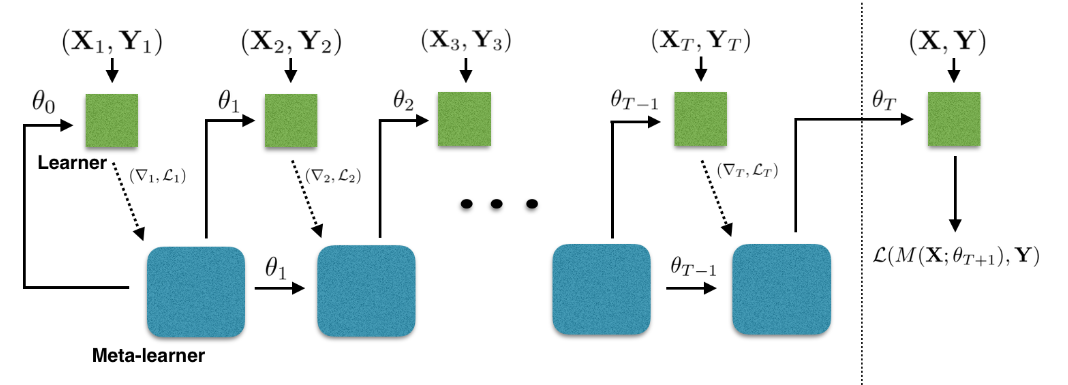}
    \caption{LSTM meta-learner computation graph. Gradients can only propagate backwards through solid edges. The base-learner is denoted as $M$. $(X_{t}, Y_{t})$ are training sets, whereas $(X, Y)$ is the test set. Source: \citet{Ravi2017}.}
    \label{fig:lstmMetaLearner}
\end{figure}

Instead of having an LSTM predict gradient updates, \citet{Ravi2017} embed the weights of the base-learner parameters into the cell state (long-term memory component) of the LSTM, giving rise to LSTM meta-learners. 
As such, the base-learner parameters $\boldsymbol{\theta}$ are literally inside the LSTM memory component (cell state). In this way, cell state updates correspond to base-learner parameter updates. This idea was inspired by the resemblance between the gradient and cell state update rules. Gradient updates often have the form as shown in \autoref{eq:gradientdescent}. The LSTM cell state update rule, in contrast, looks as follows
\begin{align}
    \boldsymbol{c}_{t} := f_{t} \odot \boldsymbol{c}_{t-1} + \alpha_{t} \odot \bar{\boldsymbol{c}}_{t},\label{eq:cellstate}
\end{align} where $f_{t}$ is the forget gate (which determines which information should be forgotten) at time $t$, $\odot$ represents the element-wise product, $\boldsymbol{c}_{t}$ is the cell state at time $t$, and $\bar{\boldsymbol{c}}_{t}$ the candidate cell state for time step $t$, and $\alpha_t$ the learning rate at time step $t$. Note that if $f_{t} = \boldsymbol{1}$ (vector of ones), $\alpha_{t} = \alpha$, $\boldsymbol{c}_{t-1} = \boldsymbol{\theta}_{t-1}$, and $\bar{\boldsymbol{c}}_{t} = - \nabla_{\boldsymbol{\theta}_{t-1}}\mathcal{L}_{\Tau_{t}}(\boldsymbol{\theta}_{t-1})$, this update is equivalent to the one used by gradient-descent. This similarity inspired  \citet{Ravi2017} to use an LSTM as meta-learner that learns to make updates for a base-learner, as shown in \autoref{fig:lstmMetaLearner}. 

More specifically, the cell state of the LSTM is initialized with $c_{0} = \boldsymbol{\theta}_{0}$, which will be adjusted by the LSTM to a good common initialization point across different tasks. 
Then, to update the weights of the base-learner for the next time step $t+1$, the LSTM computes $\boldsymbol{c}_{t+1}$ and sets the weights of the base-learner equal to that. 
There is thus a one-to-one correspondence between $\boldsymbol{c}_{t}$ and $\boldsymbol{\theta}_{t}$. The meta-learner's learning rate $\alpha_{t}$ (see \autoref{eq:cellstate}), is set equal to $\sigma(\boldsymbol{w}_{\alpha} \cdot [\nabla_{\theta_{t-1}} \mathcal{L}_{\Tau_{t}}(\boldsymbol{\theta}_{t-1}), \mathcal{L}_{\Tau_{t}}(\boldsymbol{\theta}_{t}), \theta_{t-1}, \alpha_{t-1}] + \boldsymbol{b}_{\alpha})$, where $\sigma$ is the sigmoid function. Note that the output is a vector, with values between 0 and 1, which denote the the learning rates for the corresponding parameters. Furthermore, $\boldsymbol{w}_{\alpha}$ and $\boldsymbol{b}_{\alpha}$ are trainable parameters that part of the LSTM meta-learner. In words, the learning rate at any time depends on the loss gradients, the loss value, the previous parameters, and the previous learning rate. The forget gate, $f_{t}$, determines what part of the cell state should be forgotten, and is computed in a similar fashion, but with different weights.

To prevent an explosion of meta-learner parameters, weight-sharing is used, in similar fashion to LSTM optimizers proposed by \citet{andrychowicz2016learning} (\autoref{sec:l2l}). This implies that the same update rule is applied to every weight at a given time step. The exact update, however, depends on the history of that specific parameter in terms of the previous learning rate, loss, etc. For simplicity, second-order derivatives were ignored, by assuming the base-learner's loss does not depend on the cell state of the LSTM optimizer. Batch normalization was applied to stabilize and speed up the learning process.  

In short, LSTM optimizers can learn to optimize a base-learner by maintaining a one-to-one correspondence over time between the base-learner's weights and the LSTM cell state. This allows the LSTM to exploit commonalities in the tasks, allowing for quicker optimization. However, there are simpler approaches (e.g.\ MAML \citep{Finn17}) that outperform this technique.

\subsection{Reinforcement Learning Optimizer}

\citet{li2018learning} proposed a framework that casts optimization as a reinforcement learning problem. Optimization can then be performed by existing reinforcement learning techniques\index{reinforcement learning optimizer}. At a high-level, an optimization algorithm $g$ takes as input an initial set of weights $\boldsymbol{\theta}_{0}$ and a task $\Tau_{j}$ with corresponding loss function $\mathcal{L}_{\Tau_{j}}$, and produces a sequence of new weights $\boldsymbol{\theta}_{1},\ldots ,\boldsymbol{\theta}_{T}$, where $\boldsymbol{\theta}_{T}$ is the final solution found. On this sequence of proposed new weights, we can define a loss function $\mathcal{L}$ that captures unwanted properties (e.g.\ slow convergence, oscillations, etc.). 
The goal of learning an optimizer can then be formulated more precisely as follows. We wish to learn an optimal optimizer
\begin{align}
    g^{*} = argmin_{g} \, \mathbb{E}_{\Tau_{j} \backsim p(\Tau), \boldsymbol{\theta}_{0} \backsim p(\boldsymbol{\theta}_{0})}[\mathcal{L}( g(\mathcal{L}_{\Tau_{j}},\boldsymbol{\theta}_{0}))]
\end{align}

The key insight is that the optimization can be formulated as a Partially Observable Markov Decision Process (POMDP). Then, the state corresponds to the current set of weights $\boldsymbol{\theta}_{t}$, the action to the proposed update at time step t, i.e., $\Delta \boldsymbol{\theta}_{t}$, and the policy to the function that computes the update. With this formulation, the optimizer $g$ can be learned by existing reinforcement learning techniques. 
In their paper, they used a recurrent neural network as an optimizer. At each time step, they feed it observation features, which depend on the previous set of weights, loss gradients, and objective functions, and use guided policy search to train it.  

In summary, \citet{li2018learning} made the first step towards general optimization through reinforcement learning optimizers, which were shown able to generalize across network architectures and data sets. However, the base-learner architecture that was used was quite small. The question remains whether this approach can scale to larger architectures.  

\subsection{MAML}\label{sec:maml}
\begin{figure}[!htb]
    \centering
    \includegraphics[scale=0.30]{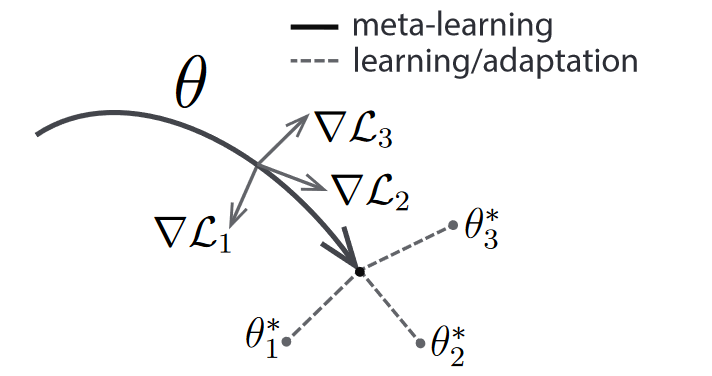}
    \caption{MAML learns an initialization point from which it can perform well on various tasks. Source: \citet{Finn17}.}
    \label{fig:maml}
\end{figure}

Model-agnostic meta-learning\index{model-agnostic meta-learning} (MAML) \citep{Finn17} uses a simple gradient-based inner optimization procedure (e.g.\ stochastic gradient descent), instead of more complex LSTM procedures or procedures based on reinforcement learning. The key idea of MAML is to explicitly optimize for fast adaptation to new tasks by learning a good set of initialization parameters $\boldsymbol{\theta}$. This is shown in \autoref{fig:maml}: from the learned initialization $\boldsymbol{\theta}$, we can quickly move to the best set of parameters for task $\Tau_{j}$, i.e., $\boldsymbol{\theta}^{*}_{j}$ for $j=1,2,3$. The learned initialization can be seen as the \textit{inductive bias} of the model, or simply the set of assumptions (encapsulated in $\boldsymbol{\theta}$) that the model makes concerning the overall task structure. 

More formally, let $\boldsymbol{\theta}$ denote the initial model parameters of a model. The goal is to quickly learn new concepts, which is equivalent to achieving a minimal loss in few gradient update steps. The amount of gradient steps $s$ has to be specified upfront, such that MAML can explicitly optimize for achieving good performance within that number of steps. Suppose we pick only one gradient update step, i.e., $s=1$. Then, given a task $\Tau_{j} = (D^{tr}_{\Tau_{j}}, D^{test}_{\Tau_{j}})$, gradient descent would produce updated parameters (fast weights)
\begin{align}
    \boldsymbol{\theta}'_{j} = \boldsymbol{\theta} - \alpha \nabla_{\boldsymbol{\theta}} \mathcal{L}_{D^{tr}_{\Tau_{j}} }(\boldsymbol{\theta}),
\end{align} specific to task $j$. The \textit{meta-loss} of quick adaptation (using $s=1$ gradient steps) across tasks can then be formulated as 
\begin{align}
    \mathit{ML} := \sum_{\Tau_{j} \backsim p(\Tau)} \mathcal{L}_{D^{test}_{\Tau_{j}}}(\boldsymbol{\theta}'_{j}) = \sum_{\Tau_{j} \backsim p(\Tau)} \mathcal{L}_{D^{test}_{\Tau_{j}}}(\boldsymbol{\theta} - \alpha \nabla_{\boldsymbol{\theta}} \mathcal{L}_{D^{tr}_{\Tau_{j}}}(\boldsymbol{\theta})),
\end{align} where $p(\Tau)$ is a probability distribution over tasks. This expression contains an inner gradient ($\nabla_{\boldsymbol{\theta}} \mathcal{L}_{\Tau_{j}}(\boldsymbol{\theta}_{j})$). As such, by optimizing this meta-loss using gradient-based techniques, we have to compute second-order gradients. One can easily see this in the computation below

\begin{align}
     \nabla_{\boldsymbol{\theta}} \mathit{ML} &=\nabla_{\boldsymbol{\theta}} \sum_{\Tau_{j} \backsim p(\Tau)}  \mathcal{L}_{D^{test}_{\Tau_{j}}}(\boldsymbol{\theta}'_{j}) \nonumber\\
    &= \sum_{\Tau_{j} \backsim p(\Tau)} \nabla_{\boldsymbol{\theta}} \mathcal{L}_{D^{test}_{\Tau_{j}}}(\boldsymbol{\theta}'_{j}) \nonumber \\
    &= \sum_{\Tau_{j} \backsim p(\Tau)}  \mathcal{L}'_{D^{test}_{\Tau_{j}}}(\boldsymbol{\theta}'_{j}) \nabla_{\boldsymbol{\theta}}(\boldsymbol{\theta}'_{j}) \nonumber \\
    &= \sum_{\Tau_{j} \backsim p(\Tau)}  \mathcal{L}'_{D^{test}_{\Tau_{j}}}(\boldsymbol{\theta}_{j}') \nabla_{\boldsymbol{\theta}} ( \boldsymbol{\theta} - \alpha \nabla_{\boldsymbol{\theta}} \mathcal{L}_{D^{tr}_{\Tau_{j}}(\boldsymbol{\theta})}) \nonumber \\
     &= \underbrace{\sum_{\Tau_{j} \backsim p(\Tau)}  \mathcal{L}'_{D^{test}_{\Tau_{j}}}(\boldsymbol{\theta}_{j}')}_\textrm{FOMAML} (\nabla_{\boldsymbol{\theta}} \boldsymbol{\theta} - \alpha \nabla_{\boldsymbol{\theta}}^{2} \mathcal{L}_{D^{tr}_{\Tau_{j}}}(\boldsymbol{\theta})),\label{eq:maml}
\end{align} where we used $\mathcal{L}'_{D^{test}_{\Tau_{j}}}(\boldsymbol{\theta}_{j}')$ to denote the derivative of the loss function with respect to the query set, evaluated at the post-update parameters $\boldsymbol{\theta}_{j}'$. The term $\alpha \nabla_{\boldsymbol{\theta}}^{2} \mathcal{L}_{D^{tr}_{\Tau_{j}}}(\boldsymbol{\theta})$ contains the second-order gradients. The computation thereof is expensive in terms of time and memory costs, especially when the optimization trajectory is large (when using a larger number of gradient updates $s$ per task). \citet{Finn17} experimented with leaving out second-order gradients, by assuming $\nabla_{\boldsymbol{\theta}}\boldsymbol{\theta}'_{j} = I$, giving us First Order MAML (FOMAML, see \autoref{eq:maml}). They found that FOMAML performed reasonably similar to MAML. This means that updating the initialization using only first order gradients $\sum_{\Tau_{j} \backsim p(\Tau)}  \mathcal{L}'_{D^{test}_{\Tau_{j}}}(\boldsymbol{\theta}_{j}')$ is roughly equal to using the full gradient expression of the meta-loss in \autoref{eq:maml}. One can extend the meta-loss to incorporate multiple gradient steps by substituting $\boldsymbol{\theta}_{j}'$ by a multi-step variant.  

MAML is trained as follows. The initialization weights $\boldsymbol{\theta}$ are updated by continuously sampling a batch of $m$ tasks $B = \{\Tau_{j} \backsim p(\Tau)\}_{i=1}^{m}$. Then, for every task $\Tau_{j} \in B$, an \textit{inner update} is performed to obtain $\boldsymbol{\theta}_{j}'$, in turn granting an observed loss $\mathcal{L}_{D^{test}_{\Tau_{j}}}(\boldsymbol{\theta}_{j}')$. These losses across a batch of tasks are used in the \textit{outer update}

\begin{align}
    \boldsymbol{\theta} := \boldsymbol{\theta} - \beta \nabla_{\boldsymbol{\theta}} \sum_{\Tau_{j} \in B} \mathcal{L}_{D^{test}_{\Tau_{j}}}(\boldsymbol{\theta}_{j}').
\end{align} 

The complete training procedure of MAML is displayed in \autoref{alg:maml}. At test-time, when presented with a new task $\Tau_{j}$, the model is initialized with $\boldsymbol{\theta}$, and performs a number of gradient updates on the task data. Note that the algorithm for FOMAML is  equivalent to \autoref{alg:maml}, except for the fact that the update on line 8 is done differently. That is, FOMAML updates the initialization with the rule $\boldsymbol{\theta} = \boldsymbol{\theta} - \beta \sum_{\Tau_{j} \backsim p(\Tau)}  \mathcal{L}'_{D^{test}_{\Tau_{j}}}(\boldsymbol{\theta}_{j}')$.

\begin{algorithm}
\caption{One-step MAML for supervised learning, by \citet{Finn17}}\label{alg:maml}
\begin{algorithmic}[1]
\State Randomly initialize $\boldsymbol{\theta}$
\While{not done}
\State Sample batch of $J$ tasks $B = \Tau_{1},\ldots ,\Tau_{J} \backsim p(\Tau)$
\For{$\Tau_{j} = (D^{tr}_{\Tau_{j}}, D^{test}_{\Tau_{j}}) \in B$}
\State Compute $\nabla_{\boldsymbol{\theta}} \mathcal{L}_{D^{tr}_{\Tau_{j}}}(\boldsymbol{\theta})$
\State Compute $\boldsymbol{\theta}_{j}' = \boldsymbol{\theta} - \alpha \nabla_{\boldsymbol{\theta}} \mathcal{L}_{D^{tr}_{\Tau_{j}}}(\boldsymbol{\theta})$
\EndFor
\State Update $\boldsymbol{\theta} = \boldsymbol{\theta} - \beta \nabla_{\boldsymbol{\theta}}\sum_{\Tau_{j} \in B} \mathcal{L}_{D^{test}_{\Tau_{j}}}(\boldsymbol{\theta}_{j}')$
\EndWhile
\end{algorithmic}
\end{algorithm} 

\citet{antoniou2018how}, in response to MAML, proposed many technical improvements that can improve training stability, performance, and generalization ability. Improvements include i) updating the initialization $\boldsymbol{\theta}$ after every inner update step (instead of after all steps are done) to increase gradient propagation, ii) using second-order gradients only after 50 epochs to increase the training speed, iii) learning layer-wise learning rates to improve flexibility, iv) annealing the meta-learning rate $\beta$ over time, and v) some Batch Normalization tweaks (keep running statistics instead of batch-specific ones, and using per-step biases). 

MAML has obtained great attention within the field of Deep Meta-Learning, perhaps due to its i)~simplicity (only requires two hyperparameters), ii)~general applicability, and iii)~strong performance. A downside of MAML, as mentioned above, is that it can be quite expensive in terms of running time and memory to optimize a base-learner for every task and compute higher-order derivatives from the optimization trajectories.

\subsection{iMAML}

Instead of ignoring higher-order derivatives (as done by FOMAML), which potentially decreases the performance compared to regular MAML, iMAML \citep{rajeswaran2019meta} approximates these derivatives in a way that is less memory-consuming. 

Let $\mathcal{A}$ denote an inner optimization algorithm (e.g., stochastic gradient descent), which takes a support set $D^{tr}_{\Tau_{j}}$ corresponding to task $\Tau_{j}$ and initial model weights $\boldsymbol{\theta}$, and produces new weights $\boldsymbol{\theta}'_{j} = \mathcal{A}(\boldsymbol{\theta}, D^{tr}_{\Tau_{j}})$. MAML has to compute the derivative 

\begin{align}
    \nabla_{\boldsymbol{\theta}} \mathcal{L}_{D^{test}_{\Tau_{j}}} (\boldsymbol{\theta}'_{j}) =  \mathcal{L}_{D^{test}_{\Tau_{j}}}'(\boldsymbol{\theta}'_{j})\nabla_{\boldsymbol{\theta}}(\boldsymbol{\theta}'_{j}) ,\label{eq:implicit1}
\end{align} where $D^{test}_{\Tau_{j}}$ is the query set corresponding to task $\Tau_{j}$. This equation is a simple result of applying the chain rule. Importantly, note that $\nabla_{\boldsymbol{\theta}}(\boldsymbol{\theta}_{j}')$ differentiates through $\mathcal{A}(\boldsymbol{\theta}, D^{tr}_{\Tau_{j}})$, while $ \mathcal{L}_{D^{test}_{\Tau_{j}}}'(\boldsymbol{\theta}'_{j})$ does not, as it represents the gradient of the loss function evaluated at $\boldsymbol{\theta}'_{j}$. \citet{rajeswaran2019meta} make use of the following lemma. 

\textit{If $(\boldsymbol{I} + \frac{1}{\lambda} \nabla^{2}_{\boldsymbol{\theta}}\mathcal{L}_{D^{tr}_{\Tau_{j}}}(\boldsymbol{\theta}'_{j}))$ is invertible (i.e., $(\boldsymbol{I} + \frac{1}{\lambda} \nabla^{2}_{\boldsymbol{\theta}}\mathcal{L}_{D^{tr}_{\Tau_{j}}}(\boldsymbol{\theta}'_{j}))^{-1}$ exists), then}

\begin{align}
    \nabla_{\boldsymbol{\theta}}(\boldsymbol{\theta}_{j}') = \left( \boldsymbol{I} + \frac{1}{\lambda} \nabla^{2}_{\boldsymbol{\theta}}\mathcal{L}_{D^{tr}_{\Tau_{j}}}(\boldsymbol{\theta}'_{j}) \right)^{-1}. \label{eq:implicit2}
\end{align} Here, $\lambda$ is a regularization parameter. The reason for this is discussed below.

Combining \autoref{eq:implicit1} and \autoref{eq:implicit2}, we have that 
\begin{align}
    \nabla_{\boldsymbol{\theta}} \mathcal{L}_{D^{test}_{\Tau_{j}}} (\boldsymbol{\theta}'_{j}) = \mathcal{L}'_{D^{test}_{\Tau_{j}}}(\boldsymbol{\theta}'_{j}) \left( \boldsymbol{I} + \frac{1}{\lambda} \nabla^{2}_{\boldsymbol{\theta}}\mathcal{L}_{D^{tr}_{\Tau_{j}}}(\boldsymbol{\theta}'_{j}) \right)^{-1}  .
\end{align}

The idea is to obtain an approximate gradient vector $\boldsymbol{g}_{j}$ that is close to this expression, i.e., we want the difference to be small
\begin{align}
    \boldsymbol{g}_{j} -  \mathcal{L}'_{D^{test}_{\Tau_{j}}}(\boldsymbol{\theta}'_{j}) \left( \boldsymbol{I} + \frac{1}{\lambda} \nabla^{2}_{\boldsymbol{\theta}}\mathcal{L}_{D^{tr}_{\Tau_{j}}}(\boldsymbol{\theta}'_{j}) \right)^{-1}   = \boldsymbol{\epsilon},
\end{align} for some small tolerance vector $\boldsymbol{\epsilon}$. If we multiply both sides by the inverse of the inverse factor, i.e., $\left( \boldsymbol{I} + \frac{1}{\lambda} \nabla^{2}_{\boldsymbol{\theta}}\mathcal{L}_{D^{tr}_{\Tau_{j}}}(\boldsymbol{\theta}'_{j}) \right)$, we get 
\begin{align}
    \boldsymbol{g}_{j}^{T} \left( \boldsymbol{I} + \frac{1}{\lambda} \nabla^{2}_{\boldsymbol{\theta}}\mathcal{L}_{D^{tr}_{\Tau_{j}}}(\boldsymbol{\theta}'_{j}) \right) \boldsymbol{g}_{j} - \boldsymbol{g}_{j}^{T} \mathcal{L}'_{D^{test}_{\Tau_{j}}}(\boldsymbol{\theta}'_{j}) = \boldsymbol{\epsilon}',
\end{align} where $\boldsymbol{\epsilon}'$ absorbed the multiplication factor. We wish to minimize this expression for $\boldsymbol{g}_{j}$, and that can be performed using optimization techniques such as the conjugate gradient algorithm \citep{rajeswaran2019meta}. This algorithm does not need to store Hessian matrices, which decreases the memory cost significantly. In turn, this allows iMAML to work with more inner gradient update steps. Note, however, that one needs to perform explicit regularization in that case to avoid overfitting. The conventional MAML did not require this, as it uses only a few number of gradient steps (equivalent to an early stopping mechanism). 

At each inner loop step, iMAML computes the meta-gradient $\boldsymbol{g}_{j}$. After processing a batch of tasks, these gradients are averaged and used to update the initialization $\boldsymbol{\theta}$. Since it does not differentiate through the optimization process, we are free to use any other (non-differentiable) inner-optimizer.

In summary, iMAML reduces memory costs significantly as it need not differentiate through the optimization trajectory, also allowing for greater flexibility in the choice of inner optimizer. Additionally, it can account for larger optimization paths. The computational costs stay roughly the same compared to MAML \citep{Finn17}. Future work could investigate more inner optimization procedures \citep{rajeswaran2019meta}.

\subsection{Meta-SGD}\label{sec:metasgd}

\begin{figure}[tb]
    \centering
    \includegraphics[width=\linewidth]{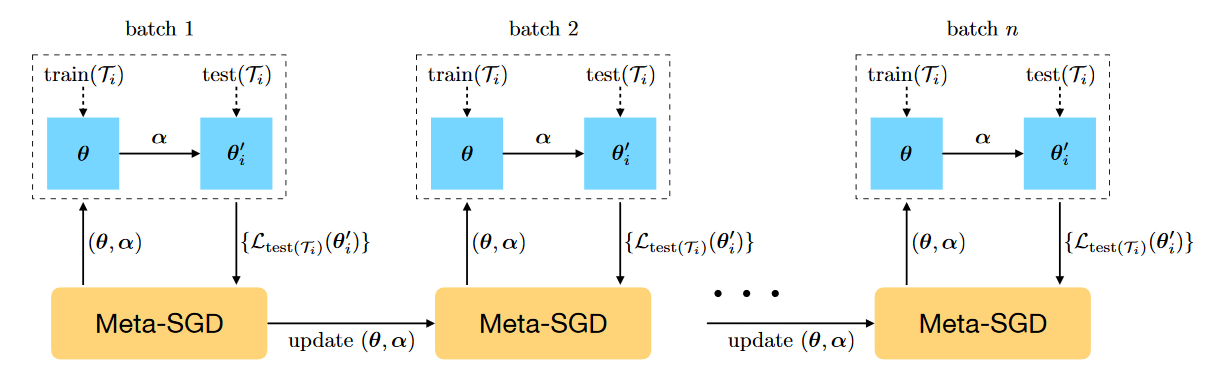}
    \caption{Meta-SGD learning process. Source: \citet{li2017metasgd}.}
    \label{fig:metaSGD}
\end{figure}

Meta-SGD \citep{li2017metasgd}, or meta-stochastic gradient descent, is similar to MAML \citep{Finn17} (\autoref{sec:maml}). However, on top of learning an initialization, Meta-SGD also learns learning rates for every model parameter in $\boldsymbol{\theta}$, building on the insight that the optimizer can be seen as a trainable entity. 

The standard SGD update rule is given in \autoref{eq:gradientdescent}. The meta-SGD optimizer uses a more general update, namely
\begin{align}
    \boldsymbol{\theta}_{j}' \gets \boldsymbol{\theta} - \boldsymbol{\alpha} \odot \nabla_{\boldsymbol{\theta}} \mathcal{L}_{D^{tr}_{\Tau_{j}}}(\boldsymbol{\theta}),
\end{align} where $\odot$ is the element-wise product. Note that this means that alpha (learning rate) is now a vector---hence the bold font--- instead of scalar, which allows for greater flexibility in the sense that each parameter has its own learning rate. The goal is to learn the initialization $\boldsymbol{\theta}$, and learning rate vector $\boldsymbol{\alpha}$, such that the generalization ability is as large as possible. More mathematically precise, the learning objective is 
\begin{align}
    min_{\boldsymbol{\alpha}, \boldsymbol{\theta}} \mathbb{E}_{\Tau_{j} \backsim p(\Tau)} [\mathcal{L}_{D^{test}_{\Tau_{j}}}(\boldsymbol{\theta}_{j}')] = \mathbb{E}_{\Tau_{j} \backsim p(\Tau)} [\mathcal{L}_{D^{test}_{\Tau_{j}}}( \boldsymbol{\theta} - \boldsymbol{\alpha} \odot \nabla_{\boldsymbol{\theta}} \mathcal{L}_{D_{\Tau_{j}}^{tr}}(\boldsymbol{\theta}) )],
\end{align} where we used a simple substitution for $\boldsymbol{\theta}_{j}'$. $\mathcal{L}_{D_{\Tau_{j}}^{tr}}$ and $\mathcal{L}_{D_{\Tau_{j}}^{test}}$ are the losses computed on the support and query set respectively. Note that this formulation stimulates generalization ability (as it includes the query set loss $\mathcal{L}_{D^{test}_{\Tau_{j}}}$, which can be observed during the meta-training phase). The learning process is visualized in \autoref{fig:metaSGD}. Note that the meta-SGD optimizer is trained to maximize generalization ability after only one update step. Since this learning objective has a fully differentiable loss function, the meta-SGD optimizer itself can be trained using standard SGD.

In summary, Meta-SGD is more expressive than MAML as it does not only learn an initialization but also learning rates per parameter. This, however, does come at the cost of an increased number of hyperparameters. 

\subsection{Reptile}

Reptile\index{reptile} \citep{nichol2018reptile} is another optimization-based technique that, like MAML \citep{Finn17}, solely attempts to find a good set of initialization parameters $\boldsymbol{\theta}$. The way in which Reptile attempts to find this initialization is quite different from MAML. It repeatedly samples a task, trains on the task, and moves the model weights towards the trained weights \citep{nichol2018reptile}. \autoref{alg:reptile} displays the pseudocode describing this simple process.

\begin{algorithm}
\caption{Reptile, by \citet{nichol2018reptile}}\label{alg:reptile}
\begin{algorithmic}[1]
\State Initialize $\boldsymbol{\theta}$
\For{$i=1,2,\ldots$}
\State Sample task $\Tau_{j} = (D^{tr}_{\Tau_{j}}, D^{test}_{\Tau_{j}})$ and corresponding loss function $\mathcal{L}_{\Tau_{j}}$
\State $\boldsymbol{\theta}'_{j} = SGD(\mathcal{L}_{D^{tr}_{\Tau_{j}}}, \boldsymbol{\theta}, k)$  \Comment{Perform $k$ gradient update steps to get $\boldsymbol{\theta}_{j}'$ }
\State $\boldsymbol{\theta} := \boldsymbol{\theta} + \epsilon(\boldsymbol{\theta}'_{j} - \boldsymbol{\theta})$ \Comment{Move initialization point $\boldsymbol{\theta}$ towards $\boldsymbol{\theta}_{j}'$}
\EndFor
\end{algorithmic}
\end{algorithm} 

\citet{nichol2018reptile} note that it is possible to treat $( \boldsymbol{\theta} - \boldsymbol{\theta}_{j}')/\alpha$ as gradients, where $\alpha$ is the learning rate of the inner stochastic gradient descent optimizer (line 4 in the pseudocode), and to feed that into a meta-optimizer (e.g.\ Adam). Moreover, instead of sampling one task at a time, one could sample a batch of $n$ tasks, and move the initialization $\boldsymbol{\theta}$ towards the average update direction $\bar{\boldsymbol{\theta}} = \frac{1}{n}\sum_{j=1}^{n}(\boldsymbol{\theta}'_{j} - \boldsymbol{\theta})$, granting the update rule $\boldsymbol{\theta} := \boldsymbol{\theta} + \epsilon \bar{\boldsymbol{\theta}}$. 

The intuition behind Reptile is that updating the initialization weights towards updated parameters will grant a good inductive bias for tasks from the same family. By performing Taylor expansions of the gradients of Reptile and MAML (both first-order and second-order), \citet{nichol2018reptile} show that the expected gradients differ in their direction. They argue, however, that in practice, the gradients of Reptile will also bring the model towards a point minimizing the expected loss over tasks. 

A mathematical argument as to why Reptile works goes as follows. Let $\boldsymbol{\theta}$ denote the initial parameters, and $\boldsymbol{\theta}^{*}_{j}$ the optimal set of weights for task $\Tau_{j}$. Lastly, let $d$ be the Euclidean distance function. Then, the goal is to minimize the distance between the initialization point $\boldsymbol{\theta}$ and the optimal point $\boldsymbol{\theta}^{*}_{j}$, i.e.,
\begin{align}
    min_{\boldsymbol{\theta}} \, \mathbb{E}_{\Tau_{j} \backsim p(\Tau)}[ \frac{1}{2}d(\boldsymbol{\theta}, \boldsymbol{\theta}^{*}_{j})^{2}].\label{eq:distanceObj}
\end{align}

\begin{figure}[htb]
    \centering
    \includegraphics[scale=0.3]{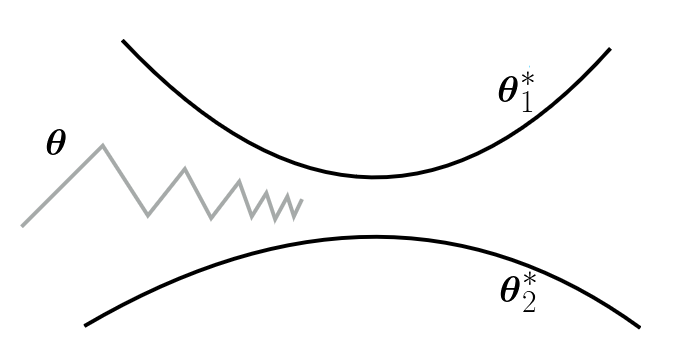}
    \caption{Schematic visualization of Reptile's learning trajectory. Here, $\boldsymbol{\theta}_{1}^*$ and $\boldsymbol{\theta}_{2}^*$ are the optimal weights for tasks $\Tau_{1}$ and $\Tau_{2}$ respectively. The initialization parameters $\boldsymbol{\theta}$ oscillate between these. Adapted from  \citet{nichol2018reptile}.}
    \label{fig:reptile}
\end{figure}

The gradient of this expected distance with respect to the initialization $\boldsymbol{\theta}$ is given by 

\begin{align}
    \nabla_{\boldsymbol{\theta}} \mathbb{E}_{\Tau_{j} \backsim p(\Tau)} [\frac{1}{2} d(\boldsymbol{\theta}, \boldsymbol{\theta}^{*}_{j})^{2}]  
    &=  \mathbb{E}_{\Tau_{j} \backsim p(\Tau)} [\frac{1}{2} \nabla_{\boldsymbol{\theta}} d(\boldsymbol{\theta}, \boldsymbol{\theta}^{*}_{j})^{2}]   \nonumber \\
    &= \mathbb{E}_{\Tau_{j} \backsim p(\Tau)} [\boldsymbol{\theta} - \boldsymbol{\theta}^{*}_{j}], 
\end{align} where we used the fact that the gradient of the squared Euclidean distance between two points $\boldsymbol{x}_{1}$ and $\boldsymbol{x}_{2}$ is the vector $2(\boldsymbol{x}_{1} - \boldsymbol{x}_{2})$. \citet{nichol2018reptile} go on to argue that performing gradient descent on this objective would result in the following update rule

\begin{align}
    \boldsymbol{\theta} &= \boldsymbol{\theta} - \epsilon \nabla_{\boldsymbol{\theta}} \frac{1}{2}d(\boldsymbol{\theta}, \boldsymbol{\theta}^{*}_{j})^{2} \nonumber \\
    &= \boldsymbol{\theta} - \epsilon(\boldsymbol{\theta}^{*}_{j} - \boldsymbol{\theta}).
\end{align} Since we do not know $\boldsymbol{\theta}^{*}_{\Tau_{j}}$, one can approximate this by term by $k$ steps of gradient descent $SGD(\mathcal{L}_{\Tau_{j}}, \boldsymbol{\theta}, k)$. In short, Reptile can be seen as gradient descent on the distance minimization objective given in \autoref{eq:distanceObj}. A visualization is shown in \autoref{fig:reptile}. The initialization $\boldsymbol{\theta}$ is moving towards the optimal weights for tasks 1 and 2 in interleaved fashion (hence the oscillations). 

In conclusion, Reptile is an extremely simple meta-learning technique, which does not need to differentiate through the optimization trajectory like, e.g., MAML \citep{Finn17}, saving time and memory costs. However, the theoretical foundation is a bit weaker due to the fact that it does not directly optimize for fast learning as done by MAML, and performance may be a bit worse than that of MAML in some settings. 

\subsection{Latent embedding optimization (LEO)}

Latent Embedding Optimization, or LEO, was proposed by \citet{rusu2018meta} to combat an issue of gradient-based meta-learners, such as MAML (see \autoref{sec:maml}), in few-shot settings ($N$-way, $k$-shot). These techniques operate in a high-dimensional parameter space using gradient information from only a few examples, which could lead to poor generalization. 

\begin{figure}[hb]
    \centering
    \includegraphics[width=\linewidth]{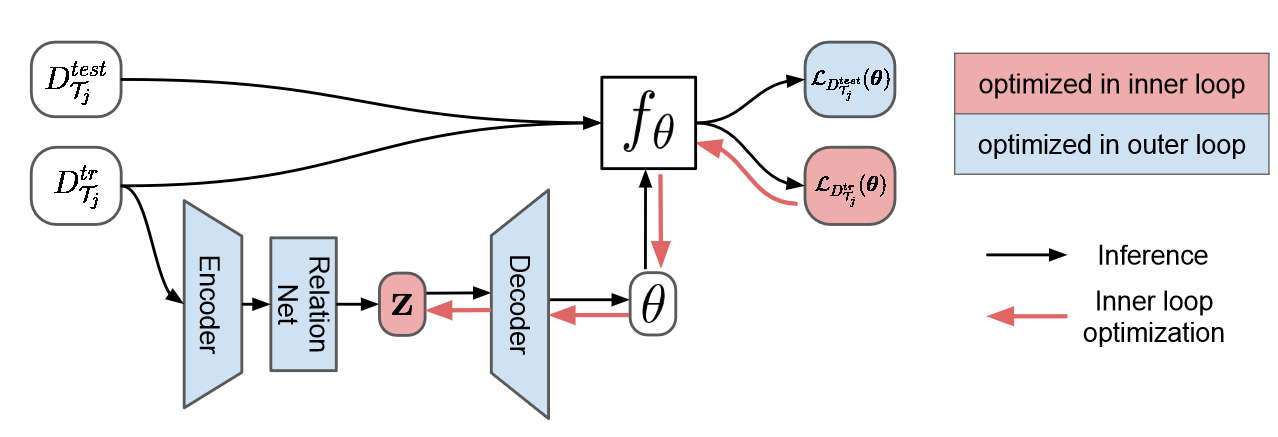}
    \caption{Workflow of LEO. adapted from \citet{rusu2018meta}. }
    \label{fig:leo}
\end{figure}

LEO alleviates this issue by learning a lower-dimensional latent embedding space, which indirectly allows us to learn a good set of initial parameters $\boldsymbol{\theta}$. Additionally, the embedding space is conditioned upon tasks, allowing for more expressivity.  In theory, LEO could find initial parameters for the entire base-learner network, but the authors only experimented with setting the parameters for the final layers. 

The complete workflow of LEO is shown in \autoref{fig:leo}. As we can see, given a task $\Tau_{j}$, the corresponding support set $D^{tr}_{\Tau_{j}}$ is fed into an encoder, which produces hidden codes for each example in that set. These hidden codes are paired and concatenated in every possible manner, granting us $(Nk)^{2}$ pairs, where $N$ is the number of classes in the training set, and $k$ the number of examples per class. These paired codes are then fed into a relation net \citep{sung2018learning} (see \autoref{sec:relnets}). The resulting embeddings are grouped by class, and parameterize a probability distribution over latent codes $\boldsymbol{z}_{n}$ (for class $n$) in a low dimensional space $\mathcal{Z}$. More formally, let $\boldsymbol{x}^{\ell}_{n}$ denote the $\ell$-th example of class $n$ in $D^{tr}_{\Tau_{j}}$. Then, the mean $\boldsymbol{\mu}^{e}_{n}$ and variance $\boldsymbol{\sigma}^{e}_{n}$ of a Gaussian distribution over latent codes for class $n$ are computed as

\begin{align}
    \boldsymbol{\mu}_{n}^{e}, \boldsymbol{\sigma}^{e}_{n} = \frac{1}{Nk^{2}}\sum_{\ell_{p}=1}^{k}\sum^{N}_{m=1} \sum_{\ell_{q}=1}^{k} g_{\boldsymbol{\phi}_{r}}\left( g_{\boldsymbol{\phi}_{e}}(\boldsymbol{x}^{\ell_{p}}_{n}), g_{\boldsymbol{\phi}_{e}}(\boldsymbol{x}^{\ell_{q}}_{m})  \right),
\end{align} where $\boldsymbol{\phi}_{r}, \boldsymbol{\phi}_{e}$ are parameters for the relation net and encoder respectively. Intuitively, the three summations ensure that every example with class $n$ in $D^{tr}_{\Tau_{j}}$ is paired with every example from all classes $n$. Given $\boldsymbol{\mu}_{n}^{e}$, and $\boldsymbol{\sigma}_{n}^{e}$, one can sample a latent code $\boldsymbol{z}_{n} \backsim N(\boldsymbol{\mu}_{n}^{e}, diag(\boldsymbol{\sigma}_{n}^{e2}))$ for class $n$, which serves as latent embedding of the task training data.

The decoder can then generate a task-specific initialization $\boldsymbol{\theta}_{n}$ for class $n$ as follows. First, one computes a mean and variance for a Gaussian distribution using the latent code
\begin{align}
    \boldsymbol{\mu}_{n}^{d}, \boldsymbol{\sigma}_{n}^{d} = g_{\boldsymbol{\phi}_{d}}(\boldsymbol{z}_{n}).
\end{align} These are then used to sample initialization weights $\boldsymbol{\theta}_{n} \backsim N(\boldsymbol{\mu}^{d}_{n}, diag(\boldsymbol{\sigma}^{d2}_{n}))$. The loss from the generated weights can then be propagated backwards to adjust the embedding space. In practice, generating such a high-dimensional set of parameters from a low-dimensional embedding can be quite problematic. Therefore, LEO uses pre-trained models, and only generates weights for the final layer, which limits the expressivity of the model.   

A key advantage of LEO is that it optimizes in a lower-dimensional latent embedding space, which aids generalization performance. However, the approach is more complex than e.g.\ MAML \citep{Finn17}, and its applicability is limited to few-shot learning settings.

\subsection{Online MAML (FTML)}

Online MAML \citep{finn2019online} is an extension of MAML \citep{Finn17} to make it applicable to \textit{online learning} settings \citep{anderson2008theory}. In the online setting, we are presented with a sequence of tasks $\Tau_{t}$ with corresponding loss functions $\{ \mathcal{L}_{\Tau_{t}} \}_{t=1}^{T}$, for some potentially infinite time horizon $T$. The goal is to pick a sequence of parameters $\{ \boldsymbol{\theta}_{t} \}_{t=1}^{T}$ that performs well on the presented loss functions. This objective is captured by the $Regret_{T}$ over the entire sequence, which is defined by \citet{finn2019online} as follows

\begin{align}
    Regret_{T} = \sum_{t=1}^{T}\mathcal{L}_{\Tau_{t}}(\boldsymbol{\theta}_{t}') - min_{\boldsymbol{\theta}} \sum_{t=1}^{T}\mathcal{L}_{\Tau_{t}}(\boldsymbol{\theta}'_{t}),
\end{align} where $\boldsymbol{\theta}$ are the initial model parameters (just as MAML), and $\boldsymbol{\theta}_{t}'$ are parameters resulting from a one-step gradient update (starting from $\boldsymbol{\theta}$) on task $t$. Here, the left term reflects the updated parameters chosen by the agent $(\boldsymbol{\theta}_{t})$, whereas the right term presents the minimum obtainable loss (in hindsight) from a single fixed set of parameters $\boldsymbol{\theta}$. Note that this setup assumes that the agent can make updates to its chosen parameters (transform its initial choice at time $t$ from $\boldsymbol{\theta}_{t}$ to $\boldsymbol{\theta}_{t}'$).

\citet{finn2019online} propose FTML (Follow The Meta Leader), inspired by FTL (Follow The Leader) \citep{hannan1957approximation, kalai2005efficient}, to minimize the regret. The basic idea is to set the parameters for the next time step ($t+1$) equal to the best parameters in hindsight, i.e.,

\begin{align}
    \boldsymbol{\theta}_{t+1} := argmin_{\boldsymbol{\theta}} \sum_{k=1}^{t}\mathcal{L}_{\Tau_{k}}(\boldsymbol{\theta}_{k}').\label{eq:ftmlGradient} 
\end{align} The gradient to perform meta-updates is then given by 

\begin{align}
    g_{t}(\boldsymbol{\theta}) := \nabla_{\boldsymbol{\theta}} \mathbb{E}_{\Tau_{k} \backsim p_{t}(\Tau)} \mathcal{L}_{\Tau_{k}}(\boldsymbol{\theta}_{k}'),
\end{align} where $p_{t}(\Tau)$ is a uniform distribution over tasks $1,\ldots ,t$ (at time $t$).

\autoref{alg:ftml} contains the full pseudocode for FTML. In this algorithm, $\mathit{MetaUpdate}$ performs a few ($N_{meta}$) meta-steps. In each meta-step, a task is sampled from $B$, together with train and test mini-batches to compute the gradient $g_{t}$ in \autoref{eq:ftmlGradient}. The initialization $\boldsymbol{\theta}$ is then updated ($\boldsymbol{\theta} := \boldsymbol{\theta} - \beta g_{t}(\boldsymbol{\theta})$), where $\beta$ is the meta-learning rate. Note that the memory usage keeps increasing over time, as at every time step $t$, we append tasks to the buffer $B$, and keep task data sets in memory.  

\begin{algorithm}
\caption{FTML by \citet{finn2019online}}\label{alg:ftml}
\begin{algorithmic}[1]
\Require{Performance threshold $\gamma$}
\State Initialize empty task buffer $B$
\For{$t = 1,\ldots$}
    \State Initialize data set $D_{t} = \emptyset$
    \State Append $\Tau_{t}$ to B
    \While{$|D_{t}| < N$}
        \State Append batch of data $\{(\boldsymbol{x}_{i}, y_{i})\}_{i=1}^{n}$ to $D_{t}$
        \State $\boldsymbol{\theta}_{t} = \mathit{MetaUpdate}(\boldsymbol{\theta}_{t}, B, t)$
        \State Compute $\boldsymbol{\theta}'_{t}$
        \If{$\mathcal{L}_{D^{test}_{\Tau_{t}}}(\boldsymbol{\theta}'_{t}) < \gamma$}
            \State Save $|D_{t}|$ as the efficiency for task $\Tau_{t}$
        \EndIf
    \EndWhile
    \State Save final performance $\mathcal{L}_{D^{test}_{\Tau_{t}}}$($\boldsymbol{\theta}'_{t}$)
    \State $\boldsymbol{\theta}_{t+1} = \boldsymbol{\theta}_{t}$
\EndFor
\end{algorithmic}
\end{algorithm}

In summary, Online MAML is a robust technique for online-learning \citep{finn2019online}. A downside of this approach is the computational costs that keep growing over time, as all encountered data are stored. Reducing these costs is a direction for future work. Also, one could experiment with how well the approach works when more than one inner gradient update steps per task are used, as mentioned by \citet{finn2019online}.

\subsection{LLAMA}

\citet{grant2018recasting} mold MAML into a probabilistic framework, such that a probability distribution over task-specific parameters $\boldsymbol{\theta}_{j}'$ is learned, instead of a single one. In this way, multiple potential solutions can be obtained for a task. The resulting technique is called LLAMA (Laplace Approximation for Meta-Adaptation). Importantly, LLAMA is only developed for supervised learning settings.

A key observation is that a neural network $f_{\boldsymbol{\theta}'_{j}}$, parameterized by updated parameters $\boldsymbol{\theta}'_{j}$ (obtained from few gradient updates using $D^{tr}_{\Tau_{j}}$), outputs class probabilities $p(y_{i}| \boldsymbol{x}_{i}, \boldsymbol{\theta}'_{j})$.
To minimize the error on the query set $D^{test}_{\Tau_{j}}$, the model must output large probability scores for the true classes.
This objective is captured in the maximum log-likelihood loss function

\begin{align}
    \mathcal{L}_{D^{test}_{\Tau_{j}}}(\boldsymbol{\theta}'_{j}) = - \sum_{\boldsymbol{x}_{i},y_{i} \in D^{test}_{\Tau_{j}}} log \, p(y_{i} | \boldsymbol{x}_{i}, \boldsymbol{\theta}'_{j}).
\end{align}

Simply put, if we see a task $j$ as a probability distribution over examples $p_{\Tau_{j}}$, we wish to maximize the probability that the model predicts the correct class $y_{i}$, given an input $\boldsymbol{x}_{i}$. This can be done by plain gradient descent, as shown in \autoref{alg:llama}, where $\beta$ is the meta-learning rate. Line 4 refers to ML-LAPLACE, which is a subroutine that computes task-specific updated parameters $\boldsymbol{\theta}'_{j}$, and estimates the negative log likelihood (loss function) which is used to update the initialization $\boldsymbol{\theta}$, as shown in \autoref{alg:llamaSUB}. \citet{grant2018recasting} approximated the quadratic curvature matrix $\hat{H}$ using K-FAC \citep{martens2015optimizing}. 

The trick is that the initialization $\boldsymbol{\theta}$ defines a distribution $p(\boldsymbol{\theta}'_{j}|\boldsymbol{\theta})$ over task-specific parameters $\boldsymbol{\theta}'_{j}$. This distribution was taken to be a diagonal Gaussian \citep{grant2018recasting}. Then, to sample solutions for a new task $\Tau_{j}$, one can simply generate possible solutions $\boldsymbol{\theta}'_{j}$ from the learned Gaussian distribution. 

\begin{algorithm}
\caption{LLAMA by \citet{grant2018recasting}}\label{alg:llama}
\begin{algorithmic}[1]
\State Initialize $\boldsymbol{\theta}$ randomly
\While{not converged}
    \State Sample a batch of $J$ tasks: $B = \Tau_{1},\ldots ,\Tau_{J} \backsim p(\Tau)$
    \State Estimate $\mathbb{E}_{(\boldsymbol{x}_{i}, y_{i}) \backsim p_{\Tau_{j}}}[-log \, p(y_{i} | \boldsymbol{x}_{i}, \boldsymbol{\theta})] \, \forall \Tau_{j} \in B$ using ML-LAPLACE
    \State $\boldsymbol{\theta} = \boldsymbol{\theta} - \beta \nabla_{\boldsymbol{\theta}}\sum_{j} \mathbb{E}_{(\boldsymbol{x}_{i}, y_{i}) \backsim p_{\Tau_{j}}}[-log \, p(y_{i} | \boldsymbol{x}_{i}, \boldsymbol{\theta})$
\EndWhile
\end{algorithmic}
\end{algorithm}

\begin{algorithm}
\caption{ML-LAPLACE \citep{grant2018recasting}}\label{alg:llamaSUB}
\begin{algorithmic}[1]
\State $\boldsymbol{\theta}'_{j} = \boldsymbol{\theta}$
\For{$k=1,\ldots ,K$}
    \State $\boldsymbol{\theta}'_{j} = \boldsymbol{\theta}'_{j} + \alpha \nabla_{\boldsymbol{\theta}'_{j}} log \, p(y_{i} \in D^{tr}_{\Tau_{j}}|\boldsymbol{\theta}'_{j}, \boldsymbol{x}_{i} \in D^{tr}_{\Tau_{j}})$
\EndFor
\State Compute curvature matrix $\hat{H} = \nabla_{\boldsymbol{\theta}'_{j}}^{2}[-log \, p(y_{i} \in D^{test}_{\Tau_{j}}| \boldsymbol{\theta}'_{j}, \boldsymbol{x}_{i} \in D^{test}_{\Tau_{j}})] + \nabla_{\boldsymbol{\theta}'_{j}}^{2}[-log \, p(\boldsymbol{\theta}'_{j}|\boldsymbol{\theta})]$
\State \Return $-log \, p(y_{i} \in D^{test}_{\Tau_{j}}| \boldsymbol{\theta}'_{j}, \boldsymbol{x}_{i} \in D^{test}_{\Tau_{j}}) + \eta \,log[det(\hat{H})]$
\end{algorithmic}
\end{algorithm}

In short, LLAMA extends MAML in a probabilistic fashion, such that one can obtain multiple solutions for a single task, instead of one. This does, however, increase the computational costs. On top of that, the used Laplace approximation (in ML-LAPLACE) can be quite inaccurate \citep{grant2018recasting}.

\subsection{PLATIPUS}

PLATIPUS \citep{finn2018probabilistic} builds upon the probabilistic interpretation of LLAMA \citep{grant2018recasting}, but learns a probability distribution over initializations $\boldsymbol{\theta}$, instead of task-specific parameters $\boldsymbol{\theta}_{j}'$. Thus, PLATIPUS allows one to sample an initialization $\boldsymbol{\theta} \backsim p(\boldsymbol{\theta})$, which can be updated with gradient descent to obtain task-specific weights (fast weights) $\boldsymbol{\theta}_{j}'$.  

\begin{algorithm}
\caption{PLATIPUS training algorithm by \citet{finn2018probabilistic}}\label{alg:platipus}
\begin{algorithmic}[1]
\State Initialize $\boldsymbol{\Theta} = \{ \boldsymbol{\mu}_{\boldsymbol{\theta}}, \boldsymbol{\sigma}^{2}_{\boldsymbol{\theta}}, \boldsymbol{v}_{q}, \boldsymbol{\gamma}_{p}, \boldsymbol{\gamma}_{q} \}$
\While{Not done}
    \State Sample batch of tasks $B = \{\Tau_{j} \backsim p(\Tau)\}_{i=1}^{m}$
    \For{$\Tau_{j} \in B$}
        \State $D^{tr}_{\Tau_{j}}, D^{test}_{\Tau_{j}} = \Tau_{j}$
        \State Compute $\nabla_{\boldsymbol{\mu}_{\boldsymbol{\theta}}} \mathcal{L}_{D^{test}_{\Tau_{j}}}(\boldsymbol{\mu}_{\boldsymbol{\theta}})$
        \State Sample $\boldsymbol{\theta} \backsim q = N(\boldsymbol{\mu}_{\boldsymbol{\theta}} - \boldsymbol{\gamma}_{q} \nabla_{\boldsymbol{\mu}_{\boldsymbol{\theta}}} \mathcal{L}_{D^{test}_{\Tau_{j}}}(\boldsymbol{\mu}_{\boldsymbol{\theta}}), \boldsymbol{v}_{q})$
        \State Compute $\nabla_{\boldsymbol{\theta}} \mathcal{L}_{D^{tr}_{\Tau_{j}}}(\boldsymbol{\theta})$
        \State Compute fast weights $\boldsymbol{\theta}'_{i} = \boldsymbol{\theta} - \alpha \nabla_{\boldsymbol{\theta}} \mathcal{L}_{D^{tr}_{\Tau_{j}}}(\boldsymbol{\theta})$
    \EndFor
    \State $p(\boldsymbol{\theta} | D^{tr}_{\Tau_{j}}) = N(\boldsymbol{\mu}_{\boldsymbol{\theta}} - \boldsymbol{\gamma}_{p} \nabla_{\boldsymbol{\mu}_{\boldsymbol{\theta}}} \mathcal{L}_{D^{tr}_{\Tau_{j}}}(\boldsymbol{\mu}_{\boldsymbol{\theta}}), \boldsymbol{\sigma}^{2}_{\boldsymbol{\theta}}  ) $
    \State Compute $\nabla_{\boldsymbol{\Theta}}\left[ \sum_{\Tau_{j}} \mathcal{L}_{D^{test}_{\Tau_{j}}}(\boldsymbol{\phi}_{i}) + D_{\mathit{KL}}( q(\boldsymbol{\theta} | D^{test}_{\Tau_{j}}), p(\boldsymbol{\theta} | D^{tr}_{\Tau_{j}}) )   \right]$
    \State Update $\boldsymbol{\Theta}$ using the Adam optimizer
\EndWhile
\end{algorithmic}
\end{algorithm}

The approach is best explained by its pseudocode, as shown in \autoref{alg:platipus}. In contrast to the original MAML, PLATIPUS introduces five more parameter vectors (line 1). All of these parameters are used to facilitate the creation of Gaussian distributions over prior initializations (or simply priors) $\boldsymbol{\theta}$. That is, $\boldsymbol{\mu}_{\boldsymbol{\theta}}$ represents the vector mean of the distributions. $\boldsymbol{\sigma}^{2}_{\boldsymbol{q}}$, and $\boldsymbol{v}_{q}$ represent the covariances of train and test distributions respectively. $\boldsymbol{\gamma}_{x}$ for $x = q,p$ are learning rate vectors for performing gradient steps on distributions $q$ (line 6 and 7) and $P$ (line 11). 

The key difference with the regular MAML is that instead of having a single initialization point $\boldsymbol{\theta}$, we now learn distributions over priors: $q$ and $P$, which are based on query and support data sets of task $\Tau_{j}$ respectively. Since these data sets come from the same task, we want the distributions $q(\boldsymbol{\theta} | D^{test}_{\Tau_{j}})$, and $p(\boldsymbol{\theta} | D^{tr}_{\Tau_{j}})$ to be close to each other. This is enforced by the Kullback–Leibler divergence ($D_{\mathit{KL}}$) loss term on line 12, which measures the distance between the two distributions. Importantly, note that $q$ (line 7) and $P$ (line 11) use vector means which are computed with one gradient update steps using the query and support data sets respectively. The idea is that the mean of the Gaussian distributions should be close to the updated mean $\boldsymbol{\mu}_{\boldsymbol{\theta}}$ because we want to enable fast learning.  As one can see, the training process is very similar to that of MAML \citep{Finn17} (\autoref{sec:maml}), with some small adjustments to allow us to work with the probability distributions over $\boldsymbol{\theta}$.

At test-time, one can simply sample a new initialization $\boldsymbol{\theta}$ from the prior distribution $p(\boldsymbol{\theta} | D^{tr}_{\Tau_{j}})$ (note that $q$ cannot be used at test-time as we do not have access to $D^{test}_{\Tau_{j}}$), and apply a gradient update on the provided support set $D^{tr}_{\Tau_{j}}$. Note that this allows us to sample multiple potential initializations $\boldsymbol{\theta}$ for the given task. 

The key advantage of PLATIPUS is that it is aware of its uncertainty, which greatly increases the applicability of Deep Meta-Learning in critical domains such as medical diagnosis \citep{finn2018probabilistic}. Based on this uncertainty, it can ask for labels of some inputs it is unsure about (active learning). A downside to this approach, however, is the increased computational costs, and the fact that it is not applicable to reinforcement learning.

\subsection{Bayesian MAML (BMAML)}

Bayesian MAML \citep{yoon2018bayesian} is another probabilistic variant of MAML that can generate multiple solutions. However, instead of learning a distribution over potential solutions, BMAML simply keeps $M$ possible solutions, and optimizes them in joint fashion. Recall that probabilistic MAMLs (e.g., PLATIPUS) attempt to maximize the data likelihood of task $\Tau_{j}$, i.e., $p(\boldsymbol{y}^{test}_{j}| \boldsymbol{\theta}'_{j})$, where $\boldsymbol{\theta}'_{j}$ are task-specific fast weights obtained by one or more gradient updates. \citet{yoon2018bayesian} model this likelihood using Stein Variational Gradient Descent (SVGD) \citep{liu2016stein}. 

To obtain $M$ solutions, or equivalently, parameter settings $\boldsymbol{\theta}^{m}$, SVGD keeps a set of $M$ \textit{particles} $\boldsymbol{\Theta} = \{ \boldsymbol{\theta}^{m} \}_{i=1}^{M}$. At iteration $t$, every $\boldsymbol{\theta}_{t} \in \boldsymbol{\Theta}$ is updated as follows

\begin{align}
    \boldsymbol{\theta}_{t+1} = \boldsymbol{\theta}_{t} + \epsilon (\phi (\boldsymbol{\theta}_{t})) \\ \text{ where } \phi(\boldsymbol{\theta}_{t}) = \frac{1}{M}\sum_{m=1}^{M} \left[ k(\boldsymbol{\theta}^{m}_{t}, \boldsymbol{\theta}_{t}) \nabla_{\boldsymbol{\theta}^{m}_{t}} log \, p(\boldsymbol{\theta}_{t}^{m}) + \nabla_{\boldsymbol{\theta}_{t}^{m}}k(\boldsymbol{\theta}^{m}_{t}, \boldsymbol{\theta}_{t}) \right]. 
\end{align} Here, $k(\boldsymbol{x}, \boldsymbol{x}')$ is a similarity kernel between $\boldsymbol{x}$ and $\boldsymbol{x}'$. The authors used a radial basis function (RBF) kernel, but in theory, any other kernel could be used. Note that the update of one particle depends on the other gradients of particles. The first term in the summation ($k(\boldsymbol{\theta}^{m}_{t}, \boldsymbol{\theta}_{t}) \nabla_{\boldsymbol{\theta}^{m}_{t}} log \, p(\boldsymbol{\theta}_{t}^{m})$) moves the particle in the direction of the gradients of other particles, based on particle similarity. The second term ($\nabla_{\boldsymbol{\theta}_{t}^{m}}k(\boldsymbol{\theta}^{m}_{t}, \boldsymbol{\theta}_{t})$) ensures that particles do not collapse (repulsive force) \citep{yoon2018bayesian}. 

These particles can then be used to approximate the probability distribution of the test labels

\begin{align}
    p(\boldsymbol{y}^{test}_{j}| \boldsymbol{\theta}'_{j}) \approx \frac{1}{M} \sum_{m=1}^{M} p(\boldsymbol{y}_{j}^{test} | \boldsymbol{\theta}^{m}_{\Tau_{j}}),
\end{align} where $\boldsymbol{\theta}_{\Tau_{j}}^{m}$ is the $m$-th particle obtained by training on the support set $D^{tr}_{\Tau_j}$ of task $\Tau_{j}$.

\citet{yoon2018bayesian} proposed a new meta-loss to train BMAML, called the \textit{Chaser Loss}. This loss relies on the insight that we want the approximated parameter distribution (obtained from the support set $p^{n}_{\Tau_{j}}(\boldsymbol{\theta}_{\Tau_{j}} | D^{tr}, \boldsymbol{\Theta}_{0})$) and true distribution $p^{\infty}_{\Tau_{j}}(\boldsymbol{\theta}_{\Tau_{j}}|D^{tr} \cup D^{test})$ to be close to each other (since the task is the same). Here, $n$ denotes the number of SVGD steps, and $\boldsymbol{\Theta}_{0}$ is the set of initial particles, in similar fashion to the initial parameters $\boldsymbol{\theta}$ seen by MAML. 
Since the true distribution is unknown, \citet{yoon2018bayesian} approximate it by running SVGD for $s$ additional steps, granting us the \textit{leader} $\boldsymbol{\Theta}^{n + s}_{\Tau_{j}}$, where the $s$ additional steps are performed on the combined support and query set. The intuition is that as the number of updates increases, the obtained distributions become more like the true ones. $\boldsymbol{\Theta}^{n}_{\Tau_{j}}$ in this context is called the \textit{chaser} as it wants to get closer to the leader. The proposed meta-loss is then given by 

\begin{align}
    \mathcal{L}_{BMAML}(\boldsymbol{\Theta}_{0}) = \sum_{\Tau_{j} \in B}\sum_{m=1}^{M} || \boldsymbol{\theta}_{\Tau_{j}}^{n,m} - \boldsymbol{\theta}_{\Tau_{j}}^{n+s,m} ||^{2}_{2}. 
\end{align} 

The full pseudocode of BMAML is shown in \autoref{alg:baymaml}. Here, $\boldsymbol{\Theta}^{n}_{\Tau_{j}}(\boldsymbol{\Theta}_{0})$ denotes the set of particles after $n$ updates on task $\Tau_{j}$, and $SG$ means ``stop gradients" (we do not want the leader to depend on the initialization, as the leader must lead).

\begin{algorithm}
\caption{BMAML by \citet{yoon2018bayesian}}\label{alg:baymaml}
\begin{algorithmic}[1]
\State Initialize $\boldsymbol{\Theta}_{0}$
\For{$t=1,\ldots$ until convergence}
\State Sample a batch of tasks B from $p(\Tau)$
\For{task $\Tau_{j} \in B$}
\State Compute chaser $\boldsymbol{\Theta}^{n}_{\Tau_{j}}(\boldsymbol{\Theta}_{0}) = SVGD_{n}(\boldsymbol{\Theta}_{0}; D^{tr}_{\Tau_{j}}, \alpha)$
\State Compute leader $\boldsymbol{\Theta}^{n+s}_{\Tau_{j}}(\boldsymbol{\Theta}_{0}) = SVGD_{s}(\boldsymbol{\Theta}^{n}_{\Tau_{j}}(\boldsymbol{\Theta}_{0}); D^{tr}_{\Tau_{j}} \cup D^{test}_{\Tau_j}, \alpha)$
\EndFor
\State $\boldsymbol{\Theta}_{0} = \boldsymbol{\Theta}_{0} - \beta \nabla_{\boldsymbol{\Theta}_{0}}\sum_{\Tau_{j} \in B} d(\boldsymbol{\Theta}^{n}_{\Tau_{j}}(\boldsymbol{\Theta}_{0}), SG(\boldsymbol{\Theta}^{n+s}_{\Tau_{j}}(\boldsymbol{\Theta}_{0})))$
\EndFor
\end{algorithmic}
\end{algorithm}

In summary, BMAML is a robust optimization-based meta-learning technique that can propose $M$ potential solutions to a task. Additionally, it is applicable to reinforcement learning by using Stein Variational Policy Gradient instead of SVGD. A downside of this approach is that one has to keep $M$ parameter sets in memory, which does not scale well. Reducing the memory costs is a direction for future work \citep{yoon2018bayesian}. Furthermore, SVGD is sensitive to the selected kernel function, which was pre-defined in BMAML. However, \citet{yoon2018bayesian} point out that it may be beneficial to learn the kernel function instead. This is another possibility for future research.

\subsection{Simple Differentiable Solvers}

\citet{Bertinetto19} take a quite different approach. That is, they pick simple base-learners that have an analytical closed-form solution. The intuition is that the existence of a closed-form solution allows for good learning efficiency. They propose two techniques using this principle, namely \textbf{R2-D2} (Ridge Regression Differentiable Discriminator), and \textbf{LR-D2} (Logistic Regression Differentiable Discriminator). We cover both in turn. 

Let $g_{\boldsymbol{\phi}}: X \rightarrow \mathbb{R}^{e}$ be a pre-trained input embedding model (e.g.\ a CNN), which outputs embeddings with a dimensionality of $e$. Furthermore, assume that we use a linear predictor function $f(g_{\boldsymbol{\phi}}(\boldsymbol{x}_{i})) = g_{\boldsymbol{\phi}}(\boldsymbol{x}_{i})W$, where $W$ is a $e \times o$ weight matrix, and $o$ is the output dimensionality (of the label). When using (regularized) Ridge Regression (done by R2-D2), one uses the optimal $W$, i.e.,

\begin{align}
    W^{*} &= \argmin_{W} \, || XW - Y||^{2}_{2} + \gamma ||W||^{2} \nonumber \\
    &= (X^{T}X + \gamma I)^{-1}X^{T}Y, \label{eq:ridgeregression}
\end{align} where $X \in \mathbb{R}^{n \times e}$ is the input matrix, containing $n$ rows (one for each embedded input $g_{\boldsymbol{\phi}}(\boldsymbol{x}_{i})$), $Y \in \mathbb{R}^{n \times o}$ is the output matrix with correct outputs corresponding to the inputs, and $\gamma$ is a regularization term to prevent overfitting. Note that the analytical solution contains the term $(X^{T}X) \in \mathbb{R}^{e \times e}$, which is quadratic in the size of the embeddings. Since $e$ can become quite large when using deep neural networks, \citet{Bertinetto19} use Woodburry's identity

\begin{align}
    W^{*} = X^{T}(XX^{T} + \gamma I)^{-1} Y,
\end{align} where $XX^{T} \in \mathbb{R}^{n \times n}$ is linear in the embedding size, and quadratic in the number of examples, which is more manageable in few-shot settings, where $n$ is very small. To make predictions with this Ridge Regression based model, one can compute 

\begin{align}
    \hat{Y} = \alpha X_{test}W^{*} + \beta,
\end{align} where $\alpha$ and $\beta$ are hyperparameters of the base-learner that can be learned by the meta-learner, and $X_{test} \in \mathbb{R}^{m \times e}$ corresponds to the $m$ test inputs of a given task. Thus, the meta-learner needs to learn $\alpha, \beta, \gamma$, and $\boldsymbol{\phi}$ (embedding weights of the CNN).  

The technique can also be applied to iterative solvers when the optimization steps are differentiable \citep{Bertinetto19}. LR-D2 uses the Logistic Regression objective and Newton's method as solver. Outputs $\boldsymbol{y} \in \{-1,+1\}^{n}$ are now binary. Let $\boldsymbol{w}$ denote a parameter row of our linear model (parameterized by $W$). Then, the $i$-th iteration of Newton's method updates $\boldsymbol{w}_{i}$ as follows

\begin{align}
    \boldsymbol{w}_{i} = (X^{T}\mbox{diag}(\boldsymbol{s}_{i})X + \gamma I)^{-1}X^{T}\mbox{diag}(\boldsymbol{s}_{i})\boldsymbol{z}_{i}, 
\end{align} where $\boldsymbol{\mu}_{i} = \sigma(\boldsymbol{w}^{T}_{i-1}X)$, $\boldsymbol{s}_{i} = \boldsymbol{\mu}_{i}(1 - \boldsymbol{\mu}_{i})$, $\boldsymbol{z}_{i} = \boldsymbol{w}^{T}_{i-1}X + (\boldsymbol{y} - \boldsymbol{\mu}_{i})/\boldsymbol{s}_{i}$, and $\sigma$ is the sigmoid function. Since the term $X^{T}\mbox{diag}(\boldsymbol{s}_{i})X$ is a matrix of size $e \times e$, and thus again quadratic in the embedding size, Woodburry's identity is also applied here to obtain
\begin{align}
    \boldsymbol{w}_{i} = X^{T}(XX^{T} + \lambda \mbox{diag}(\boldsymbol{s}_{i})^{-1})^{-1}\boldsymbol{z}_{i},
\end{align} making it quadratic in the input size, which is not a big problem since $n$ is small in the few-shot setting. The main difference compared to R2-D2 is that the base-solver has to be run for multiple iterations to obtain $W$.

In the few-shot setting, the base-level optimizers compute the weight matrix $W$ for a given task $\Tau_{i}$. The obtained loss on the query set of a task $\mathcal{L}_{D_{test}}$ is then used to update the parameters $\boldsymbol{\phi}$ of the input embedding function (e.g.\ CNN) and the hyperparameters of the base-learner.

\citet{lee2019meta} have done similar work to \citet{Bertinetto19}, but with linear Support Vector Machines (SVMs) as base-learner. Their approach is dubbed \textbf{MetaOptNet} and achieved state-of-the-art performance on few-shot image classification. 

In short, simple differentiable solvers are simple, reasonably fast in terms of computation time, but limited to few-shot learning settings. Investigating the use of other simple base-learners is a direction for future work.

\subsection{Optimization-based Techniques, in conclusion}
\begin{figure}
    \centering
    \includegraphics[width=\linewidth]{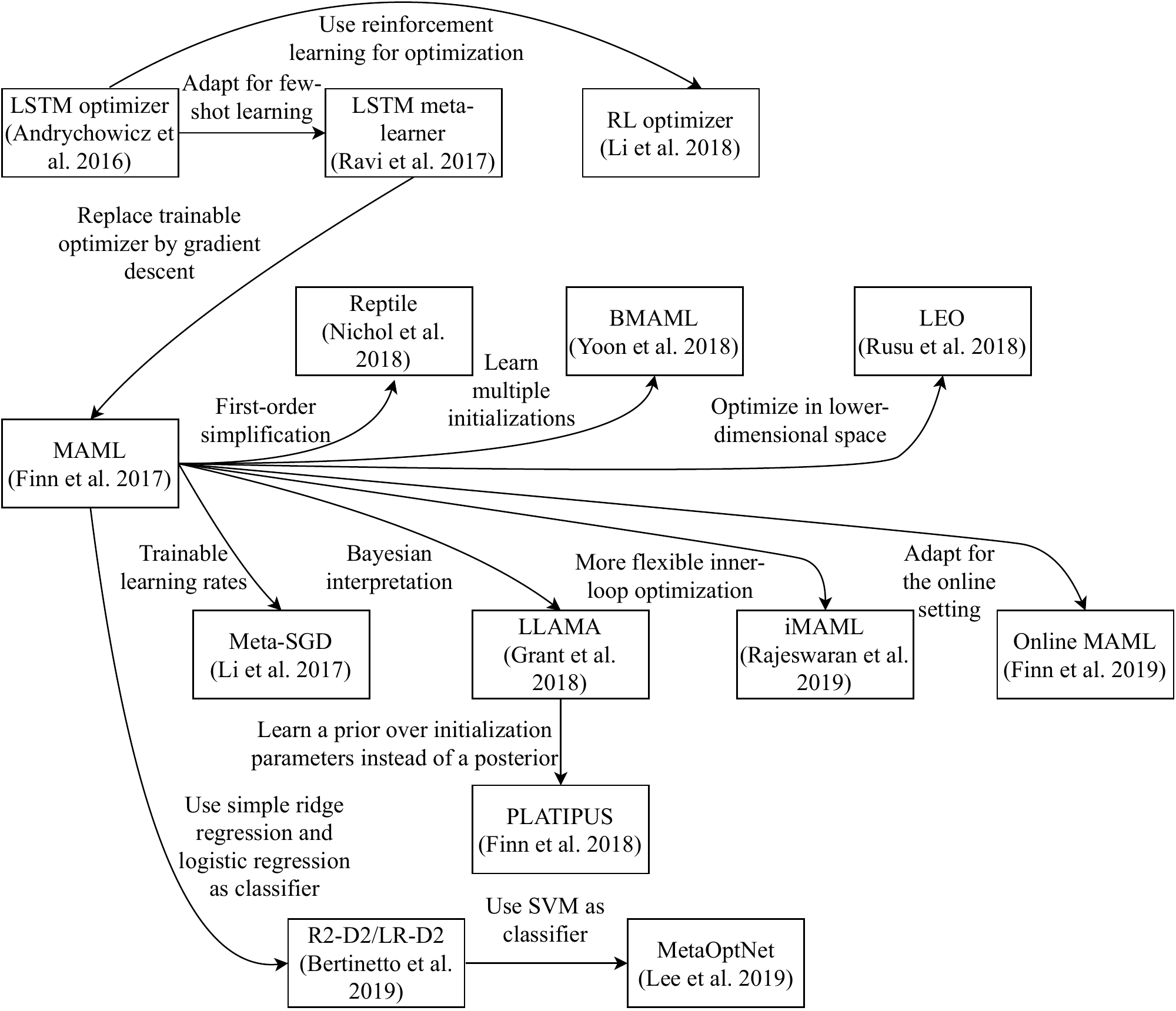}
    \caption{The relationships between the covered optimization-based meta-learning techniques. As one can see, MAML has a central position in this graph of techniques as it has inspired many other works.}
    \label{fig:optbasedrels}
\end{figure}

Optimization-based\index{meta-learning!optimization-based} aim to learn new tasks quickly through (learned) optimization procedures. Note that this closely resembles base-level learning, which also occurs through optimization (e.g., gradient descent). However, in contrast to base-level techniques, optimization-based meta-learners can learn the optimizer and/or are exposed to multiple tasks, which allows them to learn how to learn new tasks quickly.   
\autoref{fig:optbasedrels} shows the relationships between the covered optimization-based techniques.

As we can see, the LSTM optimizer \citep{andrychowicz2016learning}, which replaces hand-crafted optimization procedures such as gradient descent by a trainable LSTM, can be seen as the starting point for these optimization-based meta-learning techniques. 
\citet{li2018learning} also aim to learn the optimization procedure with reinforcement learning instead of gradient-based methods. 
The LSTM meta-learner \citep{Ravi2017} extends the LSTM optimizer to the few-shot setting by not only learning the optimization procedure but also a good set of initial weights.
This way, it can be used across tasks.
MAML \citep{Finn17} is a simplification of the LSTM meta-learner as it replaces the trainable LSTM optimizer by hand-crafted gradient descent.
MAML has received considerable attention within the field of deep meta-learning, and has, as one can see, inspired many other works. 

Meta-SGD is an enhancement of MAML that not only learns the initial parameters, but also the learning rates \citep{li2017metasgd}. 
LLAMA \citep{grant2018recasting}, PLATIPUS \citep{finn2018probabilistic}, and online MAML \citep{finn2019online} extend MAML to the active and online learning settings. 
LLAMA and PLATIPUS are probabilistic interpretations of MAML, which allow them to sample multiple solutions for a given task and quantify their uncertainty.
BMAML \citep{yoon2018bayesian} takes a more discrete approach as it jointly optimizes a discrete set of $M$ initializations. 
iMAML \citep{rajeswaran2019meta} aims to overcome the computational expenses associated with the computation of second-order derivatives, which is needed by MAML. 
Through implicit differentiation, they also allow for the use of non-differentiable inner loop optimization procedures. 
Reptile \citep{nichol2018reptile} is an elegant first-order meta-learning algorithm for finding a set of initial parameters and removes the need for computing higher-order derivatives. 
LEO \citep{rusu2018meta} tries to improve the robustness of MAML by optimizing in lower-dimensional parameter space through the use of an encoder-decoder architecture. 
Lastly, R2-D2, LR-D2 \citep{Bertinetto19}, and \cite{lee2019meta} use simple classical machine learning methods (ridge regression, logistic regression, SVM, respectively) as a classifier on top of a learned feature extractor.

A key advantage of optimization-based approaches is that they can achieve better performance on wider task distributions than, e.g., model-based approaches \citep{finn2018meta}. However, optimization-based techniques optimize a base-learner for every task that they are presented with and/or learn the optimization procedure, which is computationally expensive \citep{hospedales2020meta}. 

Optimization-based meta-learning is a very active area of research. We expect future work to be done in order to reduce the computational demands of these methods and improve the solution quality and level of generalization. We think that benchmarking and reproducibility research will play an important role in these improvements.

\section{Concluding Remarks}\label{sec:conclusion}

In this section, we give a helicopter view of all that we discussed, and the field of Deep Meta-Learning in general. We will also discuss challenges and future research. 

\subsection{Overview}

In recent years, there has been a shift in focus in the broad meta-learning community. Traditional algorithm selection and hyperparameter optimization for classical machine learning techniques (e.g.\ Support Vector Machines, Logistic Regression, Random Forests, etc.) have been augmented by Deep Meta-Learning, or equivalently, the pursuit of self-improving neural networks that can leverage prior learning experience to learn new tasks more quickly. Instead of training a new model from scratch for different tasks, we can use the same (meta-learning) model across tasks. As such, meta-learning can widen the applicability of powerful deep learning techniques to domains where fewer data are available and computational resources are limited.

Deep Meta-Learning techniques are characterized by their meta-objective, which allows them to maximize performance across various tasks, instead of a single one, as is the case in base-level learning objectives. This meta-objective is reflected in the training procedure of meta-learning methods, as they learn on a set of different meta-training tasks. The few-shot setting lends itself nicely towards this end, as tasks consist of few data points. This makes it computationally feasible to train on many different tasks, and it allows us to evaluate whether a neural network can learn new concepts from few examples. 
Task construction for training and evaluation does require some special attention. That is, it has been shown beneficial to match training and test conditions \citep{vinyals2016matching}, and perhaps train in a more difficult setting than the one that will be used for evaluation \citep{snell2017prototypical}. 

On a high level, there are three categories of Deep Meta-Learning techniques, namely i)~metric-\index{meta-learning!metric-based}, ii)~model-\index{meta-learning!model-based}, and iii)~optimization-based\index{meta-learning!optimization-based} ones, which rely on i) computing input similarity, ii) task embeddings\index{task embedding} with states, and iii) task-specific updates, respectively. Each approach has strengths and weaknesses. Metric-learning techniques are simple and effective \citep{garcia2017few} but are not readily applicable outside of the supervised learning setting \citep{hospedales2020meta}. Model-based techniques, on the other hand, can have very flexible internal dynamics, but lack generalization ability to more distant tasks than the ones used at meta-train time \citep{finn2018meta}. Optimization-based approaches have shown greater generalizability, but are in general computationally expensive, as they optimize a base-learner for every task \citep{finn2018meta, hospedales2020meta}.

\autoref{tab:categorization} provides a concise, tabular overview of these approaches. Many techniques have been proposed for each one of the categories, and the underlying ideas may vary greatly, even within the same category. \autoref{tab:overview}, therefore, provides an overview of all methods and key ideas that we have discussed in this work, together with their applicability to supervised learning (SL) and reinforcement learning (RL) settings, key ideas, and benchmarks that were used for testing them. 
\autoref{tab:performance} displays an overview of the 1- and 5-shot classification performances (reported by the original authors) of the techniques on the frequently used miniImageNet benchmark. 
Moreover, it displays the used backbone (feature extraction module) as well as the final classification mechanism. 
From this table, it becomes clear that the 5-shot performance is typically better than the 1-shot performance, indicating that data scarcity is a large bottleneck for achieving good performance. 
Moreover, there is a strong relationship between the expressivity of the backbone and the performance. 
That is, deeper backbones tend to give rise to better classification performance.
The best performance is achieved by MetaOptNet, yielding a 1-shot accuracy of $64.09$\% and a 5-shot accuracy of $80.00$\%.
Note however that MetaOptNet used a deeper backbone than most of the other techniques.

\begin{table}[phtb]
    \begin{tabularx}{\linewidth}{lllll}
        \toprule 
         Name & Backbone & Classifier & 1-shot & 5-shot \\
         \midrule 
         \textbf{Metric-based} \\
         \, Siamese nets & - & - & -   \\
         \, Matching nets & 64-64-64-64 & Cosine sim. & $43.56 \pm 0.84$ & $55.31 \pm 0.73$     \\
         \, Prototypical nets &  64-64-64-64 & Euclidean dist. & $49.42 \pm 0.78$ & $68.20 \pm 0.66$  \\
         \, Relation nets &64-96-128-256 &  Sim. network & $50.44 \pm 0.82$ & $65.32 \pm 0.70$ \\
         \, ARC & - & 64-1 dense & $49.14 \pm -$ & - \\
         \, GNN  & 64-96-128-256 & Softmax & $50.33 \pm 0.36$& $66.41 \pm 0.63$ \\
         \midrule
         \textbf{Model-based}  \\
         \, RMLs  & - & - & - \\
         \, MANNs & - & - & - \\
         \, Meta nets & 64-64-64-64-64 & 64-Softmax & $49.21 \pm 0.96$ & - \\
         \, SNAIL  & Adj. ResNet-12 &  Softmax & $55.71 \pm 0.99$ & $68.88 \pm 0.92$ \\
         \, CNP  & - & - & - \\
         \, Neural stat. & - & - & - \\
         \midrule
         \textbf{Opt.-based} &  & &  \\
         \, LSTM optimizer & - & - & -  \\
         \, LSTM ml. & 32-32-32-32 & Softmax & $43.44 \pm 0.77$ & $60.60 \pm 0.71$  \\
         \, RL optimizer  & - & - & -  \\
         \, MAML  & 32-32-32-32 & Softmax & $48.70 \pm 1.84$ & $63.11 \pm 0.92$  \\
         \, iMAML  & 64-64-64-64 & Softmax & $49.30 \pm 1.88$ & - \\
         \, Meta-SGD  & 64-64-64-64 & Softmax & $50.47 \pm 1.87$ & $64.03 \pm 0.94$ \\
         \, Reptile  & 32-32-32-32 & Softmax & $48.21 \pm 0.69$& $66.00 \pm 0.62$  \\
         \, LEO  & WRN-28-10 & Softmax & $61.76 \pm 0.08$ & $77.59 \pm 0.12$  \\
         \, Online MAML  & - & -& - \\
         \, LLAMA  & 64-64-64-64 & Softmax & $49.40 \pm 1.83$ & -  \\
         \, PLATIPUS & - & -& - \\
         \, BMAML & 64-64-64-64-64 & Softmax & $53.80 \pm 1.46$ & - \\
         \, Diff. solvers  & \\
         \, \, \, \, R2-D2 & 96-192-384-512 & Ridge regr. &$51.8 \pm 0.2$ & $68.4 \pm 0.2$ \\
         \, \, \, \, LR-D2 &  96-192-384-512 & Log. regr. & $51.90 \pm 0.20$ & $68.70 \pm 0.20$ \\
         \, \, \, \, MetaOptNet & ResNet-12 & SVM & {$\boldsymbol{64.09 \pm 0.62}$} & {$\boldsymbol{80.00 \pm 0.45}$} \\
         \bottomrule
    \end{tabularx}
    \caption{Comparison of the accuracy scores of the covered meta-learning techniques on 1- and 5-shot miniImageNet classification. Scores are taken from the original papers. The $\pm$ indicates the 95\% confidence interval. The backbone is the used feature extraction module. The classifier column shows the final layer(s) that were used to transform the features into class predictions. Used abbreviations: ``sim.": similarity, ``Adj.": adjusted, and ``dist.": distance, ``log.": logistic, ``regr.": regression, ``ml.": meta-learner, ``opt.": optimization.}
    \label{tab:performance}
\end{table}

\subsection{Open Challenges and Future Work}

Despite the great potential of Deep Meta-Learning techniques, there are still open challenges, which we discuss here.  

\autoref{fig:depthandperformance}  in Section~\ref{sec:intro} displays the accuracy scores of the covered meta-learning techniques on 1-shot miniImageNet classification. 
Techniques that were not tested in this setting by the original authors are omitted.
As we can see, the performance of the techniques is related to the expressivity of the used backbone (ordered in increasing order on the x-axis).
For example, the best-performing techniques, LEO and MetaOptNet, use the largest network architectures.
Moreover, the fact that different techniques use different backbones poses a problem as it is difficult to fairly compare their classification performance.
An obvious question arises to which degree the difference in performance is due to methodological improvements, or due to the fact that a better backbone architecture was chosen.
For this reason, we think that it would be useful to perform a large-scale benchmark test where techniques are compared when they use the same backbones. 
This would also allow us to get a more clear idea of how the expressivity of the feature extraction module affects the performance. 

Another challenge of Deep Meta-Learning techniques is that they can be susceptible to the \textit{memorization problem} (meta-overfitting), where the neural network has memorized tasks seen at meta-training time and fails to generalize to new tasks. More research is required to better understand this problem.  Clever task design and meta-regularization may prove useful to avoid such problems \citep{yin2020Meta-Learning}. 

Another problem is that most of the meta-learning techniques discussed in this work are evaluated on narrow benchmark sets. This means that the data that the meta-learner used for training are not too distant from the data used for evaluating its performance. As such, one may wonder how well these techniques are able to adapt to  more distant tasks. \citet{chen2019closer} showed that the ability to adapt to new tasks decreases as they become more distant from the tasks seen at training time. Moreover, a simple non-meta-learning baseline (based on pre-training and fine-tuning) can outperform state-of-the-art meta-learning techniques when meta-test tasks come from a different data set than the one used for meta-training.

In reaction to these findings, \citet{triantafillou2019meta} have recently proposed the Meta-Dataset benchmark, which consists of various previously used meta-learning benchmarks such as Omniglot \citep{lake2011one} and ImageNet \citep{deng2009imagenet}. This way, meta-learning techniques can be evaluated in more challenging settings where tasks are diverse. Following \citet{hospedales2020meta}, we think that this new benchmark can prove to be a good means towards the investigation and development of meta-learning algorithms for such challenging scenarios. 

As mentioned earlier in this section, Deep Meta-Learning has the appealing prospect of widening the applicability of deep learning techniques to more real-world domains. For this, increasing the generalization ability of these techniques is very important. Additionally, the computational costs associated with the deployment of meta-learning techniques should be small. While these techniques can learn new tasks quickly, meta-training can be quite computationally expensive. Thus, decreasing the required computation time and memory costs of Deep Meta-Learning techniques remains an open challenge. 

Some real-world problems demand systems that can perform well in online, or active learning settings. The investigation of Deep Meta-Learning in these settings \citep{finn2018probabilistic, yoon2018bayesian, finn2019online, munkhdalai2017meta, vuorio2018meta} remains an important direction for future work.  

Yet another direction for future research is the creation of \textit{compositional} Deep Meta-Learning systems, which instead of learning flat and associative functions $\boldsymbol{x} \rightarrow y$,  organize knowledge in a \textit{compositional} manner. This would allow them to decompose an input $\boldsymbol{x}$ into several (already learned) components $c_{1}(\boldsymbol{x}),\ldots ,c_{n}(\boldsymbol{x})$, which in turn could help the performance in low-data regimes \citep{tokmakov2019learning}. 

The question has been raised whether contemporary Deep Meta-Learning techniques actually learn how to perform rapid learning, or simply learn a set of robust high-level features, which can be (re)used for many (new) tasks. \citet{raghu2020rapid} investigated this question for the most popular Deep Meta-Learning technique MAML and found that it largely relies on feature reuse. It would be interesting to see whether we can develop techniques that rely more upon fast learning, and what the effect would be on performance.   

Lastly, it may be useful to add more meta-abstraction levels, giving rise to, e.g., meta-meta-learning, meta-meta-...-learning \citep{hospedales2020meta, schmidhuber1987evolutionary}.

\begin{acknowledgements}
Thanks to Herke van Hoof for an insightful discussion on LLAMA. Thanks to Pavel Brazdil for his encouragement and feedback on a preliminary version of this work.
\end{acknowledgements}

%
%

\bibliographystyle{spbasic}      
\bibliography{refs}   

%
%

\end{document}